\documentclass[10pt,twocolumn,letterpaper]{article}
\usepackage[accsupp]{axessibility}
\usepackage[pagenumbers]{cvpr} % To force page numbers, e.g. for an arXiv version

\definecolor{cvprblue}{rgb}{0.21,0.49,0.74}
\usepackage[pagebackref,breaklinks,colorlinks,allcolors=cvprblue]{hyperref}

\def\paperID{14314} % *** Enter the Paper ID here
\def\confName{CVPR}
\def\confYear{2025}

\usepackage{url}

\usepackage{enumitem}
\usepackage{fancybox}
\usepackage{booktabs}
\usepackage{graphicx}
\usepackage{multirow}
\usepackage{makecell}
\usepackage{textcomp}
\usepackage{bbding}
\usepackage{pifont}
\usepackage{color, colortbl}
\usepackage{adjustbox}
\usepackage{float}

\newcommand\blfootnote[1]{%
  \begingroup
  \renewcommand\thefootnote{}\footnote{#1}%
  \addtocounter{footnote}{-1}%
  \endgroup
}

\usepackage[capitalize]{cleveref}
\definecolor{Gray}{gray}{0.94}
\definecolor{Red}{RGB}{255, 46, 23}
\definecolor{Green}{RGB}{0, 171, 79}
\definecolor{cred}{rgb}{0.85, 0.1, 0.15}
\definecolor{cgreen}{rgb}{0.25, 0.68, 0.28}
\definecolor{royalblue}{rgb}{0.25, 0.5, 0.75}
\definecolor{grayblue}{rgb}{0.9, 0.92, 0.95}
\hypersetup{
colorlinks=true,
linkcolor=red,
urlcolor=magenta
}

\usepackage{xargs}

\title{RAP: Retrieval-Augmented Personalization for Multimodal Large\\ Language Models}

\author{Haoran Hao$^{1,2*}$, Jiaming Han$^{1*}$, Changsheng Li$^{3}$, Yu-Feng Li$^{2}$, Xiangyu Yue$^{1,4\dag}$ \vspace{0.2cm}\\
$^{1}$MMLab, The Chinese University of Hong Kong \\
$^{2}$National Key Laboratory for Novel Software Technology, Nanjing University\\
$^{3}$Beijing Institute of Technology \ \ $^{4}$SHIAE, CUHK
}

\begin{document}
\twocolumn[{%
\maketitle
\renewcommand\twocolumn[1][]{#1}%
\vspace{-0.35in}
\begin{center}
    \centering
  \includegraphics[width=0.96\linewidth]{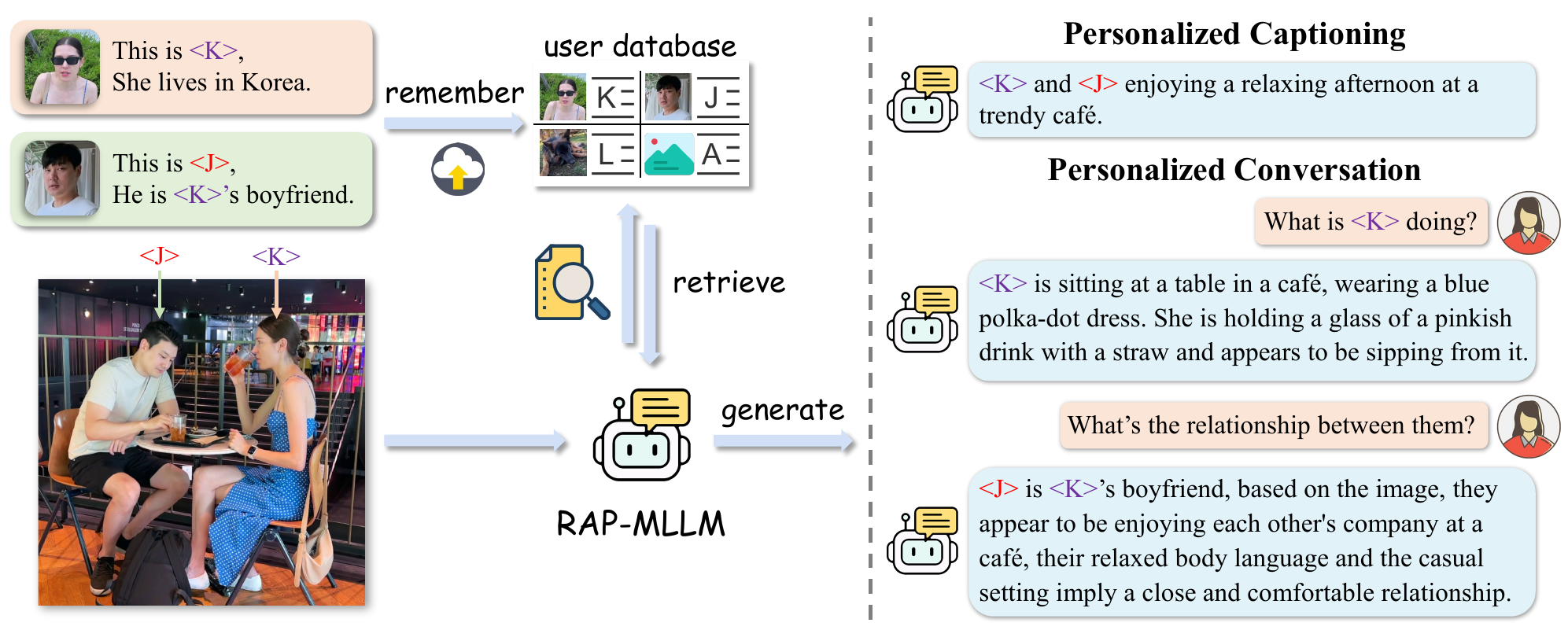}
    \vspace{-0.25cm}
    \captionof{figure}{Introduce some user-specific concepts to our RAP-MLLM, it can remember them and achieve excellent performance in a variety of personalized multimodal generation tasks.}
    \label{fig:teaser}
\end{center}
}]
\blfootnote{$^*$ Equal contribution\ \ $^{\dagger}$ Corresponding author}

\begin{abstract}
The development of large language models (LLMs) has significantly enhanced the capabilities of multimodal LLMs (MLLMs) as general assistants. 
However, lack of user-specific knowledge still restricts their application in human's daily life. In this paper, we introduce the \textbf{R}etrieval \textbf{A}ugmented \textbf{P}ersonalization~(RAP) framework for MLLMs' personalization. Starting from a general MLLM, we turn it into a personalized assistant in three steps. (a) Remember: We design a key-value database to store user-related information, \eg, user's name, avatar and other attributes. (b) Retrieve: When the user initiates a conversation, RAP will retrieve relevant information from the database using a multimodal retriever. (c) Generate: The input query and retrieved concepts' information are fed into MLLMs to generate personalized, knowledge-augmented responses. Unlike previous methods, RAP allows real-time concept editing via updating the external database.
To further improve generation quality and alignment with user-specific information, we design a pipeline for data collection and create a specialized dataset for personalized training of MLLMs. Based on the dataset, we train a series of MLLMs as personalized multimodal assistants.
By pretraining on large-scale dataset, RAP-MLLMs can generalize to infinite visual concepts without additional finetuning.
Our models demonstrate outstanding flexibility and generation quality across a variety of tasks, such as personalized image captioning, question answering and visual recognition. The code, data and models are available at \url{https://hoar012.github.io/RAP-Project/}.
\end{abstract}
\section{Introduction}
\label{sec:intro}

\begin{table*}[t]
\centering
\caption{\textbf{Comparison of Different Personalization Methods.} RAP needs only 1 image with its personalized description, showing outstanding convenience and flexibility in practical applications.}
\label{Cost}
\vspace{-3mm}
\setlength{\tabcolsep}{1.1mm}{\begin{tabular}{lcccccccc}
\toprule
&\multicolumn{2}{c}{Number of Image} &\multicolumn{4}{c}{Data Requirements for Personalization} &\multicolumn{2}{c}{Support}\\
\cmidrule(lr){2-3}\cmidrule(lr){4-7}\cmidrule(lr){8-9}
Method & Positive& Negative& Caption & Description &Question-Answer &Recognition &Real-time edit & Text-only QA \\
\midrule
Fine-tuning & n & - &Yes &Yes& No& No& \ding{55} & \ding{51}\\
MyVLM~\cite{myvlm} & n & 150 & Yes& No &Yes& Yes&\ding{55} & \ding{55}\\
Yo'LLaVA~\cite{yollava} & n & 200 & No & No &Yes& Yes& \ding{55} & \ding{51}\\
\rowcolor{Gray}RAP(Ours) & 1 & - & No &Yes &No& No&\ding{51} & \ding{51}\\
\bottomrule
\end{tabular}}
\vspace{-1mm}
\end{table*}

Recently, the development of large language models (LLMs) has significantly enhanced their language processing and generating capabilities~\cite{survey_llm}. Building on this foundation, the integration of visual and textual ability through vision-language alignment brings powerful multimodal LLMs (MLLMs)~\cite{survey_mllm, GPT4, gemini, llava, internlm, onellm, internvideo2}. 
MLLMs have shown significant improvement in various tasks, such as image description and question answering, highlighting their potential as humans' assistants. However, their lack of user-specific knowledge continues to limit their effectiveness as personalized assistants in daily life.

A qualified personalized assistant first should be able to recognize and remember user-related concepts, such as the dog named \textlangle Lala\textrangle\ adopted by the user. Although existing MLLMs have been trained on large-scale datasets and possess strong recognition and classification capabilities, directly transferring this knowledge to a user's personal concepts remains challenging. For instance, current leading MLLMs cannot remember your dog's name, even if you have mentioned it before, and they lack awareness of your identity and preferences.
Furthermore, the assistant should generate responses tailored to the user's preferences and requirements. However, collecting extensive personal data to train a unique assistant for each user is impractical.

To address this issue, the personalization of MLLMs has gained increasing attention, with several approaches already being proposed.
MyVLM~\cite{myvlm} utilizes external classification heads to recognize specific concepts, and learns an embedding for each concept to personalize the outputs of vision language models (VLMs)~\cite{llava, li2023blip}. Another concurrent work, Yo'LLaVA~\cite{yollava}, learns a few special tokens to represent each concept. 
However, both approaches require continuous learning and model updates as new concepts emerge. As shown in Table~\ref{Cost}, they require multiple labeled images of the target concept along with a large number of negative images, making data collection a significant challenge. Alternatively, fine-tuning the model for each new concept also incurs substantial computational costs. 
This presents a challenge in dynamic, ever-changing real-world scenarios, where the computing power of users' personal devices is often limited, and all data must be stored locally for privacy concerns.

To address these challenges, we propose the \textbf{R}etrieval \textbf{A}ugmented \textbf{P}ersonalization~(RAP) framework, designed to allow MLLMs to update their supported concepts without additional training. Specifically, RAP works in three key steps. (a) Remember: RAP includes a designed database to help remember each concept via storing its image and basic information, \eg, name, avatar and other attributes. (b)~Retrieve: When a user initiates a conversation, RAP will retrieve relevant information from the database using a multimodal retriever. (c) Generate: The input query and retrieved concepts information are incorporated into the MLLM's input for personalized, knowledge-augmented generation. As shown in Table~\ref{Cost}, our RAP requires only one image per concept with its related information for personalization. At the same time, it allows users to make real-time adjustments to the model's outputs by modifying their personal databases, eliminating the need for retraining. Examples of real-time concept editing are presented in Table~\ref{appendix:editing}.

% As shown in Table~\ref{Cost}, our RAP framework requires only a single image per concept, along with its related information, for personalization. Additionally, it enables users to adjust the model’s outputs in real time by modifying their personal databases, eliminating the need for retraining while allowing on-the-fly editing of personalized generations. Examples of real-time concept editing are presented in Table~\ref{appendix:editing}.

Another significant challenge is the lack of large-scale datasets for training MLLMs' personalized generation capabilities. To address this, we design a pipeline to collect extensive training data and create a comprehensive dataset, which enables to train MLLMs to effectively understand and utilize user-related information for generation. Based on this dataset, we train LLaVA~\cite{llava} and Phi3-V~\cite{hanoona2024LLaVA++} as novel personalized assistants and evaluate their performance across various tasks, including personalized image captioning, question answering, and visual recognition. Experimental results demonstrate that our RAP-MLLMs excel in a wide range of personalized generation tasks, showcasing excellent generation quality and flexibility.

Our contributions are summarized as follows: 
\begin{itemize}[leftmargin=18 pt, itemsep= 1 pt,topsep = -1 pt]
\item We propose the RAP framework for MLLMs' personalization, allowing models pre-trained on our dataset to adapt to diverse users and infinite new concepts without further training.
\item We develop a pipeline for collecting large-scale data and create a dataset specifically designed for the personalized training and evaluation of MLLMs. This dataset enables us to train a series of MLLMs to function as personalized assistants.
\item Our models demonstrate exceptional performance across various personalized multimodal generation tasks, including image captioning and question answering. Additionally, they exhibit a strong capability to recognize personal concepts within images.
\end{itemize}

\section{Related Work}
\noindent \textbf{Multimodal Large Language Models.}
Recently, numerous advanced large language models (LLMs)~\cite{llama, llama-adapter, chiang2023vicuna, taori2023stanford, abdin2024phi} have been proposed, showing remarkable performance in addressing a wide range of tasks. The rapid development of LLMs has led to the emergence of multimodal LLMs (MLLMs)~\cite{GPT4, gemini, llava, internlm, onellm, minigpt, internvideo2}, which excel in general visual understanding and complex reasoning tasks. For instance, LLaVA~\cite{llava, improvedbaselinesvisualinstruction} and MiniGPT-4~\cite{minigpt} align visual and language modalities through visual instruction tuning, showcasing impressive capabilities in multimodal conversations. GPT4RoI~\cite{gpt4roi} and RegionGPT~\cite{regiongpt} enhance fine-grained understanding and reasoning for specific regions by training on region-level instruction datasets. Despite these advancements in tasks such as image captioning and question answering, the lack of user-specific knowledge restricts the generation of personalized content. In this work, we focus on the personalization of MLLMs, enabling them to remember and understand user-specific concepts, and generate personalized content tailored to user preferences.

\noindent \textbf{Personalization of MLLMs.}
In the realm of artificial intelligence, personalization typically refers to the process of customizing a system, application, or model to meet individual needs and preferences~\cite{metapersonalizingvlm, personalizedllms, ERAGent, wu2024understanding}. Substantial efforts have been made to generate images of a user's personal objects or within certain contexts~\cite{Dreambooth, custom_diffusion, personalizedresidualsconceptdriventexttoimage, imageworthword, ipadapter, shi2024instantbooth, blip_diffusion}. For example, Dreambooth~\cite{Dreambooth} employs transfer learning in text-to-image diffusion models via fine-tuning all parameters for new concepts. In this paper, we mainly aim at enabling MLLMs to remember and understand user-specific concepts, and generate personalized language outputs. Several studies have focused on the personalization of MLLMs, among which the most relevant works are MyVLM~\cite{myvlm} and Yo'LLaVA~\cite{yollava}. MyVLM introduces the task of personalizing VLMs. It utilizes external classification heads to recognize specific concepts, and learns an embedding for each concept to personalize the outputs of VLMs. Yo'LLaVA personalizes LLaVA by extending its vocabulary and learning specific tokens for each concept. However, both approaches require continuous model updates as new concepts emerge, which presents challenges in dynamic real-world applications.
In this work, we propose the RAP framework for the personalization of MLLMs, enabling models pre-trained on our dataset to continuously update supported concepts without additional fine-tuning.

\noindent \textbf{Retrieval Augmented Generation.}
Retrieval-based methods for incorporating external knowledge have been effective in improving generation across various knowledge-intensive tasks~\cite{rallm, retrieval_mm, selfRAG, RAGlongcontext, RAGrobust, Ra-dit}. DPR~\cite{DPR} introduces Dense Passage Retrieval, marking a shift from sparse to dense retrieval techniques. Later, MuRAG~\cite{chen2022murag} proposes to use multimodal knowledge to augment language generation. Self-Rag~\cite{selfRAG} introduces special tokens to make retrieval adaptive and controllable. ERAGent~\cite{ERAGent} presents a comprehensive system for retrieval-augmented language models. With the advancements in MLLMs, RAG has been widely applied to multimodal generative tasks. For instance, FLMR~\cite{FLMR} employs multi-dimensional embeddings to capture finer-grained relevance between queries and documents, achieving significant improvement in the RA-VQA setting. While existing methods primarily enhance models' performance by retrieving from external knowledge bases, few of them consider the personalization task. Although RAG has been applied to image generation~\cite{rag_diffution, ReMoDiff} and image captioning~\cite{EVCAP, ramos2023retrieval}, there is currently no existing work that focuses on personalizing MLLMs through RAG, to the best of our knowledge.

\section{Retrieval Augmented Personalization}
Existing MLLMs typically align other modalities with language. For instance, LLaVA~\cite{llava} projects visual tokens into text space, and then generates subsequent tokens using an LLM. While these MLLMs perform well in various tasks, the lack of memory and comprehension of personal concepts hinders effective user-specific responses. In this work, we mainly focus on personalizing MLLMs to generate tailored language responses, such as creating personalized captions for user's images and answering questions about personal concepts. 
In this section, we detail the implementation steps of our proposed Retrieval-Augmented Personalization~(RAP) framework. Unlike previous approaches that usually necessitate additional data collection and further training to learn new concepts, after pre-training on our dataset, RAP-MLLMs can adapt to diverse users and infinite new concepts without further training. In section~\ref{frame}, we present the RAP framework that is applicable to various types of MLLMs, and then in section~\ref{dataset}, we provide details of the proposed dataset.

\subsection{RAP Framework}
\label{frame}

\begin{figure*}[t]
  \centering
  \includegraphics[width=0.9\linewidth]{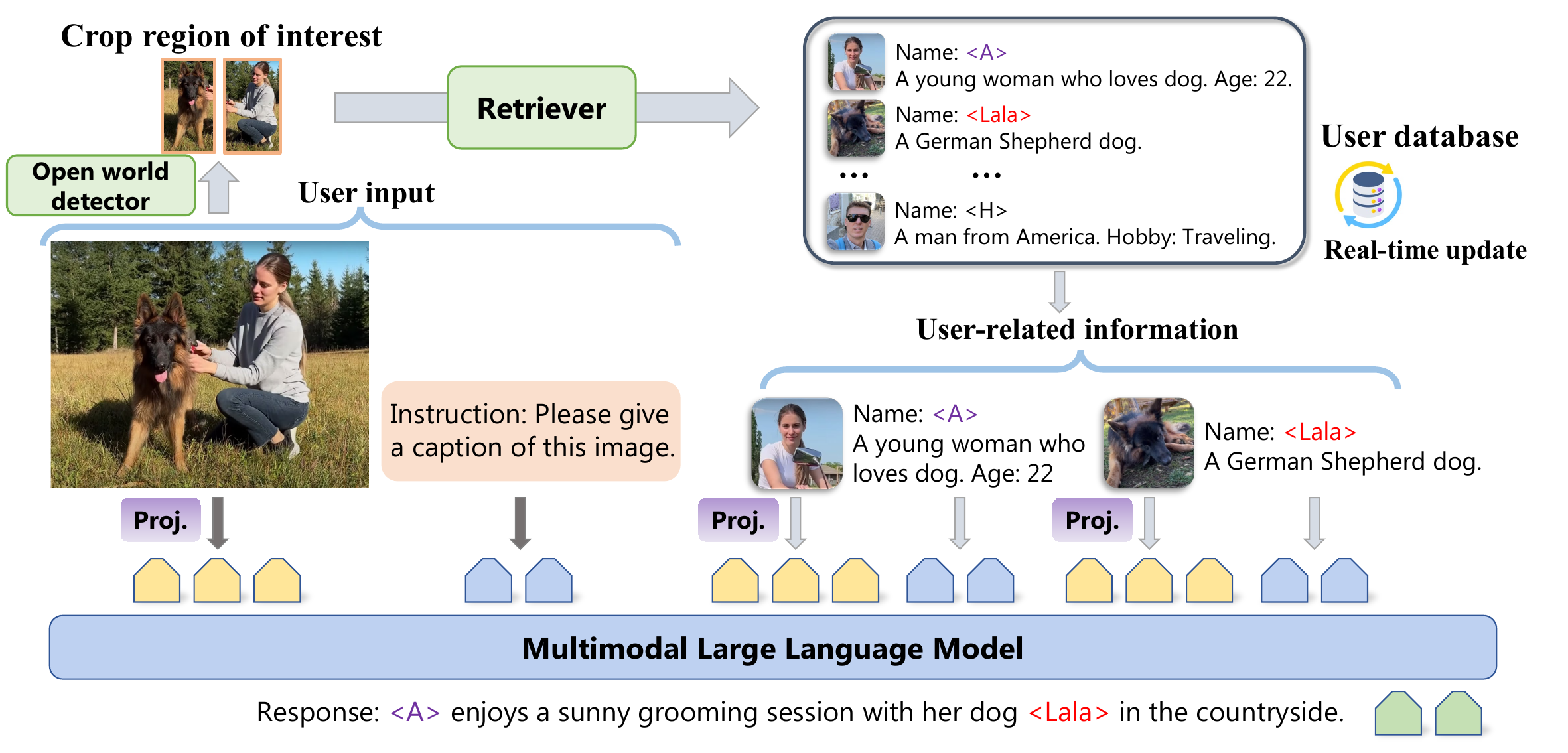}
  \vspace{-0.4cm}
  \caption{\textbf{Retrieval-Augmented Personalization Framework}. Region-of-interest detected by an open world detector are used to retrieve concepts from the database. The images and information of the retrieved concepts are then integrated into the input for the MLLM.}
  \vspace{-0.25cm}
  \label{framework}
\end{figure*}

RAP works in three steps: Remember, Retrieve and Generate. An overview of the framework is shown in Figure~\ref{framework}.

\noindent \textbf{Remember.}
The premise of personalization is that the model can remember personal concepts and relevant information, such as the dog named \textlangle Lala\textrangle\ adopted by \textlangle A\textrangle. To facilitate this, we construct a database $\mathcal{M}$ to store personal concepts, which comprises an avatar $\mathbf{I}_j$, a name along with a brief description $\mathbf{T}_j$ for each concept. The key $k_i$ for each concept in the database is its visual feature, obtained by feeding its image $\mathbf{I}_j$ into a pre-trained image encoder $\mathcal{E}(\cdot)$. Examples of our database are presented in Figure~\ref{framework}. When a user initiates a conversation, the input can be represented as $Q=(\mathbf{X}_v, \mathbf{X}_q)$, which may include both image $\mathbf{X}_v$ and some textual instructions $\mathbf{X}_q$. The first step involves identifying possible concepts within the input image that have been previously stored in the database. 
Previous methods~\cite{myvlm} typically need to learn an external classifier to determine whether a concept appears in the input image, which requires a substantial amount of training data and can only apply to specific concept. To enhance the generalizability of the recognition process, we do not construct specific modules for each concept. Instead, we employ a universal detection model, such as YOLO~\cite{YOLO} and YOLO-World~\cite{yoloworld}, as recognition model $\mathcal{R}(\cdot)$. Given the predefined setting $P$ that specifies which categories should be remembered, the region-of-interest can be acquired via
$\mathbf{X}_u=\mathcal{R}(\mathbf{X}_v, {\mathbf{X}_q}|{P})$.

\noindent \textbf{Retrieve.}
Identified region-of-interest will be used as query to retrieve from the database. For each recognized component $\mathbf{X}_u^i$, we feed the image crop into the image encoder $\mathcal{E}(\cdot)$ to get its visual feature $v_i=\mathcal{E}(\mathbf{X}_u^i)$, which is a n-dimensional vector. 
Then we calculate the euclidean distance between the visual feature and each key $k_j\in \mathcal{M}$, which is calculated as $Dist(v_i, k_j)=\left \| v_i - k_j \right \| $. The Top-K image-text pairs $\{(\mathbf{I}_1, \mathbf{T}_1), (\mathbf{I}_2, \mathbf{T}_2),\cdots (\mathbf{I}_K, \mathbf{T}_K)\}$ with the lowest distances are selected from the database. We also introduce retrieval using concept names, such as $\left \langle \text{sks} \right \rangle $ for a unique concept. When the user mentions the name of an object documented in the database, our model retrieves its related information from the database. This also enables our model to respond to text-only queries effectively.

\noindent \textbf{Generate.}
Each pair $\mathbf{M}_j = (\mathbf{I}_j, \mathbf{T}_j)$ provides related information about a user's personal concept and will be incorporated into the input of the MLLM. Take LLaVA~\cite{llava} as an example. The image $\mathbf{I}_j$ is first encoded by a pre-trained vision encoder, such as CLIP~\cite{clip}, to obtain their visual tokens $\mathbf{Z}_j$. These image tokens are then projected by a projector into language tokens $\mathbf{H}_j^v$, which could be understood by the language model. Simultaneously, corresponding text information $\mathbf{T}_j$ are transformed into text tokens $\mathbf{H}_j^q$. This process is also applied to both $\mathbf{X}_v$ and $\mathbf{X}_q$. All these tokens are incorporated into the MLLM's input to generate language response. During training, we keep parameters of both the detector and retriever frozen, just train the MLLM's parameters $\theta$. Given the length $L$ of the output sequence, the probability of the target answer $\mathbf{X}_a$ is computed as:
\begin{equation}
\prod_{i=1}^{L} p_{\theta}(\mathbf{X}_{a,i}|\mathbf{X}_v,\mathbf{X}_q,\mathbf{M}_1,\cdots \mathbf{M}_K, \mathbf{X}_{a,<i}).
\end{equation}

\subsection{Personalization Dataset}
\begin{figure*}[t]
  \centering
  \includegraphics[width=0.9\linewidth]{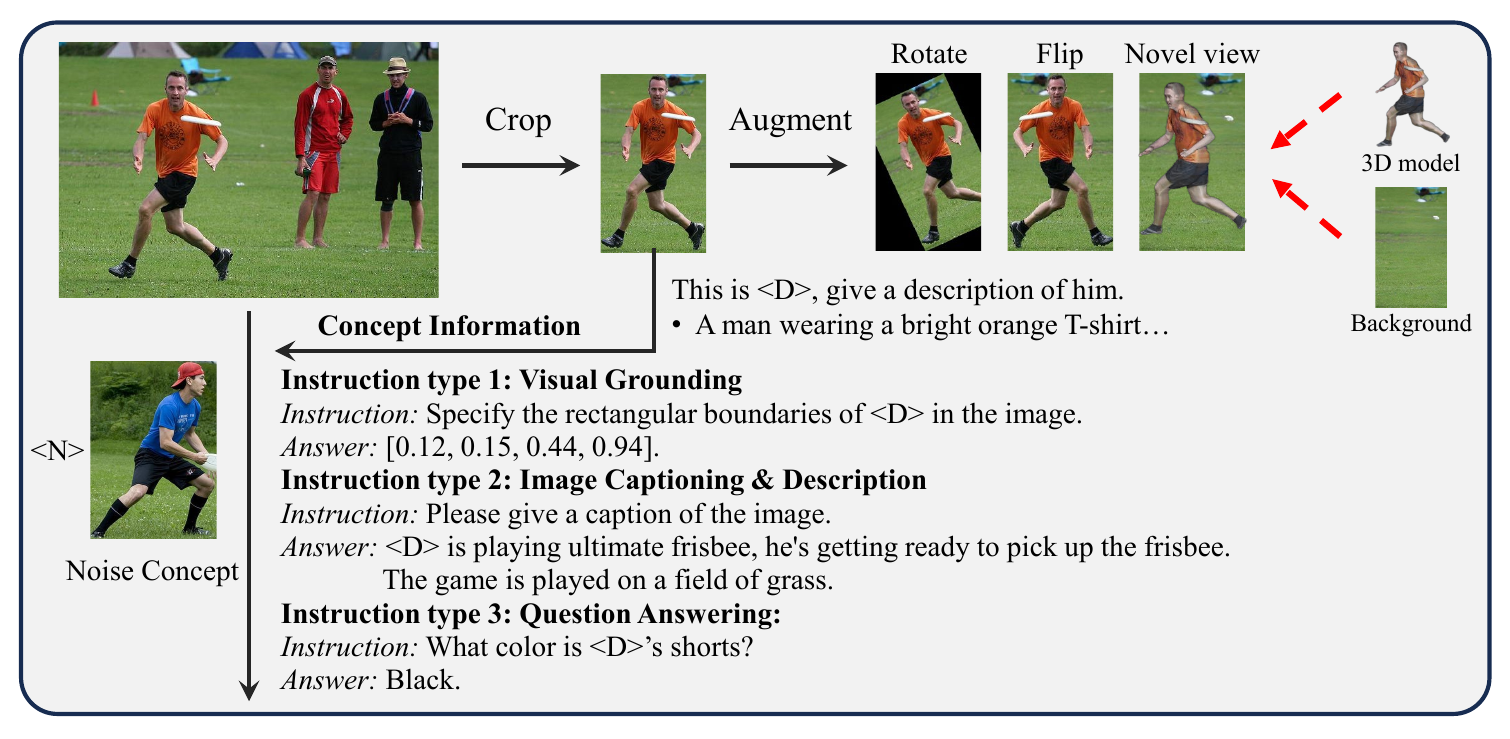}
  \vspace{-0.4cm}
  \caption{\textbf{Our Pipeline for Data Collection.} We first crop the target concept from the image based on the dataset annotations and then query Gemini~\cite{gemini} to generate its personalized description. We also apply data augmentation to diversify these cropped images. Then we combine them with the original image to derive a series of instructions and answers from Gemini. When noise concepts are included in the additional information, the answer remains unchanged, helping to train the MLLMs' ability to filter out irrelevant concepts.}
  \vspace{-0.4cm}
  \label{data-pipeline}
\end{figure*}
\label{dataset}
Most existing MLLMs struggle to generate personalized outputs even if additional concept information is provided, and there is currently no large-scale dataset for personalized training of MLLMs. To this end, we design a pipeline for data creation and curate a novel dataset specifically for the personalized training of MLLMs. We use Gemini-1.5~\cite{gemini} to generate annotations for our dataset. An overview of our pipeline and dataset is presented in Figure~\ref{data-pipeline}.

The first component of our dataset is dedicated to visual grounding. In this task, an MLLM is trained to determine whether a specific concept is present in an image, particularly identifying if the person or object in a reference image appears in the given image. When a positive match is detected, we also require the model to provide the bounding box for the identified concept. For single-concept grounding, we primarily use the RefCOCO dataset~\cite{refcoco}. Based on RefCOCO’s annotations, we crop target concepts from the images and assign names to them, which serve as references for specific concepts. We then query Gemini to generate concise descriptions about properties of the concepts in these cropped regions, by which we construct a large-scale database including numerous different concepts. The training data pairs images and these descriptions as queries and the corresponding bounding boxes as outputs. However, data generated in this way is insufficient to simulate the complexity of real-world recognition, especially when the target concept in the reference and input image is captured from different perspectives. To address this, we incorporate the ILSVRC2015-VID video object detection dataset~\cite{imagenet}, TAO~\cite{TAO} and CustomConcept101~\cite{custom_diffusion} to enrich our dataset. For multi-object grounding, we use the Object365 dataset~\cite{object365} to construct our training data.

The second component of our dataset is designed for instruction following. This section includes training data for tasks such as image captioning, image description, and question answering. For the image captioning and description data, we provide cropped images of target concepts, accompanied by their names and related information from the large-scale database, then query Gemini to generate a caption or description that reflects the concepts depicted in the entire image. For question answering, we first design a set of seed questions to serve as examples. These examples are used to prompt the annotator, Gemini, to generate new questions and corresponding answers. This iterative process facilitates the creation of a rich and diverse collection of conversations that MLLMs can learn from. We construct such data using RefCOCO~\cite{refcoco}, Object365~\cite{object365}, TAO~\cite{TAO} and CustomConcept101~\cite{custom_diffusion} dataset.

To enhance alignment with real-world scenarios, it is essential to collect data featuring the same identity in various environments. Thus, we also include multiple images about the same individual from the CelebA dataset~\cite{celebA} and produce question answering data about the individual. To further diversify the dataset, we apply image editing techniques for data augmentation. This includes performing random rotations and flips on the cropped images, as well as generating novel views of the concepts by diffusion models. Specifically, we use Inpaint-Anything \cite{inpaint-anything} to separate the foreground from the background, and use Wonder3D~\cite{long2024wonder3d} and SiTH~\cite{ho2024sith} to synthesize novel views of foreground object or person, respectively. Finally, we combine these elements to generate images of the target concept from different perspectives.

To support multi-concept personalized generation, it is necessary to retrieve multiple potential concepts from the database. In the generation step, the MLLM must prioritize accurate and contextually relevant information.  Considering that retrieval results can be inaccurate, potentially leading to unreasonable answers, we construct negative samples by adding noise concepts to the input while preserving the original output. This approach trains the model's discrimination ability. By exposing the MLLM to both relevant and irrelevant information during training, it learns to discern and filter out noise concepts, thereby enhancing its robustness during inference. Additionally, we include a subset of the LLaVA-Instruct-665k visual instruction dataset~\cite{improvedbaselinesvisualinstruction} to retain general knowledge from the original MLLM. Further details about our dataset can be found in Appendix \ref{appendix:dataset}. 
\begin{table*}[t]
\centering
\caption{\textbf{Qualitative Comparison on Image Captioning.} Image examples of target concepts are shown in the left and captions generated are shown in the right. We use \textcolor{cgreen}{green} text to denote correct target concepts.}
\label{tab:Qualitative Comparison}
\vspace{-3mm}
\begin{tabular}{cc}
\toprule
Image & Caption\\
\midrule
\makecell[c]{
\begin{minipage}[b]{0.3\linewidth}
    \centering
    {\includegraphics[width=4.6cm]{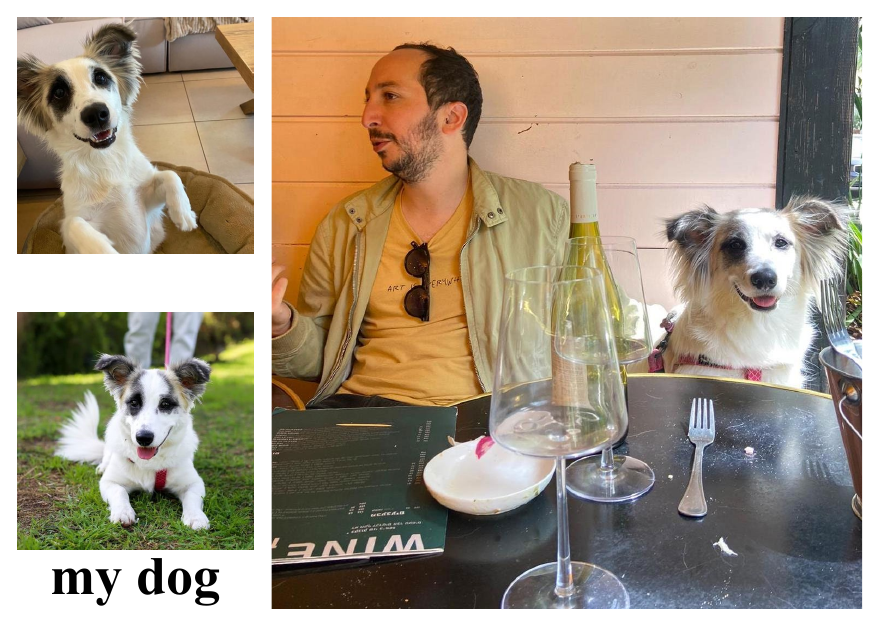}}
\end{minipage}
}& \makecell[c]{
\begin{minipage}[b]{0.62\linewidth}
    {LLaVA: A man is sitting at a table with a dog, and there are wine glasses and a fork on the table. \\
    LLaVA-LoRA: \textcolor{cred}{\textlangle collie dog\textrangle} looking pleased as she shares a meal with her owner.\\
    MyVLM: \textcolor{cgreen}{\textlangle my dog\textrangle} positioned on a chair by a black table, \textcolor{cred}{holding a wine glass in her hand. A white dog sits on the floor beside her\dots} \\
    \adjustbox{bgcolor=Gray,minipage=[t]{\linewidth}}{\textbf{RAP-LLaVA}(Ours): \textcolor{cgreen}{\textlangle my dog\textrangle} is a very good boy, and he loves to sit at a table with his owner. They are enjoying a meal.}}
\end{minipage}
} \\
\midrule
\makecell[c]{
\begin{minipage}[b]{0.3\linewidth}
    \centering
    {\includegraphics[width=4.6cm]{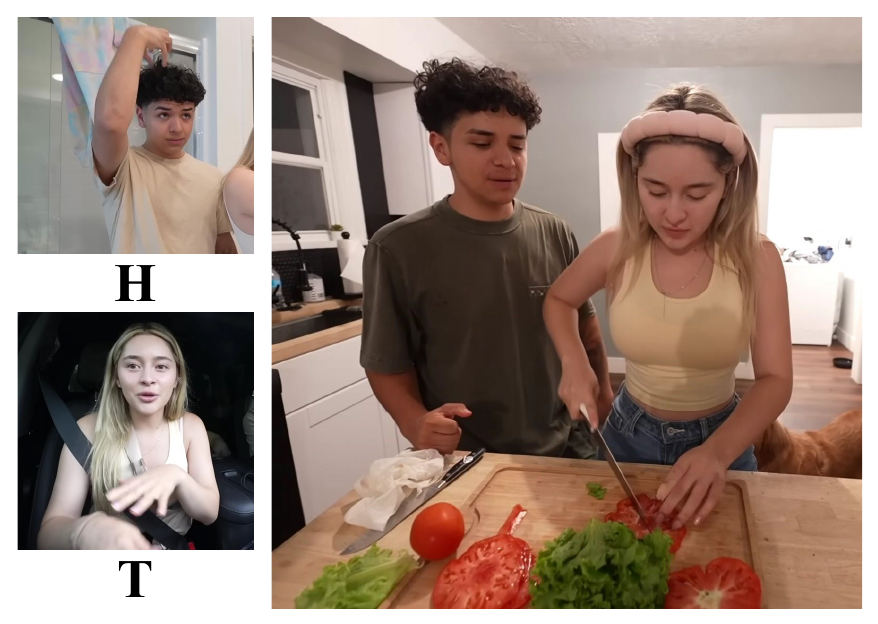}}
\end{minipage}
}& \makecell[c]{
\begin{minipage}[b]{0.62\linewidth}
    {LLaVA: A man and a woman are standing in a kitchen, preparing food together. The woman is cutting lettuce on a cutting board, while the man watches her. There are several tomatoes \dots
    \vspace{0.5mm}\\
    LLaVA-LoRA: \textcolor{cgreen}{\textlangle H\textrangle} and \textcolor{cred}{\textlangle K\textrangle} are preparing a meal together.    
    \vspace{0.5mm}\\
    MyVLM: \textcolor{cgreen}{\textlangle T\textrangle} and her friend \textcolor{cgreen}{\textlangle H\textrangle} are looking very serious as they take in the scenery.
    \vspace{0.5mm}\\
    \adjustbox{bgcolor=Gray,minipage=[t]{\linewidth}}{\textbf{RAP-LLaVA}(Ours): \textcolor{cgreen}{\textlangle H\textrangle} is helping \textcolor{cgreen}{\textlangle T\textrangle} prepare a salad in the kitchen.}}
\end{minipage}
} \\
\midrule
\makecell[c]{
\begin{minipage}[b]{0.3\linewidth}
    \centering
    {\includegraphics[width=4.6cm]{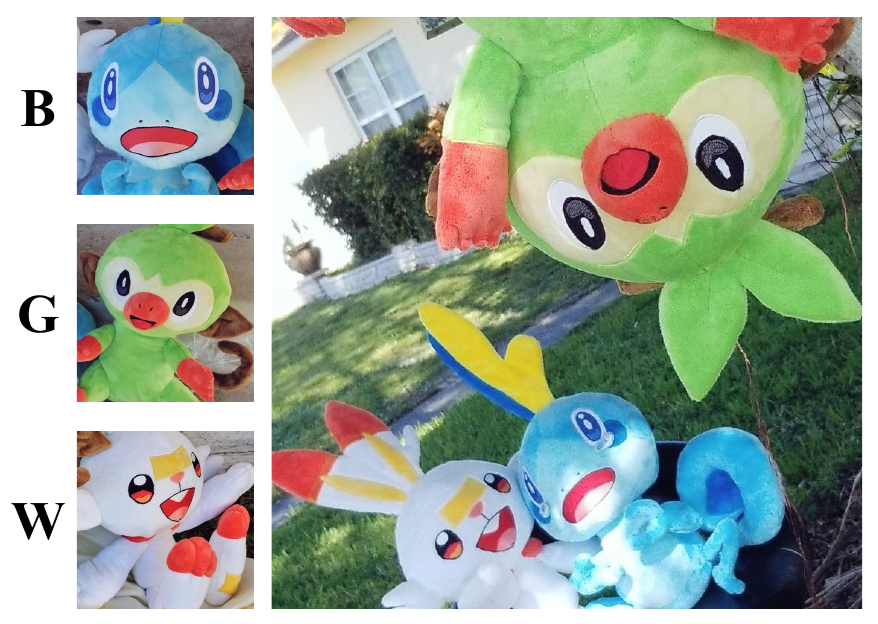}}
\end{minipage}
}& \makecell[c]{
\begin{minipage}[b]{0.62\linewidth}
    {Phi3-V: A group of stuffed animals, including a blue one, are sitting on a black surface.
    \vspace{1mm}\\
    LLaVA-LoRA: \textcolor{cgreen}{\textlangle B\textrangle}, \textcolor{cgreen}{\textlangle G\textrangle} and \textcolor{cgreen}{\textlangle W\textrangle} are happily exploring the grassland.
    \vspace{1mm}\\
    MyVLM: \textcolor{cgreen}{\textlangle G\textrangle} and his crew are always ready to jump into a new adventure.
    \vspace{1mm}\\
    \adjustbox{bgcolor=Gray,minipage=[t]{\linewidth}}{\textbf{RAP-Phi3-V}(Ours): \textcolor{cgreen}{\textlangle W\textrangle} is hanging out with \textcolor{cgreen}{\textlangle G\textrangle} and \textcolor{cgreen}{\textlangle B\textrangle} on the lawn. They are having a great time playing!}}
\end{minipage}
} \\
\bottomrule
\end{tabular}
\end{table*}

\section{Experiment}

\noindent \textbf{Implementation Details.}
We conduct experiments on LLaVA-1.5-13B~\cite{improvedbaselinesvisualinstruction} and Phi3-V-3.8B~\cite{hanoona2024LLaVA++}, resulting in two personalized MLLMs, RAP-LLaVA and RAP-Phi3-V. We select YOLO-Worldv2~\cite{yoloworld} as the detector and construct a multimodal retriever using Facebook AI Similarity Search (FAISS)~\cite{faiss}, employing a pre-trained CLIP ViT-L/14-336~\cite{clip} as the visual encoder. Due to the context length limitation of the backbone language model, for RAP-LLaVA and RAP-Phi3-V, we retrieve the 2 and 3 different concepts with the highest similarity, respectively. More implementation details can be found in Appendix~\ref{appendix:expetiment}.

\noindent \textbf{Training.} 
In the training phase, we %freeze the detector and retriever, 
skip the recognition and retrieval procedures, instead perform instruction tuning to train the MLLMs. We adhere to most settings from the original experiment of LLaVA~\cite{improvedbaselinesvisualinstruction}, except for using a maximum learning rate of 1e-4 and training for 1 epoch. We employ low-rank adapters~\cite{lora} to reduce the number of trainable parameters, and train our models on 8 A100 GPUs with a valid batch size of 64.

\begin{table*}[t]
\begin{minipage}[t]{.55\textwidth}
\caption{\textbf{Quantitative Evaluation on Image Captioning.} We report Recall, Precision and F1-score in the table, the best result in each metric is bold and the second is underlined.}
\label{caption}
\vspace{-3mm}
\setlength{\tabcolsep}{1.mm}{\begin{tabular}{l|c|ccc}
\toprule
Method  & LLM&Recall & Precision & F1-score \\
\midrule
LLaVA~\cite{improvedbaselinesvisualinstruction} + Retriever &Vicuna-13B& 1.260 & 48.76 & 2.450 \\
\midrule
LLaVA-LoRA~\cite{lora} &Vicuna-13B& 82.97 & 93.28 & 87.82 \\
MyVLM-LLaVA~\cite{myvlm}  &Vicuna-13B& 84.65 & 86.37 & 85.50 \\
\midrule
\rowcolor{Gray}RAP-LLaVA &Vicuna-13B&\textbf{93.51} &\textbf{96.47} & \textbf{94.97} \\
\rowcolor{Gray}RAP-Phi3-V &Phi3-V-3.8B& \underline{88.14} & \underline{95.10} & \underline{91.49}  \\
\bottomrule
\end{tabular}}
\end{minipage}\hspace {5mm}
\begin{minipage}[t]{.4\textwidth}
\captionof{figure}{Performance under \textbf{varying number of personalized concepts.}}
\label{fig:varying_num_concepts}
\vspace{-3mm}
\centering
{\includegraphics[width=5.4cm]{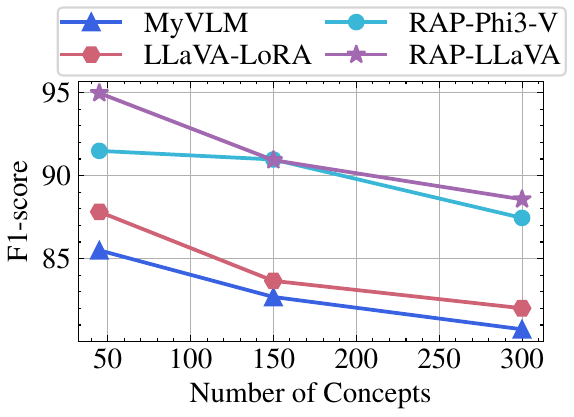}}
\end{minipage}
\end{table*}

\begin{table*}[t]
\centering
\caption{\textbf{Quantitative Evaluation on Question Answering and Visual Recognition.} The best result in each setting is bold and the second is underlined. Evaluation results of GPT-4V~\cite{GPT4} are also provided as reference. Weighted results are computed as arithmetic means.}
\label{QA}
\vspace{-3mm}
\setlength{\tabcolsep}{1.4mm}{\begin{tabular}{l|c|c|c|ccc|ccc}
\toprule
\multirow{2}{*}{Method}  &\multirow{2}{*}{LLM} &\multirow{2}{*}{Train}&\multirow{2}{*}{\#Image}& \multicolumn{3}{c|}{Question Answering} &\multicolumn{3}{c}{Visual Recognition} \\
&&& &Visual & Text-only &Weighted& Positive & Negative & Weighted\\
\midrule
GPT-4V~\cite{GPT4} + Prompt &GPT-4V &\ding{55}& 1 & 0.866 & 0.982 & 0.924 & 0.809&0.992&0.901\\
GPT-4V~\cite{GPT4} + Prompt &GPT-4V &\ding{55}& 5 & 0.887& 0.987 & 0.937 & 0.851&0.998 & 0.925\\
\midrule
LLaVA~\cite{improvedbaselinesvisualinstruction} &Vicuna-13B &\ding{55}& - & 0.899 & 0.659 & 0.779&0.000 & \textbf{1.000} &0.500 \\
LLaVA~\cite{improvedbaselinesvisualinstruction} + Retriever&Vicuna-13B &\ding{55}& 1 & 0.912 & 0.863 & 0.887& \textbf{1.000} & 0.025 &0.513 \\
\midrule
LLaVA-LoRA~\cite{lora} &Vicuna-13B &\ding{51} & 1 & 0.900 &0.583 & 0.741 & 0.988 & 0.662 & 0.825\\
LLaVA-LoRA~\cite{lora} &Vicuna-13B &\ding{51} & 5 & \underline{0.935} & 0.615 & 0.775 & \underline{0.997} & 0.444 & 0.721 \\
MyVLM-LLaVA~\cite{myvlm} &Vicuna-13B &\ding{51} & 5 & 0.912 & - & - & 0.994 & 0.845 & 0.919 \\
Yo'LLaVA~\cite{yollava} &Vicuna-13B &\ding{51} & 5 & 0.929 & \underline{0.883} &\underline{0.906} &0.949 & 0.898 & 0.924\\
\midrule
\rowcolor{Gray}RAP-LLaVA(Ours) &Vicuna-13B &\ding{55} & 1 & \underline{0.935} & \textbf{0.938} & \textbf{0.936} & 0.979 & 0.982 & \textbf{0.980}\\
\rowcolor{Gray}RAP-Phi3-V(Ours) &Phi3-V-3.8B &\ding{55} & 1 & \textbf{0.941} & 0.850 & 0.896 & 0.922 & \underline{0.988} & \underline{0.955}\\
\bottomrule
\end{tabular}}
\end{table*}

\subsection{Personalized Image Captioning}
\label{sec:exp:caption}

In this section, we evaluate our models on generating personalized image captions with user-specific concepts. We extend the dataset introduced by MyVLM~\cite{myvlm} via adding 16 new concepts, including both objects and humans, forming 8 concept pairs that appear together. For each pair, there are 8-13 images used for testing. This multi-concept setting presents additional challenges for personalization.

\noindent \textbf{Settings.} We compare our models with MyVLM~\cite{myvlm} and fine-tuning based method LLaVA-LoRA~\cite{lora}. For LLaVA-LoRA and MyVLM, the training dataset contains 1 image accompanied by 5 captions for each concept. 
For LLaVA-LoRA, we train it with captions of the training images for 3 epochs, applying low-rank adapters~\cite{lora} and the same hyperparameters as our models. For MyVLM, following their training process, we first train the classification head with the positive and 150 negative images, then train the corresponding concept embedding with the provided captions for each concept.
For our models, we construct a database where each concept is represented by a cropped image and a text description. Details of our database could be found in Appendix \ref{appendix:database}. All remaining images are used as test samples. This evaluation process is repeated three times with different seeds, resulting in a total of 1,182 images used for evaluation, and we report the average results. 

\noindent \textbf{Qualitative Comparison.} In Table~\ref{tab:Qualitative Comparison}, we present image captions generated by different methods to make a comparison. While LLaVA~\cite{improvedbaselinesvisualinstruction} and Phi3-V~\cite{hanoona2024LLaVA++} generally provide brief and clear captions for most test images, their lack of understanding of the user-specific concepts restricts them from generating a more personalized caption. LLaVA-LoRA and MyVLM can generate personalized captions, however, the limited training data often results in imprecise outputs, particularly noticeable when multiple concepts are present in the same image. 
In contrast, our models generate clear and accurate captions based on the database content, which also ensures the reliability of the outputs.
Additional examples of personalized captions generated by the models could be found in Appendix~\ref{appendix:demo}.

\noindent \textbf{Quantitative Evaluation.} We employ recall, precision and the comprehensive metric F1-score as our evaluation metrics. Recall is calculated as the percentage of correct occurrences of target concepts, while precision is the ratio of correct concept names to the total number of concept names presented. The experimental results are shown in Table \ref{caption}. 
Notably, the classification heads of MyVLM exhibit higher error rates when the number of positive images is limited, leading to weaker performance. Even with additional concept information provided through retrieval, the vanilla LLaVA~\cite{improvedbaselinesvisualinstruction} still fails to effectively accomplish the personalized generation task, underscoring the necessity of the proposed dataset. Our models demonstrate superior performance in both recall and precision metrics, highlighting the advantages of our RAP-MLLMs in data efficiency.

\subsection{Personalized Question Answering}
\label{sec:exp:QA}
\noindent \textbf{Settings.} In this section, we evaluate different methods on the benchmark of personalized question answering introduced by Yo'LLaVA \cite{yollava}, which contains both visual and text-only questions about user concepts. For each concept, we generate a description to serve as its information in our database. For LLaVA-LoRA, we feed these descriptions and corresponding images to train the model to describe the concept's properties. Additionally, we incorporate text-only queries and answers to enhance the model's textual understanding. The training dataset for Yo'LLaVA and MyVLM consists of 5 positive images with question answering pairs and 200 negative images per concept. For GPT-4V \cite{GPT4}, images and related concept information are provided as a supplementary prompt. The evaluation metric is accuracy. Additional details are provided in Appendix~\ref{appendix:expetiment}.

\noindent \textbf{Results and Analysis.} 
The experimental results are provided in Table~\ref{QA}. LLaVA and LLaVA-LoRA both perform well in visual based question answering, because substantial information of the target concept can be obtained from the images. However, their performance is quite poor when images of the target concept mentioned in the question are not available. MyVLM performs well in visual question answering but does not support text-only question answering. Yo'LLaVA excels in text-only question answering, but its performance is still limited by the insufficient information provided by the learned tokens of a concept. In contrast, our models demonstrate balanced performance in both visual and text-only question answering. By providing a single image, our RAP-LLaVA surpasses baseline methods and achieves performance comparable to that of GPT-4V.

\noindent \textbf{Visual Recognition.}
We also evaluate the models' recognition abilities for a more comprehensive comparison. MLLMs are required to determine whether a personal concept is present in an image. We query them with "Is \textlangle sks\textrangle\ in the image? Answer with a single word.”, where \textlangle sks\textrangle\ is replaced by corresponding concept name. For positive images, the desired response is "Yes" and "No" for negative. Results show that without knowledge of personal concepts, the vanilla LLaVA consistently produces negative responses. After training on target concepts, LLaVA-LoRA, MyVLM and YoLLaVA tend to give positive responses, but struggle to differentiate between similar concepts, resulting in weaker performance on negative images. Our models demonstrate exceptional performance in both positive and negative scenarios, achieving the best overall results.

% \subsection{Cost of Personalization}
% \label{sec:exp:cost}
% We further compare the costs of personalization. As shown in Table~\ref{Cost}, existing methods usually struggle with continuous updates or have high demands for training data. For fine-tuning based method like LLaVA-LoRA, while they can achieve satisfactory performance, fine-tuning the model each time a new concept emerges incurs substantial computational costs. MyVLM and Yo'LLaVA learn an embedding or some new tokens to represent the new concept without updating the pre-trained MLLM's parameters, however, they require multiple labeled images of the target concept and a large number of negative images, which poses significant challenges for data collection. In contrast, our RAP requires only 1 image with its related information provided by the user, achieving outstanding performance across various personalized generation tasks. At the same time, by modifying images and descriptions in the database, RAP enables real-time editing of personalized generation settings. We present examples of real-time concept editing in Table~\ref{appendix:editing}.

\subsection{Ablation Study}
\label{sec:exp:ablation}

\noindent \textbf{Influence of Number of Learned Concepts.} In real-world scenario, users' personal databases typically expand over time. Next, we evaluate the performance of various methods with varying numbers of learned concepts. We extend the database with hundreds of new concepts selected from the RefCOCO dataset~\cite{refcoco}, ensuring no overlap with the test dataset. 
For LLaVA-LoRA and MyVLM, we provide images containing the target concepts along with their captions as training data, and assess models' performance on the original test dataset. The results are presented in Figure~\ref{fig:varying_num_concepts}. More learned concepts result in increased recognition errors, leading to a decline in performance for each model. Our RAP-MLLMs maintain the highest performance under different settings.

\begin{table}[t]
\caption{We evaluate model's performance with perfect retrieval, and test contributions of text information and dataset components.}
\label{ablation}
\vspace{-3mm}
\setlength{\tabcolsep}{1mm}{\begin{tabular}{l|lll}
\toprule
Setting & Recall & Precision & F1-score \\
\midrule
\rowcolor{Gray}RAP-LLaVA &93.51 &96.47 & 94.97\\
\textit{Skip retrieval} & 96.16 {\scriptsize(\textcolor{Green}{+2.7})}   &100.0 {\scriptsize(\textcolor{Green}{+3.5})} & 98.04 {\scriptsize(\textcolor{Green}{+3.1})} \\
- \textit{Text information}  & 94.91 {\scriptsize(\textcolor{Green}{+1.4})} & 88.66 {\scriptsize(\textcolor{Red}{-7.8})} & 91.68 {\scriptsize(\textcolor{Red}{-3.3})}\\
- \textit{Data augmentation}  & 89.25 {\scriptsize(\textcolor{Red}{-4.3})} & 98.01 {\scriptsize(\textcolor{Green}{+1.5})} & 93.42 {\scriptsize(\textcolor{Red}{-1.6})}\\
- \textit{Negative samples}  & 95.74 {\scriptsize(\textcolor{Green}{+2.2})} & 58.21 {\scriptsize(\textcolor{Red}{-38.3})} & 72.40 {\scriptsize(\textcolor{Red}{-22.6})} \\ 
\bottomrule
\end{tabular}}
\vspace{-2mm}
\end{table}

\noindent \textbf{Generation Ability of MLLM.} We skip the recognition and retrieval processes, providing the MLLM with relevant information of each concept present in the image to evaluate the generation capability of the trained MLLM. The results, shown in Table~\ref{ablation}, indicate that when relevant concept information is supplied, our RAP-LLaVA achieves superior generation performance, obtaining 100\% precision without outputting irrelevant concepts as well as a higher recall rate.

\noindent \textbf{Text Information.}
We remove additional text information to examine its impact on personalization. The results are shown in Table~\ref{ablation} and \ref{appendix:QA_ablation} (Appendix). This additional text provides extra information that helps the model understand specific concepts, effectively improving the accuracy of the answers and benefiting personalized generation. Furthermore, it is especially helpful for question answering, as it can provide information not directly available from the image, such as a user's age and hobbies, or a pet's preferred food, as shown in the example in Table~\ref{appendix:editing} of the Appendix.

\noindent \textbf{Dataset Composition.}
We conduct experiments to assess the contribution of each component in our dataset. First, we remove data generated through data augmentation and train the model. The results displayed in Table~\ref{ablation} indicate an obvious decrease in the recall metric for image captioning, resulting in a lower overall performance. We further exclude constructed negative samples from the dataset and retrain the model, then we find that it performs poorly on precision metric. This suggests a diminished ability to discriminate against noise concepts not present in the image.

\noindent \textbf{Additional Ablation.} We  conduct ablation studies on \textbf{retriever's performance}, and examine the impact of \textbf{retrieving different numbers of concepts}. We also evaluate models' performance on \textbf{several multimodal benchmarks}, and the results demonstrate that RAP-LLaVA retains most general knowledge of the original LLaVA, while achieving superior performance in knowledge intensive tasks. Due to space limit, we put the results in Appendix Section~\ref{appendix:ablation}.

\section{Conclusion}
In this paper, we introduce the RAP framework for personalizing MLLMs. This framework enables MLLMs to understand infinite user-specific concepts, generate personalized captions and respond to user-related queries. To enhance the quality of the generated content and better align outputs with user configuration, we curate a large-scale dataset for personalized training of MLLMs and train a series of MLLMs to function as personalized assistants. Experimental results show that RAP-MLLMs achieve exceptional performance in various personalized generation tasks, while allowing real-time adjustments to generation settings.

\noindent\textbf{Acknowledgements.} This work is partially supported by the National Natural Science Foundation of China (Grant No. 62306261), and The Shun Hing Institute of Advanced Engineering (SHIAE) Grant (No. 8115074).

{
    \small
    \bibliographystyle{ieeenat_fullname}
    \bibliography{main}
}

\clearpage
\clearpage
\maketitlesupplementary
\appendix

\section{Appendix Overview}
\begin{itemize}[leftmargin=18 pt, itemsep= 3 pt,topsep = 1pt]
\item Section \ref{appendix:eval}: Additional evaluations of our models.
\item Section \ref{appendix:expetiment}: More experimental details.
\item Section \ref{appendix:dataset}: More details of the RAP dataset.
\item Section \ref{appendix:demo}: Additional demonstrations.
\item Section \ref{appendix:limitation}: Analysis of limitations of our work.
\item Section \ref{appendix:ethics}: Analysis of potential ethics issues.
\item Section \ref{appendix:database}: Examples of the personalized database.
\end{itemize}

\section{Additional Evaluation Results}
\label{appendix:eval}

\subsection{Ablation Study}
\label{appendix:ablation}

\noindent \textbf{Dataset Composition.} We conduct ablation experiments on the question answering and recognition benchmark, experimental results are present in Table~\ref{appendix:QA_ablation}. The results further demonstrate that our data augmentation and the constructed negative samples also contribute to the model's performance.

% \begin{table}[H]
% \centering
% \caption{Image captioning performance without text information.}
% \label{appendix:ablation:caption}
% \vspace{-3mm}
% \setlength{\tabcolsep}{1.3mm}{\begin{tabular}{l|lll}
% \toprule
% Setting & Recall & Precision & F1-score \\
% \midrule
% \rowcolor{Gray}RAP-LLaVA &93.51 &96.47 & 94.97\\
% - \textit{Text information}  & 94.91 {\scriptsize(\textcolor{Green}{+1.4})} & 88.66 {\scriptsize(\textcolor{Red}{-7.8})} & 91.68 {\scriptsize(\textcolor{Red}{-3.3})}\\
% \bottomrule
% \end{tabular}}
% \end{table}

% \noindent \textbf{Text information.}
% We remove additional text information to examine its impact on personalization. The results are shown in Table~\ref{appendix:ablation:caption} and \ref{appendix:QA_ablation}. This additional text provides extra information that helps the model understand specific concepts, effectively improving the accuracy of the answers and benefiting personalized generation. Furthermore, it is especially helpful for question answering, as it can provide information not directly obtainable from the image, such as a user's age and hobbies, or a pet's preferred food, as illustrated in the example in Table~\ref{appendix:editing}.

% \begin{table}[H]
% \centering
% {\includegraphics[width=5.5cm]{figure/retriever.pdf}}
% \vspace{-3mm}
% \captionof{figure}{\textbf{Performance of Our Retriever}. Top-K recall rates under varying database size N.}
% \label{fig:retriever}
% \end{table}

\noindent\resizebox{1\linewidth}{!}{%
\centering
\begin{minipage}[t]{.5\textwidth}
\vspace{-4mm}
\centering
{\includegraphics{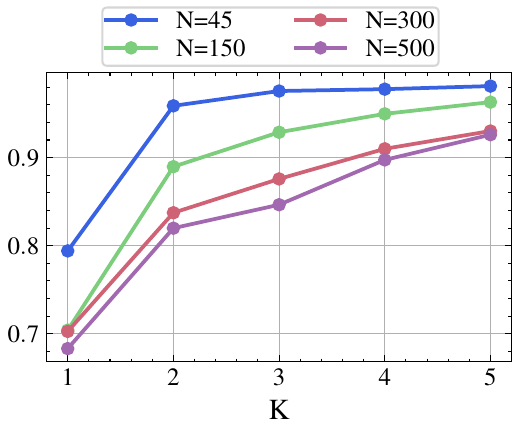}}
\captionsetup{font=LARGE}
\vspace{-8mm}
\captionof{figure}{\textbf{Retriever's Top-K Recall} under varying database size N.}
\label{fig:retriever-recall}
\end{minipage}
\hspace{1mm}
\begin{minipage}[t]{.5\textwidth}
\vspace{-4mm}
\centering
{\includegraphics{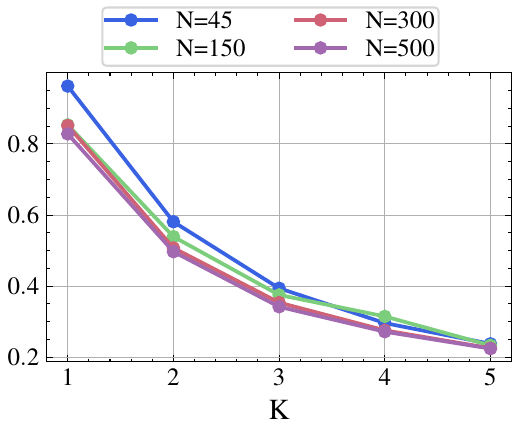}}
\captionsetup{font=LARGE}
\vspace{-8mm}
\captionof{figure}{\textbf{Retriever's  Top-K Precision} under varying database size N.}
\label{fig:retriever-precision}
\end{minipage}
}
\vspace{1mm}

\noindent \textbf{Retriever.} The retriever's performance is crucial for a RAG system. We assess the retriever's performance on the personalized captioning dataset. We use the detection model to identify potential concepts and retrieve the K concepts with the highest similarity from the database. Figure~\ref{fig:retriever-recall} and Figure~\ref{fig:retriever-precision} show the Top-K recall and precision for different values of K and database sizes N. The results indicate that as the database size increases, the retriever's performance declines. While a larger K generally enhances recall, it also introduces more irrelevant concepts, leading to a drop in precision. Notably, even with 500 personal concepts to remember, the Top-5 recall rate can still exceed 90\%, which guarantees the effectiveness of our RAP framework.

\begin{table}[H]
\centering
{\includegraphics[width=5cm]{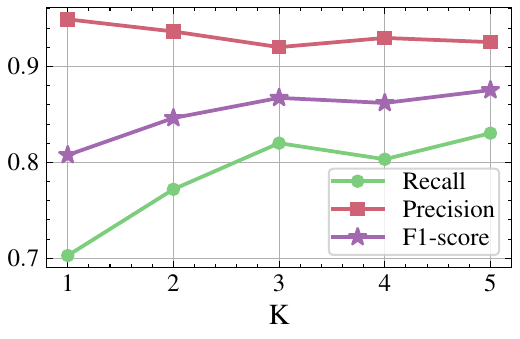}}
\vspace{-3mm}
\captionof{figure}{\textbf{Captioning Performance} with varying number of retrieved concepts.}
\label{fig:caption-K}
\end{table}

\noindent \textbf{Impact of Retrieving Different Numbers of Concepts.}
We train a new model, RAP-LLaVA-OneVision (0.5B) \cite{llavaonevision}, to analyze the impact of the number of retrieved concepts in detail. As shown in Figure \ref{fig:caption-K}, retrieving more concepts from a database of 300 concepts improves the recall metric of image captioning, but also introduces more noise, leading to a drop in precision. This reflects the trade-off between supporting a larger set of personalized concepts and maintaining generation precision.

\begin{table}[H]
\centering
\caption{\textbf{Evaluation on Knowledge-intensive Multimodal Benchmarks}. KB: Knowledge Base.}
\label{tab:addition}
\vspace{-3mm}
\setlength{\tabcolsep}{1mm}{\begin{tabular}{l|ccc}
\toprule
Method  & MMMU~\cite{mmmu}  & InfoSeek~\cite{infoseek}\\
\midrule
LLaVA~\cite{improvedbaselinesvisualinstruction}  & 0.364 & 0.205 \\
LLaVA-LoRA~\cite{lora}  & 0.359 &0.205 \\
\rowcolor{Gray}RAP-LLaVA & 0.361 & 0.218\\
\rowcolor{Gray}RAP-LLaVA(With KB) & \textbf{0.369} & \textbf{0.344}  \\
\bottomrule
\end{tabular}}
\vspace{-3mm}
\end{table}

\begin{table*}[th]
\centering
\caption{Ablation studies on \textbf{Question Answering and Visual Recognition.} Weighted results are computed as arithmetic means.}
\label{appendix:QA_ablation}
\vspace{-3mm}
\setlength{\tabcolsep}{2mm}{\begin{tabular}{l|lll|lll}
\toprule
\multirow{2}{*}{Method}  & \multicolumn{3}{c|}{Question Answering} &\multicolumn{3}{c}{Visual Recognition} \\
&Visual & Text &Weighted& Positive & Negative & Weighted\\
\midrule
\rowcolor{Gray}RAP-LLaVA & 0.935 & 0.938 & 0.936 & 0.979 & 0.982 & 0.980\\
- \textit{Text information} & 0.935 & 0.908 {\scriptsize(\textcolor{Red}{-0.030})} & 0.921 {\scriptsize(\textcolor{Red}{-0.015})} & 0.988 {\scriptsize(\textcolor{Green}{+0.009})} & 0.930 {\scriptsize(\textcolor{Red}{-0.052})} & 0.959 {\scriptsize(\textcolor{Red}{-0.021})}\\
- \textit{Data augmentation}  & 0.924 {\scriptsize(\textcolor{Red}{-0.011})} & 0.918 {\scriptsize(\textcolor{Red}{-0.020})} & 0.921 {\scriptsize(\textcolor{Red}{-0.015})} & 0.943 {\scriptsize(\textcolor{Red}{-0.036})} & 0.988 {\scriptsize(\textcolor{Green}{+0.006})} & 0.965 {\scriptsize(\textcolor{Red}{-0.015})}\\
- \textit{Negative samples} & 0.918 {\scriptsize(\textcolor{Red}{-0.017})} & 0.933 {\scriptsize(\textcolor{Red}{-0.005})} & 0.925 {\scriptsize(\textcolor{Red}{-0.011})} & 0.958 {\scriptsize(\textcolor{Red}{-0.021})} & 0.985 {\scriptsize(\textcolor{Green}{+0.003})} & 0.971 {\scriptsize(\textcolor{Red}{-0.009})}\\
\bottomrule
\end{tabular}}
\end{table*}

\noindent \textbf{Multimodal Benchmark.}
We also evaluate our model's performance on several traditional multimodal benchmarks, including MMMU \cite{mmmu} and InfoSeek \cite{infoseek}. We assess our models' performance both with and without external knowledge base. For MMMU~\cite{mmmu}, we use 30K images paired with corresponding captions from Wikipedia as the external knowledge base. During testing, we retrieve the three most similar images based on the question's image and incorporate only the textual knowledge to the input. For InfoSeek~\cite{infoseek}, we randomly sample 5K questions from the validation set and construct a knowledge base containing 50K entities from Wikipedia database provided by the authors, which includes all relevant entities associated with the questions. For each question, we retrieve the most similar entity and add only the textual knowledge to the input.

We evaluate on the validation set of MMMU, and 5K questions sampled from the validation set of InfoSeek. We use the official scripts to get the results, which are presented in Table \ref{tab:addition}. From the results, our RAP-LLaVA retains most general knowledge of the original LLaVA~\cite{improvedbaselinesvisualinstruction}. It also equips the MLLM with the ability to retrieve information from an external knowledge base, demonstrating superior performance in knowledge intensive tasks.

\begin{table*}[t]
\small
\centering
\setlength{\tabcolsep}{9pt}
\caption{\textbf{Quantitative Evaluation with Standard Image Captioning Metrics}. We compute standard image captioning metrics for personalized captions generated by RAP-LLaVA. For each image, we treat all 5 augmented captions as the set of ground truth captions. The results are averaged across all three evaluations and all concepts.}
\vspace{-3mm}
\centering
\begin{tabular}{l c c c c c c c c}
    \toprule
                 Model & B1  & B2  &  B3 &  B4 & METEOR & ROUGE\_L &  CIDEr & SPICE \\
    \midrule                            LLaVA~\cite{improvedbaselinesvisualinstruction} & $0.177$ & $0.100$ & $0.057$ & $0.032$ & $0.138$ & $0.240$	& $0.428$ & $0.136$ \\
      LLaVA~\cite{improvedbaselinesvisualinstruction} + Retriever & $0.087$ & $0.038$ & $0.014$ & $0.005$ & $0.074$ & $0.132$	& $0.025$ & $0.020$ \\
      MyVLM-LLaVA~\cite{myvlm} & $0.195$ & $0.103$ & $0.061$ & $0.036$ & $0.213$ & $0.263$ & $0.435$ & $0.145$ \\
      RAP-LLaVA  & $0.178$ & $0.094$ & $0.058$ & $0.037$ & $0.215$ & $0.189$ & $0.388$ & $0.108$ \\
    \bottomrule
\end{tabular}
\label{tab:captioning_metrics}
\end{table*}

\subsection{Standard Image Captioning Metrics}
We further evaluate using standard image captioning metrics. The results are shown in Table~\ref{tab:captioning_metrics}. Despite the additional concept information is offered, LLaVA’s performance declines due to its inability to effectively utilize the information. Although our RAP-LLaVA is not trained on these concepts, it still achieves comparable performance on general image captioning tasks while enabling personalization.

\subsection{Robustness to Retrieval Errors}
As described in Section~\ref{dataset}, we construct negative samples to enhance models' robustness to retrieval errors. Figure~\ref{fig:retriever-recall} and \ref{fig:retriever-precision} show the retriever's recall and precision under different database sizes. As the database size grows, the retriever's recall and precision decrease, which means a higher likelihood of retrieving irrelevant concepts. Despite this, our models maintain outstanding performance, as shown in Figure~\ref{fig:varying_num_concepts}, which demonstrates their robustness to retrieval errors. 
Additionally, results in Table \ref{ablation} show that our models perform even better when the retrieval process is error-free. In Table \ref{appendix:ret_error}, we provide examples to show model's robustness to retrieval errors. Thanks to negative samples in the training dataset, RAP-MLLMs can distinguish irrelevant concepts and generate accurate responses.

\begin{table}[H]
\centering
{\includegraphics[width=5.5cm]{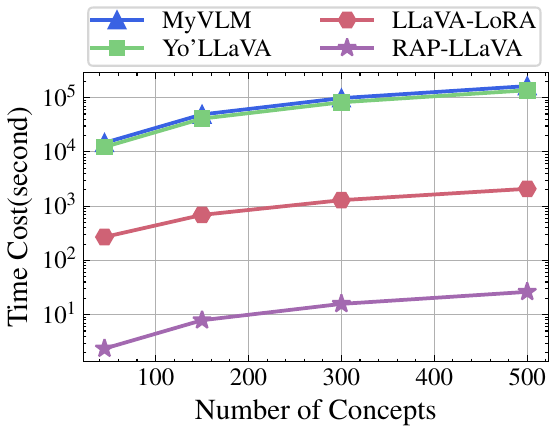}}
\vspace{-3mm}
\captionof{figure}{\textbf{Time Cost of Personalization}. We conduct experiment with 2 A800 GPUs.}
\label{fig:time_cost}
\end{table}
\subsection{Time Cost of Personalization}
We also evaluate the time cost associated with different methods for learning a set of user's concepts. The results are presented in Figure~\ref{fig:time_cost}. MyVLM has to train an external recognition model for each concept and learn an embedding to adjust the model's outputs. Similarly, Yo'LLaVA needs to learn new tokens for each concept. During the optimization process, both approaches necessitate multiple forward and backward pass of the MLLM, resulting in significant time consumption. In contrast, our RAP only requires time for encoding the image and adding its embedding to the database, which can be accomplished in just a few seconds. This significantly enhances the convenience and practicality of our models in practical applications.

\section{More Experimental Details}
\label{appendix:expetiment}

\textbf{Implementation Details.} We utilize YOLO-Worldv2-X~\cite{yoloworld} as the detection model, setting detection classes to include all categories stored in the database to reduce the interventions from unrelated objects. We construct a multimodal retriever using Facebook AI Similarity Search (FAISS) \cite{faiss}, employing a pre-trained CLIP ViT-L/14-336 \cite{clip} as the visual encoder. Each key in the database is generated by inputting a concept's image into the CLIP visual encoder, resulting in a 768-dimensional vector. Considering the restriction of context length of the backbone language model, we retrieve the 2 most similar images from the database for each region of interest. Then we select 2 and 3 different concepts with the highest similarity among all as supplementary inputs for RAP-LLaVA and RAP-Phi3-V, respectively.

\noindent\textbf{Baselines.}
For MyVLM, we find that when the training data is very limited, it is quite hard for the classification head to work effectively. Therefore, we use data augmentation to help improve its performance. Specifically, we crop the single image into several pieces containing the target concept to improve the accuracy of classification heads. To distinguish between multiple different concepts that may appear in the image, we use \textlangle sks1\textrangle, \textlangle sks2\textrangle $\dots$ as concept identifiers. For YoLLaVA, we present its experimental results reported in the original paper~\cite{yollava}. For GPT-4V, reference images and descriptions are provided as additional prompts, thus it can generate responses about the target concept.

\noindent\textbf{Multi-concept Data Collection.}
We collect videos from YouTube and sample frames from them. For each video, we detect multiple concepts and use Gemini \cite{gemini} to find frames that contain both concepts in a given pair. For each pair, there are 8 to 13 images used for testing. Examples of these concept pairs are shown in Table \ref{tab:Qualitative Comparison}. We generate five captions for each image to enhance the concept learning of baseline methods.

\begin{table*}[ht]
\begin{minipage}[t]{.6\textwidth}
\centering
\captionof{figure}{Composition of our dataset.}
\label{fig:composition}
\centering
\vspace{-3mm}
{\includegraphics[width=6.5cm]{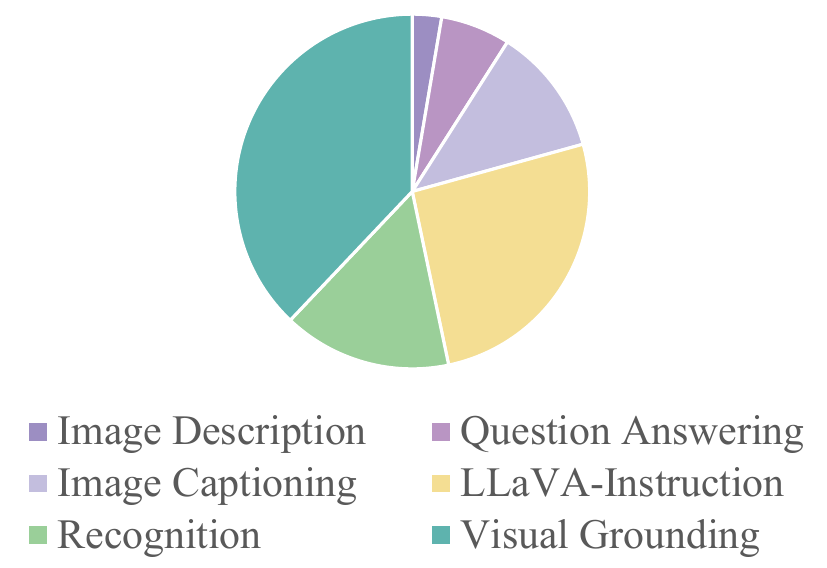}}
\end{minipage}\hspace {1mm}
\begin{minipage}[t]{.3\textwidth}
\centering
\caption{Statistics of our dataset.}
\label{tab:dataset_statistics}
\vspace{-3mm}
\setlength{\tabcolsep}{3mm}{
\begin{tabular}{l|c}
\toprule
Type  & Size\\
\midrule
Visual Grounding &100K\\
Recognition  &40K\\
Caption \& Description& 37K\\
Question Answering & 16K\\
LLaVA-Instruction & 67K\\
\midrule
\textbf{Total} &\textbf{260K}\\
\bottomrule
\end{tabular}
}
\end{minipage}
\end{table*}

\begin{table}[ht]
\centering
\caption{Data source.}
\label{tab:dataset_source}
\vspace{-3mm}
\setlength{\tabcolsep}{1mm}{\small{\begin{tabular}{l|c}
\toprule
Type  & Source Dataset\\
\midrule
\multirow{2}{*}{Visual Grounding} &RefCOCO~\cite{refcoco}, TAO~\cite{TAO}\\
& ILSVRC2015-VID~\cite{imagenet}, Object365~\cite{object365}\\
\midrule
Recognition  &CustomConcept101~\cite{custom_diffusion}, CelebA~\cite{celebA}\\
\midrule
\multirow{2}{*}{Caption \& Description} &RefCOCO~\cite{refcoco}, TAO~\cite{TAO}\\
& Object365 \cite{object365}, CustomConcept101~\cite{custom_diffusion}\\
\midrule
\multirow{3}{*}{Question Answering} & RefCOCO~\cite{refcoco}, TAO~\cite{TAO}\\
& Object365 \cite{object365}, CustomConcept101~\cite{custom_diffusion}\\
& CelebA~\cite{celebA}\\
\midrule
LLaVA-Instruction& LLaVA-Instruct-665K~\cite{improvedbaselinesvisualinstruction}\\
\bottomrule
\end{tabular}}}
\end{table}

\section{Details of Dataset}
\label{appendix:dataset}

\subsection{Dataset Composition}
\label{appendix:dataset composition}
\begin{itemize}[leftmargin=18 pt, itemsep= 3 pt,topsep = 1pt]
\item We provide a summary of the composition of our dataset in Figure~\ref{fig:composition}, which visually represents the distribution of different components.
\item Table~\ref{tab:dataset_statistics} presents detailed numerical data for each part.
\item In Table~\ref{tab:dataset_source}, we specify the sources for each component of our dataset.
\end{itemize}

\subsection{Instructions}
\label{appendix:dataset instructions}
In this section, we present the instruction templates used to create our dataset:
\begin{itemize}[leftmargin=18 pt, itemsep= 3 pt,topsep = 1pt]
\item Table \ref{tab:inst_for_grounding} contains instructions for visual grounding and recognition.
\item Table \ref{tab:inst_caption} includes example instructions for image captioning.
\item Table \ref{tab:inst_description} presents example instructions for image description.
\item Table \ref{tab:seed_q} presents example questions used for question answering synthesis.
\end{itemize}

\section{Additional Demonstrations}
\label{appendix:demo}
In this section, we provide more qualitative results obtained by various models. 

\begin{itemize}[leftmargin=18 pt, itemsep= 3 pt,topsep = 1pt]
\item In Table~\ref{appendix:editing}, we demonstrate how our models achieve real-time editing of concepts by modifying the database.
\item In Table~\ref{appendix:updating}, we demonstrate the real-time addition of new concepts by updating the database.
\item In Table~\ref{tab:additional QA}, we present qualitative results on personalized conversation of RAP-LLaVA.
\item In Table~\ref{tab:additional QA2}, we present qualitative results on personalized conversation of RAP-Phi3-V.
\item In Table~\ref{tab:Additional Qualitative Comparison1}, we present additional image captions generated by RAP-LLaVA and other methods. 
\item In Table~\ref{tab:Additional Qualitative Comparison2}, we present additional image captions generated by RAP-Phi3-V and other methods. 
\item In Table~\ref{tab:Additional description1}, we provide demonstrations of image description generated by RAP-LLaVA and LLaVA. 
\item In Table~\ref{tab:Additional description2}, we provide demonstrations of image description generated by RAP-Phi3-V and Phi3-V. 
\item In Table~\ref{tab:additional recognition} and \ref{tab:additional recognition2}, we provide results on visual recognition of RAP-LLaVA. It also has the ability to give precise bounding box of specific concept in the image.
\end{itemize}

% real-time editing
\begin{table}[t]
\centering
\caption{\textbf{Examples of Robustness to Retrieval Errors}. Our RAP-MLLMs can distinguish noisy concepts and generate accurate answers.}
\label{appendix:ret_error}
\vspace{-3mm}
\begin{tabular}{cc}
\toprule
Image& Retrieved Concepts\\
\midrule
\makecell[c]{
\begin{minipage}[b]{0.4\linewidth}
    \centering
    {\includegraphics[width=3.5cm]{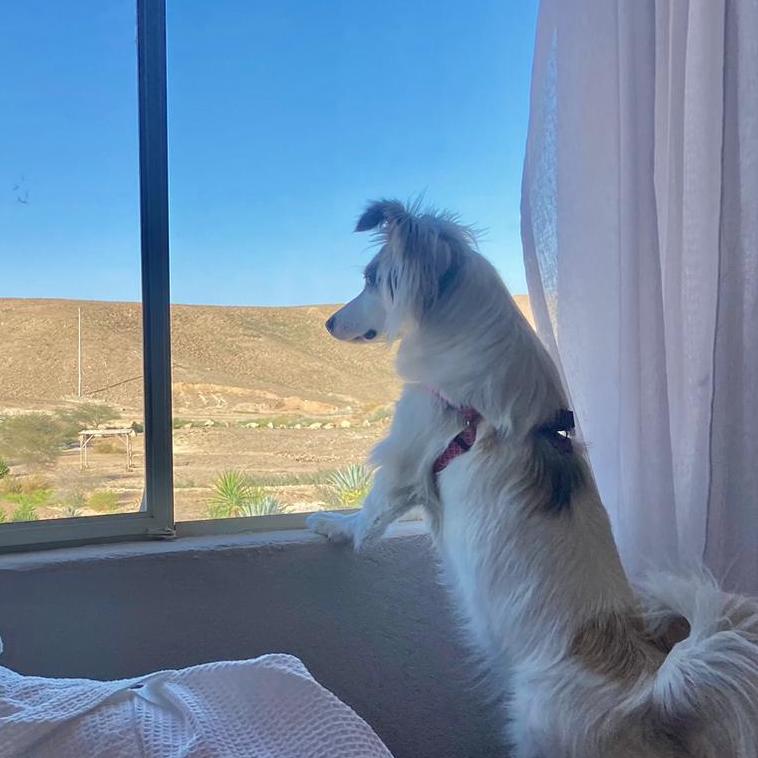}}
\end{minipage}
}
& \makecell[c]{
\begin{minipage}[b]{0.4\linewidth}
    \centering
    {\includegraphics[width=3.7cm]{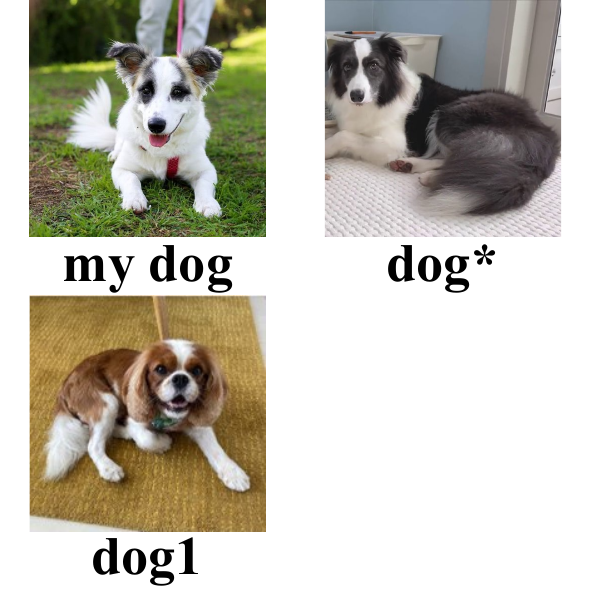}}
\end{minipage}
} \\
\midrule
\multicolumn{2}{l}{\makecell[c]{
\begin{minipage}[t]{1\linewidth} % 使用 [t] 使 minipage 顶部对齐
\raggedright % 设置内部文字左对齐
Question: Give a caption of this image.\\
\adjustbox{bgcolor=Gray,minipage=[t]{\linewidth}}{\textbf{RAP-Phi3-V}: \textcolor{cgreen}{\textlangle my dog\textrangle} is looking out the window.}\vspace{1mm}\\
Question: Can you see \textlangle dog*\textrangle\ in this image?\\
\adjustbox{bgcolor=Gray,minipage=[t]{\linewidth}}{\textbf{RAP-Phi3-V}: 
No, \textlangle dog*\textrangle\ is not visible in this picture.}
\end{minipage}
}}\\
\bottomrule
\end{tabular}
\end{table}

% real-time editing
\begin{table*}[t]
\centering
\caption{\textbf{Examples of Concept Editing}. Based on the information recorded in the database, our RAP-LLaVA can provide reliable and accurate answers.}
\label{appendix:editing}
\vspace{-3mm}
\begin{tabular}{ccc}
\toprule
Image& Concept & Information in the Database\\
\midrule
\makecell[c]{
\begin{minipage}[b]{0.2\linewidth}
    \centering
    {\includegraphics[width=1.5cm]{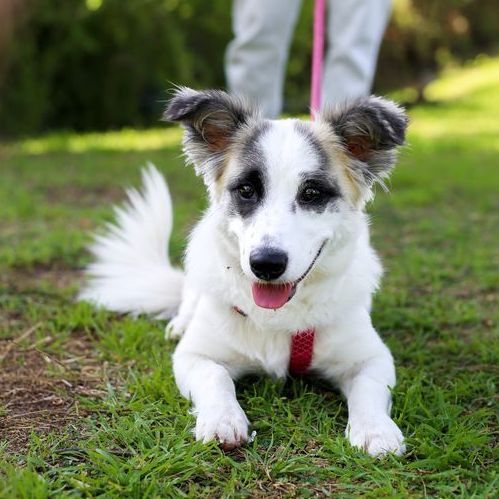}}
\end{minipage}
} &\makecell[c]{
\begin{minipage}[b]{0.12\linewidth}
    \centering
    {my dog}
\end{minipage}
}&\makecell[c]{
\begin{minipage}[b]{0.58\linewidth}
    {1. A white and gray dog with long fur. He has black eyes. His favorite food is \underline{chicken}.\\
     2. A white and gray dog with long fur. He has black eyes. His favorite food is \underline{beef}.}
\end{minipage}
} \\
\midrule
\multicolumn{2}{l}{\makecell[c]{
\begin{minipage}[b]{0.32\linewidth}
    \centering
    {\includegraphics[width=3.5cm]{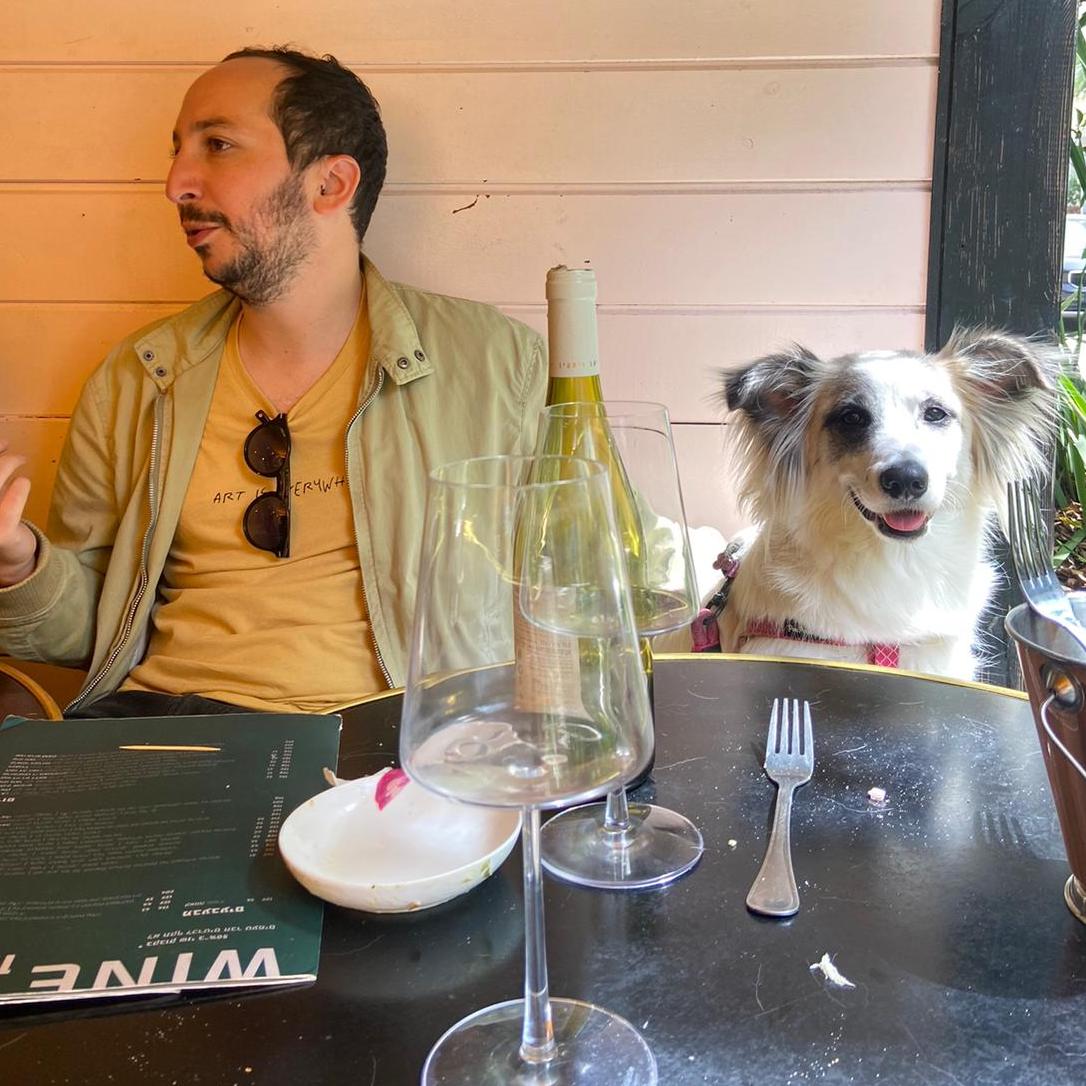}}
\end{minipage}
}}
& \makecell[c]{
\begin{minipage}[b]{0.61\linewidth}
    {Question: What is \textlangle my dog\textrangle's favorite food?\vspace{2mm}\\
    1. \adjustbox{bgcolor=Gray,minipage=[t]{\linewidth}}{\textbf{RAP-LLaVA}: \underline{Chicken}. }\vspace{2mm}\\
    2. \adjustbox{bgcolor=Gray,minipage=[t]{\linewidth}}{\textbf{RAP-LLaVA}: His favorite food is \underline{beef}. }
    }
\end{minipage}
} \\
\bottomrule
\end{tabular}
\end{table*}
\begin{table*}[t]
\centering
\caption{\textbf{Examples of Concept Updating}. The first caption is generated when toy2 not yet stored in the database. Once the new concept is added, RAP-LLaVA can recognize both toy1 and toy2.}
\label{appendix:updating}
\vspace{-3mm}
\begin{tabular}{ccc}
\toprule
Image& Concept & Information in the Database\\
\midrule
\makecell[c]{
\begin{minipage}[b]{0.2\linewidth}
    \centering
    {\includegraphics[width=1.5cm]{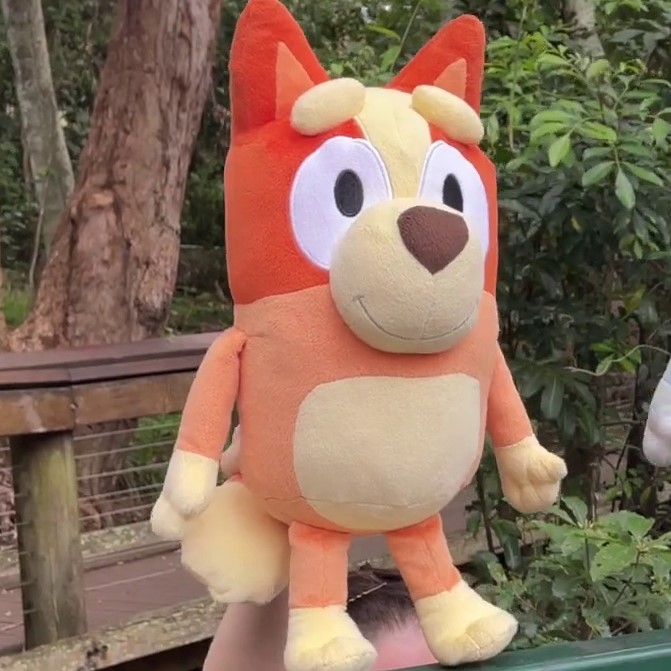}}
\end{minipage}
}&\makecell[c]{
\begin{minipage}[b]{0.12\linewidth}
    \centering
    {toy1}
\end{minipage}
} &\makecell[c]{
\begin{minipage}[b]{0.58\linewidth}
    {A plush toy. It is orange with a yellow belly and a brown nose.}
\end{minipage}
} \\
\midrule
\makecell[c]{
\begin{minipage}[b]{0.2\linewidth}
    \centering
    {\includegraphics[width=1.5cm]{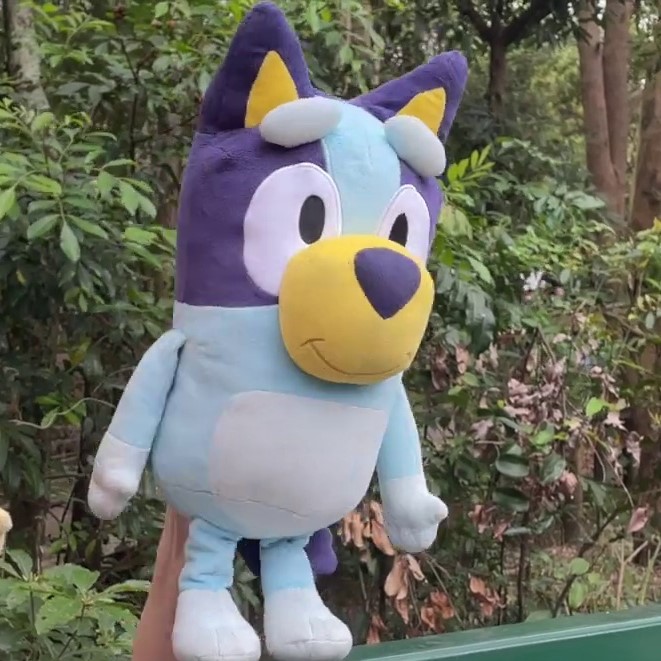}}
\end{minipage}
}&\makecell[c]{
\begin{minipage}[b]{0.12\linewidth}
    \centering
    {toy2}
\end{minipage}
} &\makecell[c]{
\begin{minipage}[b]{0.58\linewidth}
    {This is a plush toy of the bluey character. It is a light blue color with a purple patch on its head, and its ears are yellow. }
\end{minipage}
} \\
\midrule
\multicolumn{2}{l}{\makecell[c]{
\begin{minipage}[b]{0.32\linewidth}
    \centering
    {\includegraphics[width=3.5cm]{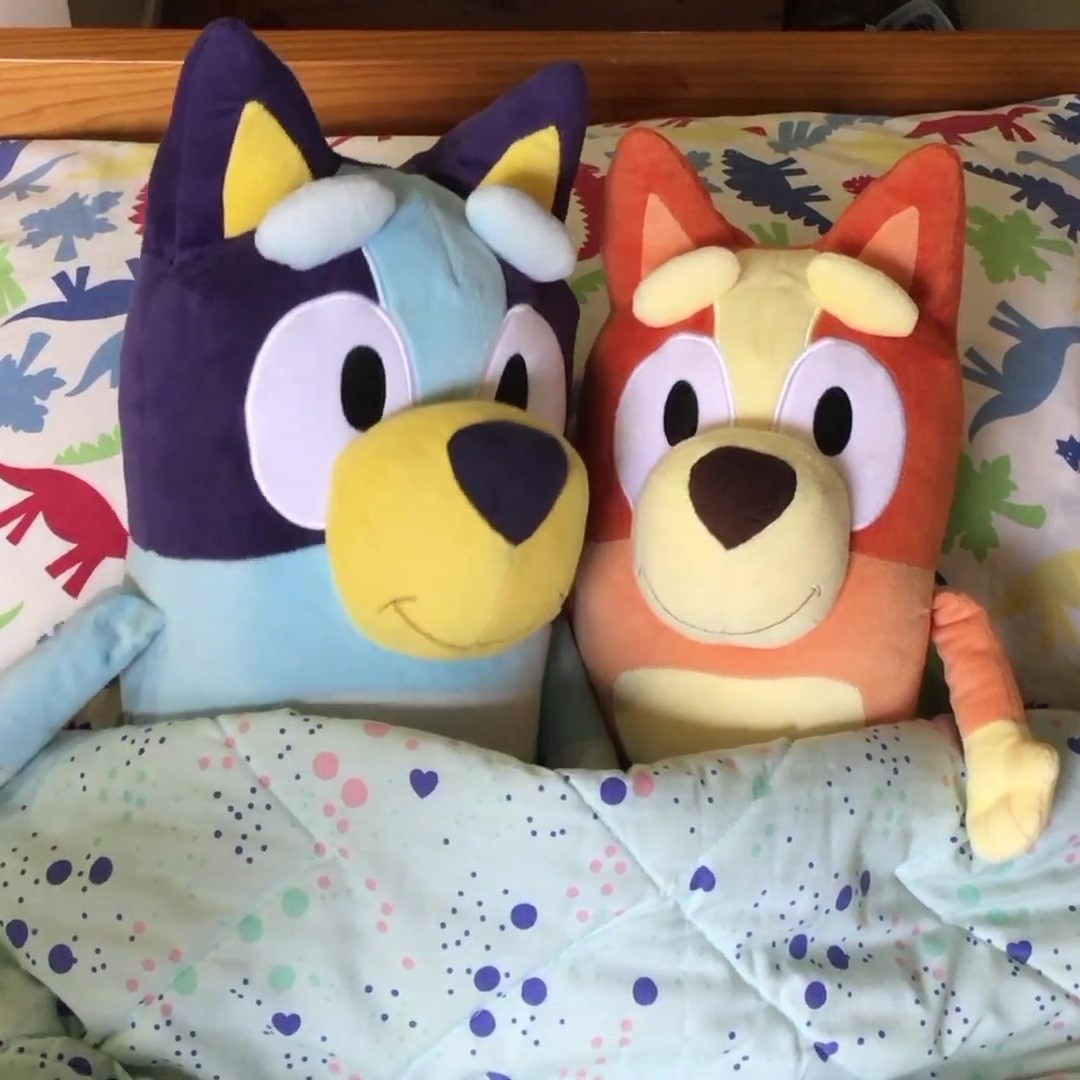}}
\end{minipage}
}}
& \makecell[c]{
\begin{minipage}[b]{0.61\linewidth}
    {Question: Give a caption of this image.\vspace{1.5mm}\\
    1. \adjustbox{bgcolor=Gray,minipage=[t]{\linewidth}}{\textbf{RAP-LLaVA}: \textcolor{cgreen}{\textlangle toy1\textrangle} is ready for bed! He's snuggled up with his friend, a blue and yellow dog plushie. They're both looking so cozy and ready for a good night's sleep! }\vspace{2mm}\\
    (Add new concept \textlangle toy2\textrangle\ to the database) \vspace{1mm}\\
    2. \adjustbox{bgcolor=Gray,minipage=[t]{\linewidth}}{\textbf{RAP-LLaVA}: \textcolor{cgreen}{\textlangle toy1\textrangle} and \textcolor{cgreen}{\textlangle toy2\textrangle} are ready for bed! They are both snuggled under the covers with their favorite stuffed animals. }
    }
\end{minipage}
} \\
\bottomrule
\end{tabular}
\end{table*}

% Question Answering
\begin{table*}[t]
\centering
\caption{Examples of personalized conversations obtained by RAP-LLaVA.}
\label{tab:additional QA}
\vspace{-3mm}
\begin{tabular}{ccc}
\toprule
Image& Concept & Information in the Database\\
\midrule
\makecell[c]{
\begin{minipage}[b]{0.2\linewidth}
    \centering
    {\includegraphics[width=1.5cm]{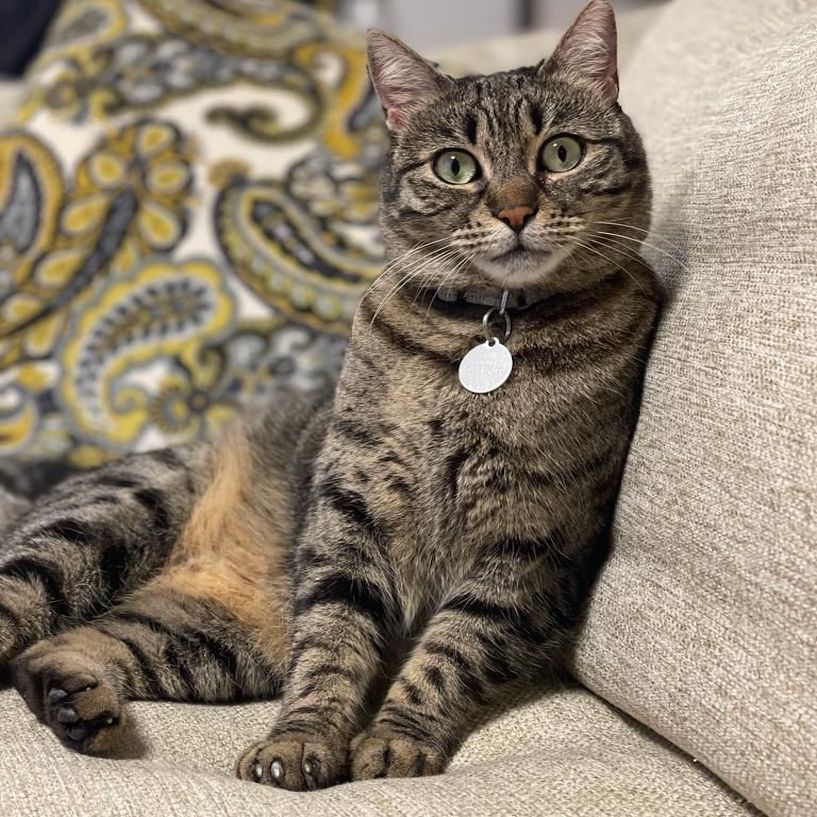}}
\end{minipage}
} &\makecell[c]{
\begin{minipage}[b]{0.12\linewidth}
    \centering
    {my cat}
\end{minipage}
}&\makecell[c]{
\begin{minipage}[b]{0.58\linewidth}
    {A tabby cat with its eyes closed is laying on a light grey surface. The cat's fur is a mix of light and dark brown stripes.}
\end{minipage}
} \\
\midrule
\multicolumn{2}{l}{\makecell[c]{
\begin{minipage}[b]{0.3\linewidth}
    \centering
    {\includegraphics[width=3.5cm]{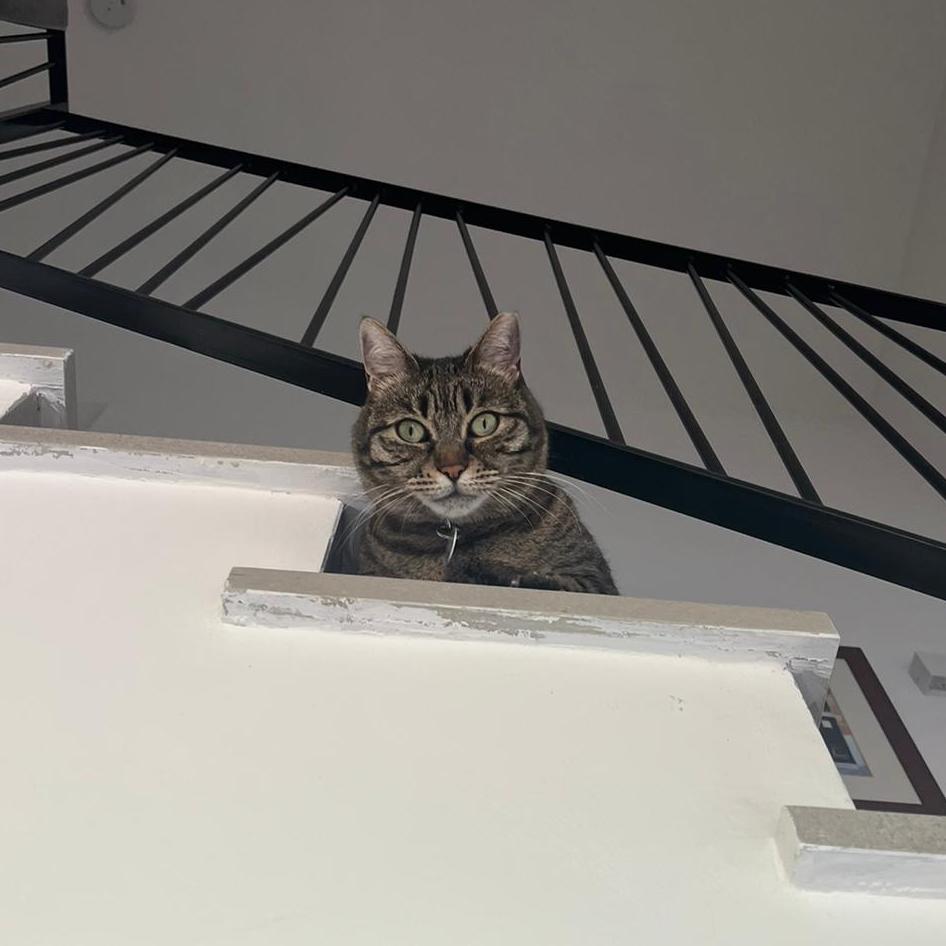}}
\end{minipage}
}}
& \makecell[c]{
\begin{minipage}[b]{0.62\linewidth}
    {
    Question: Where is \textlangle my cat\textrangle? \vspace{1mm}\\
    \adjustbox{bgcolor=Gray,minipage=[t]{\linewidth}}{\textbf{RAP-LLaVA}:
    \textlangle my cat\textrangle\ is on a staircase. }\vspace{2mm}\\
    Question: What is the type of \textlangle my cat\textrangle? \vspace{1mm}\\
    \adjustbox{bgcolor=Gray,minipage=[t]{\linewidth}}{\textbf{RAP-LLaVA}:
    \textlangle my cat\textrangle\ is a tabby cat. }\vspace{2mm}\\
    Question: What is the color of \textlangle my cat\textrangle? \vspace{1mm}\\
    \adjustbox{bgcolor=Gray,minipage=[t]{\linewidth}}{\textbf{RAP-LLaVA}:
    \textlangle my cat\textrangle\ is a tabby cat with a mix of light and dark brown stripes. }
    }
\end{minipage}
} \\
\midrule
Image& Concept & Information in the Database\\
\midrule
\makecell[c]{
\begin{minipage}[b]{0.2\linewidth}
    \centering
    {\includegraphics[width=1.5cm]{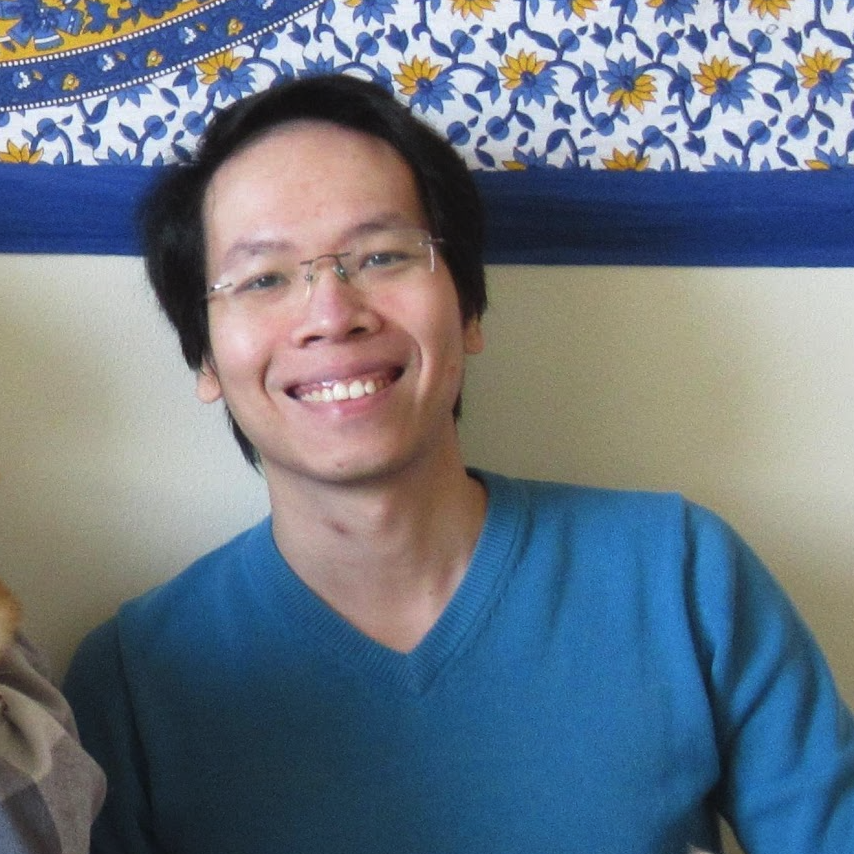}}
\end{minipage}
} &\makecell[c]{
\begin{minipage}[b]{0.12\linewidth}
    \centering
    {A}
\end{minipage}
}&\makecell[c]{
\begin{minipage}[b]{0.58\linewidth}
    {\textlangle A\textrangle\ is a smiling Asian man with short black hair, wearing glasses and a grey sweater.}
\end{minipage}
} \\
\midrule
\multicolumn{2}{l}{\makecell[c]{
\begin{minipage}[b]{0.3\linewidth}
    \centering
    {\includegraphics[width=3.5cm]{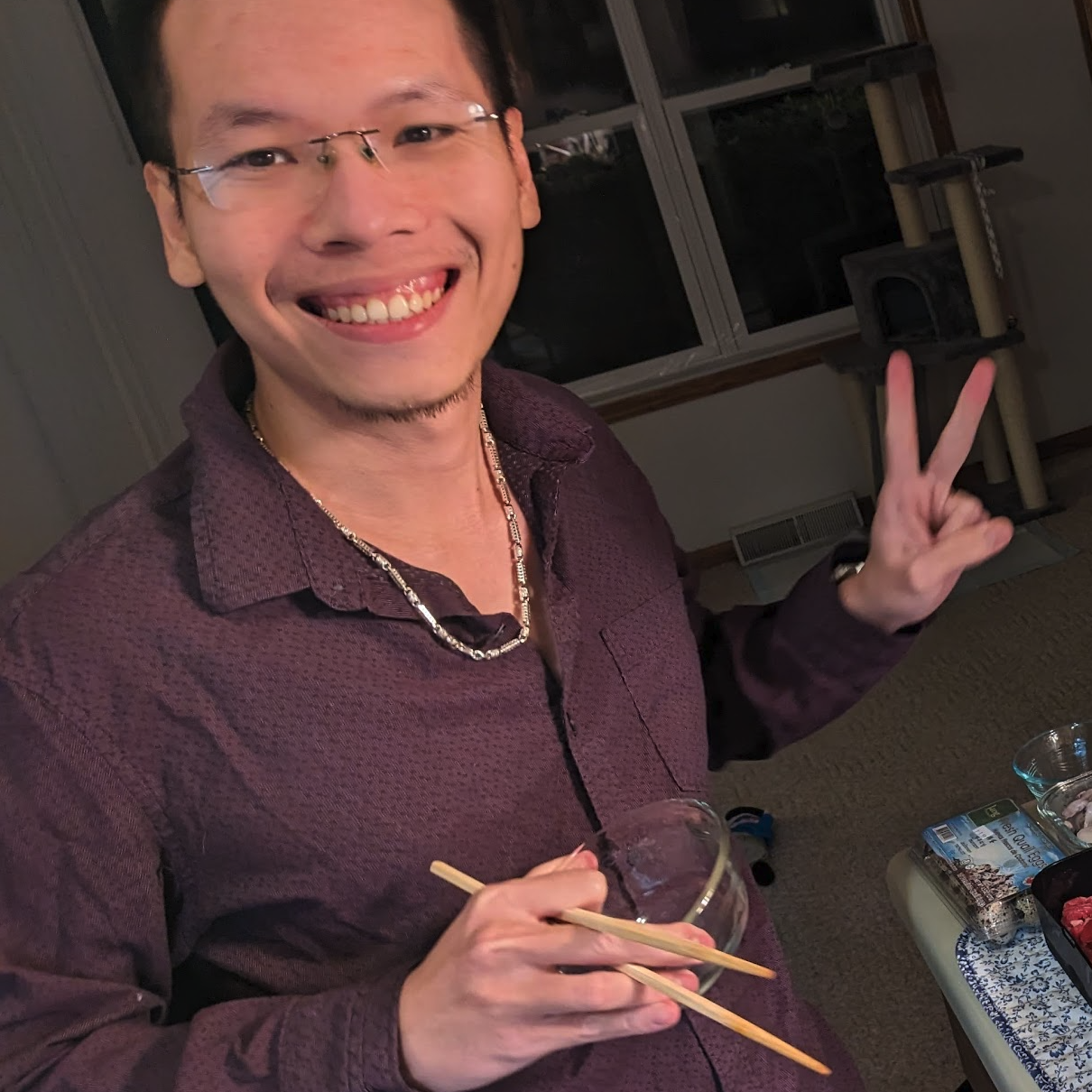}}
\end{minipage}
}}
& \makecell[c]{
\begin{minipage}[b]{0.62\linewidth}
    {
    Question: Can you see \textlangle A\textrangle\ in this photo? \vspace{1mm}\\
    \adjustbox{bgcolor=Gray,minipage=[t]{\linewidth}}{\textbf{RAP-LLaVA}: Yes,
    \textlangle A\textrangle\ is captured in this image.}\vspace{2mm}\\
    Question: What color is \textlangle A\textrangle’s shirt? \vspace{1mm}\\
    \adjustbox{bgcolor=Gray,minipage=[t]{\linewidth}}{\textbf{RAP-LLaVA}:
    \textlangle A\textrangle\ is wearing a purple shirt. }\vspace{2mm}\\
    Question: What is  \textlangle A\textrangle’s right hand holding? \vspace{1mm}\\
    \adjustbox{bgcolor=Gray,minipage=[t]{\linewidth}}{\textbf{RAP-LLaVA}:
    Chopsticks. }
    }
\end{minipage}
} \\
\midrule
Image& Concept & Information in the Database\\
\midrule
\makecell[c]{
\begin{minipage}[b]{0.2\linewidth}
    \centering
    {\includegraphics[width=1.5cm]{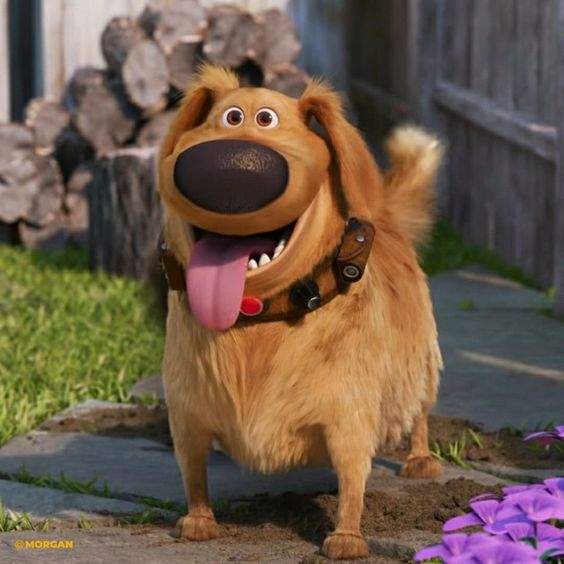}}
\end{minipage}
} &\makecell[c]{
\begin{minipage}[b]{0.12\linewidth}
    \centering
    {dug}
\end{minipage}
}&\makecell[c]{
\begin{minipage}[b]{0.58\linewidth}
    {\textlangle dug\textrangle\ is a golden retriever from the movie Up. He has a brown coat and a big, goofy smile, wearing a collar with some round objects.}
\end{minipage}
} \\
\midrule
\multicolumn{2}{l}{\makecell[c]{
\begin{minipage}[b]{0.3\linewidth}
\centering
(Text-only conversation)
\end{minipage}
}}
& \makecell[c]{
\begin{minipage}[b]{0.62\linewidth}
    {
    Question: Can you describe \textlangle dug\textrangle\ in detail? \vspace{1mm}\\
    \adjustbox{bgcolor=Gray,minipage=[t]{\linewidth}}{\textbf{RAP-LLaVA}:
    \textlangle dug\textrangle\ is a golden retriever from the movie Up. He has a brown coat and a big, goofy smile. He is wearing a collar with some round objects. He looks very happy and playful. }
    }
\end{minipage}
} \\
\midrule
Image& Concept & Information in the Database\\
\midrule
\makecell[c]{
\begin{minipage}[b]{0.2\linewidth}
    \centering
    {\includegraphics[width=1.5cm]{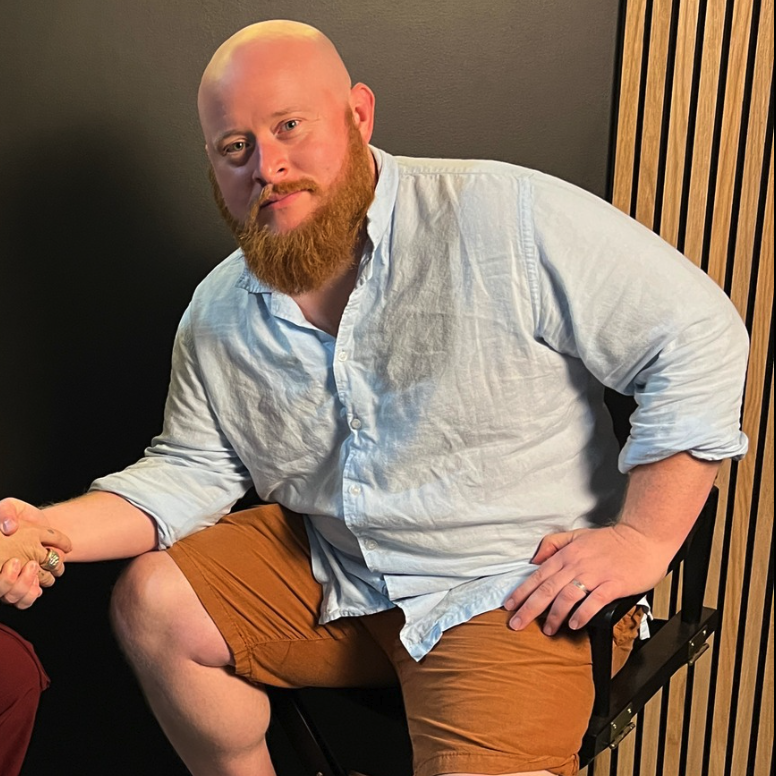}}
\end{minipage}
} &\makecell[c]{
\begin{minipage}[b]{0.12\linewidth}
    \centering
    {P}
\end{minipage}
}&\makecell[c]{
\begin{minipage}[b]{0.58\linewidth}
    {\textlangle P\textrangle\ is a bald man with a red beard.}
\end{minipage}
} \\
\midrule
\multicolumn{2}{l}{\makecell[c]{
\begin{minipage}[b]{0.3\linewidth}
\centering
(Text-only conversation)
\end{minipage}
}}
& \makecell[c]{
\begin{minipage}[b]{0.62\linewidth}
    {
    \vspace{2mm}Question: Can you describe \textlangle P\textrangle's appearence in detail? \vspace{1mm}\\
    \adjustbox{bgcolor=Gray,minipage=[t]{\linewidth}}{\textbf{RAP-LLaVA}:
    \textlangle P\textrangle\ is a bald man with a red beard. He is wearing orange shorts and a blue shirt. He is sitting in a black director's chair with his legs crossed.}
    }
\end{minipage}
} \\
\bottomrule
\end{tabular}
\end{table*}
\begin{table*}[t]
\centering
\caption{Examples of personalized conversations obtained by RAP-Phi3-V.}
\label{tab:additional QA2}
\vspace{-3mm}
\begin{tabular}{ccc}
\toprule
Image& Concept & Information in the Database\\
\midrule
\makecell[c]{
\begin{minipage}[b]{0.2\linewidth}
    \centering
    {\includegraphics[width=1.5cm]{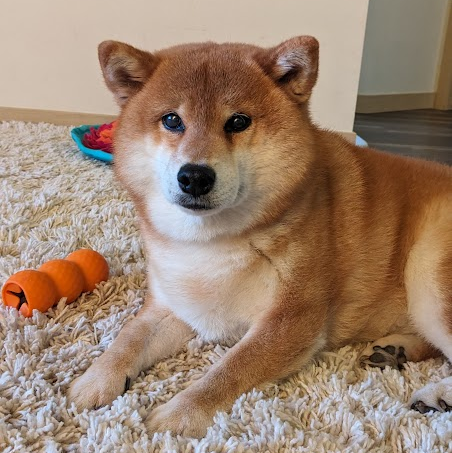}}
\end{minipage}
} &\makecell[c]{
\begin{minipage}[b]{0.12\linewidth}
    \centering
    {bo}
\end{minipage}
}&\makecell[c]{
\begin{minipage}[b]{0.58\linewidth}
    {\textlangle bo\textrangle\ is a well-groomed, medium-sized Shiba Inu with a thick, cinnamon-colored coat, cream accents, alert eyes, and a black collar.}
\end{minipage}
} \\
\midrule
\multicolumn{2}{l}{\makecell[c]{
\begin{minipage}[b]{0.3\linewidth}
    \centering
    {\includegraphics[width=3.5cm]{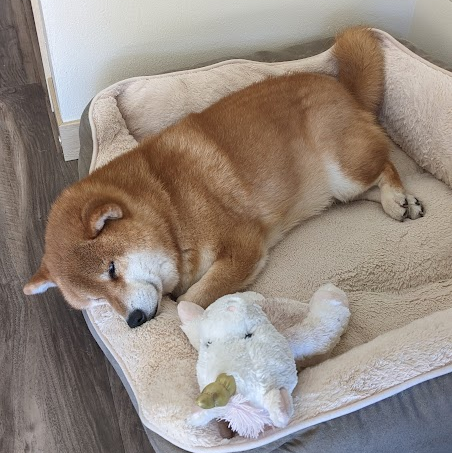}}
\end{minipage}
}}
& \makecell[c]{
\begin{minipage}[b]{0.62\linewidth}
    {
    Question: What is \textlangle bo\textrangle\ doing? \vspace{0.5mm}\\
    \adjustbox{bgcolor=Gray,minipage=[t]{\linewidth}}{\textbf{RAP-Phi3-V}:
    \textlangle bo\textrangle\ is lying on its side on a fluffy, gray dog bed. It is resting its head on a white stuffed toy. \textlangle bo\textrangle\ has a thick, cinnamon-colored coat with cream accents, alert eyes, and a black collar. \textlangle bo\textrangle\ looks very relaxed and content.}\vspace{1mm}\\
    Question: Where do you think he likes to play? \vspace{0.5mm}\\
    \adjustbox{bgcolor=Gray,minipage=[t]{\linewidth}}{\textbf{RAP-Phi3-V}:
    It's hard to say for sure, but it's possible that \textlangle bo\textrangle\ enjoys going to the park. Many dogs like to go to the park to play and get exercise. }
    }
\end{minipage}
} \\
\midrule
Image& Concept & Information in the Database\\
\midrule
\makecell[c]{
\begin{minipage}[b]{0.2\linewidth}
    \centering
    {\includegraphics[width=1.5cm]{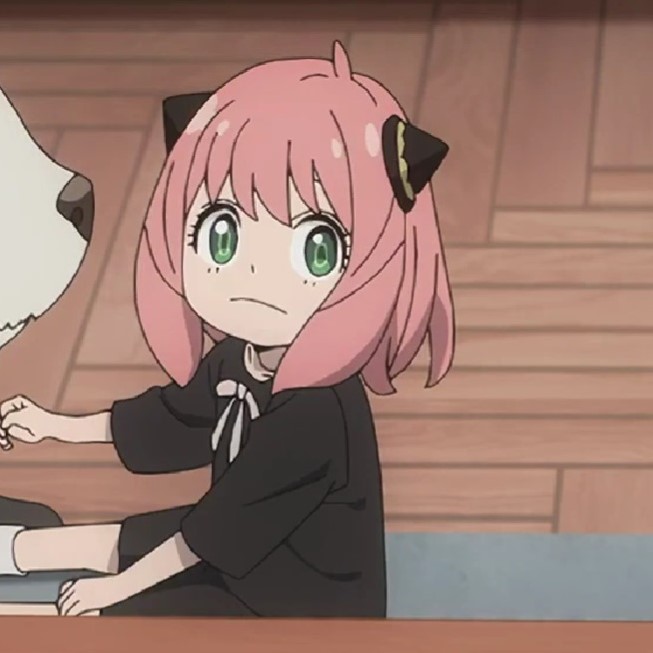}}
\end{minipage}
} &\makecell[c]{
\begin{minipage}[b]{0.12\linewidth}
    \centering
    {Anya}
\end{minipage}
}&\makecell[c]{
\begin{minipage}[b]{0.58\linewidth}
    {A young girl with pink hair and big green eyes.}
\end{minipage}
} \\
\midrule
\makecell[c]{
\begin{minipage}[b]{0.2\linewidth}
    \centering
    {\includegraphics[width=1.5cm]{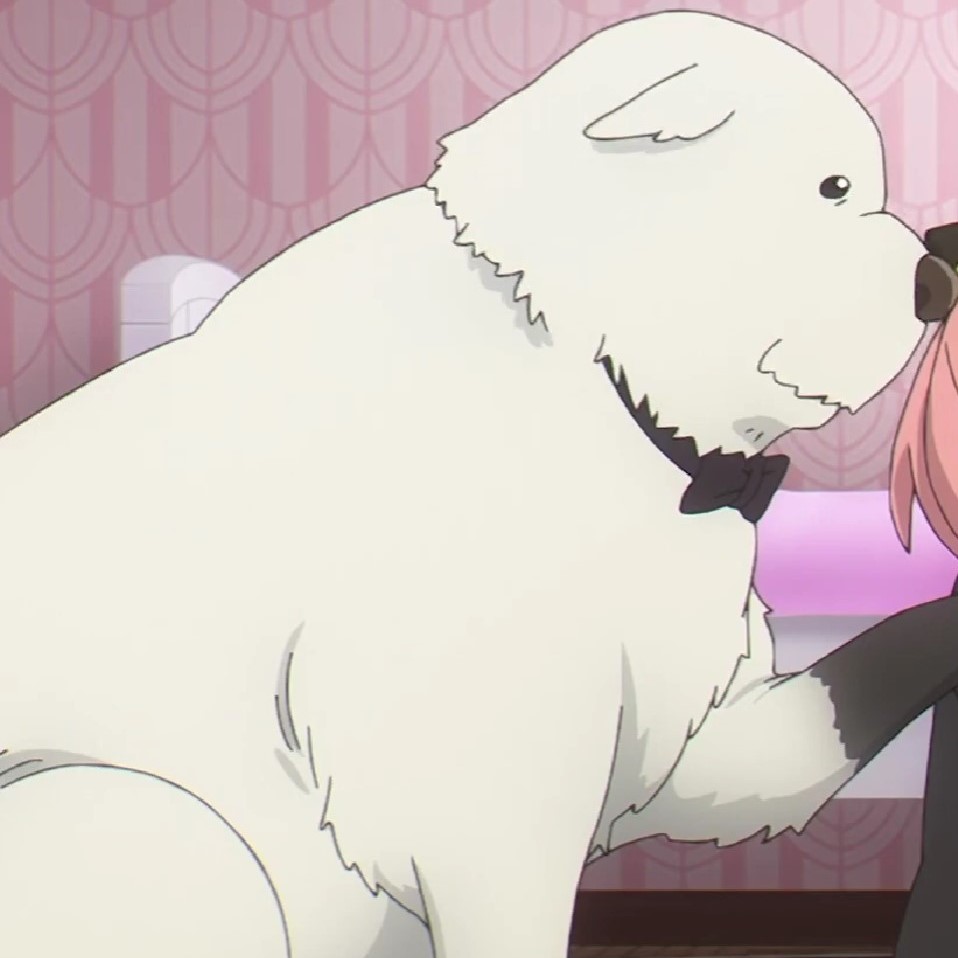}}
\end{minipage}
} &\makecell[c]{
\begin{minipage}[b]{0.12\linewidth}
    \centering
    {Bond}
\end{minipage}
}&\makecell[c]{
\begin{minipage}[b]{0.58\linewidth}
    {A white dog wearing a black bow tie.}
\end{minipage}
} \\
\midrule
\multicolumn{2}{l}{\makecell[c]{
\begin{minipage}[b]{0.3\linewidth}
    \centering
    {\includegraphics[width=3.5cm]{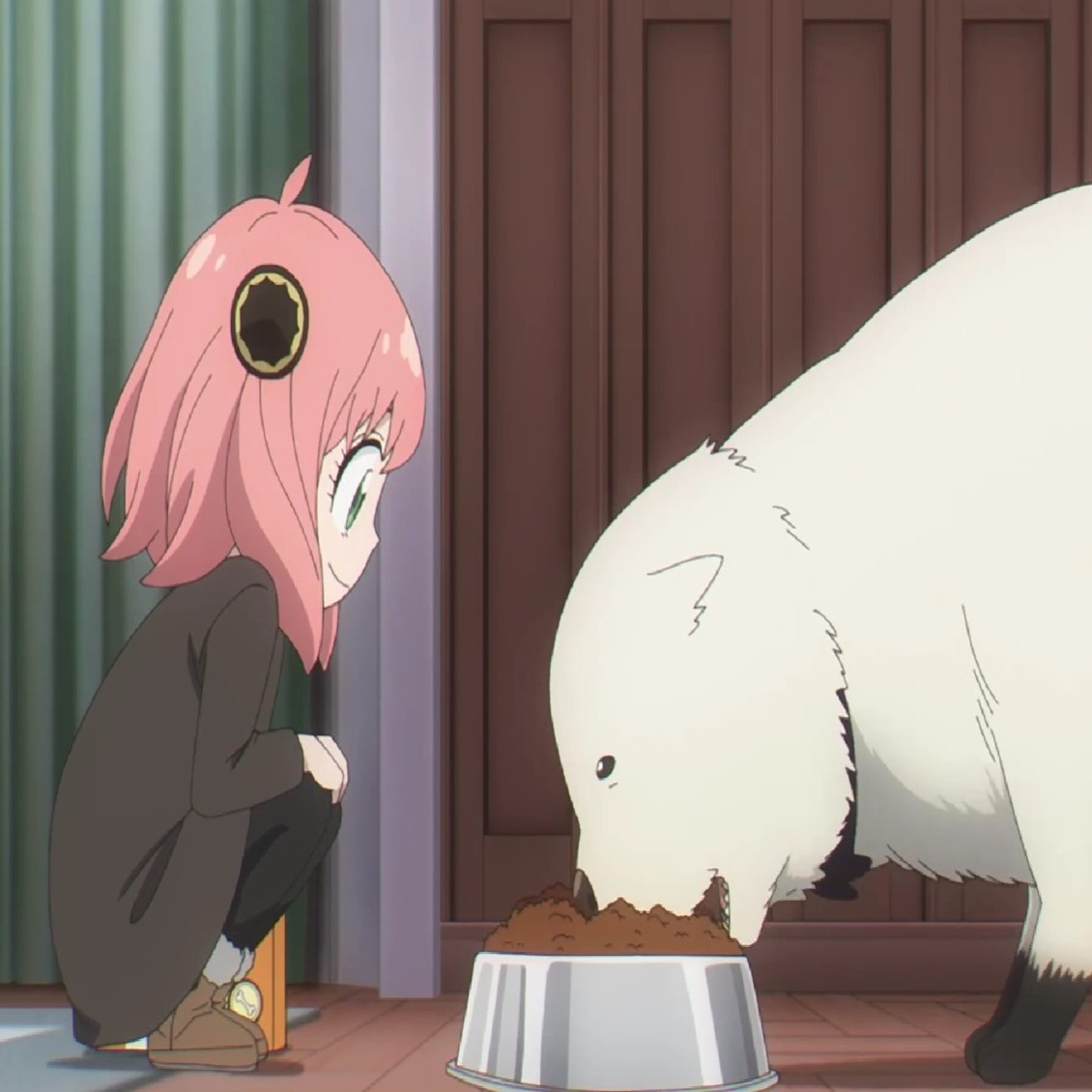}}
\end{minipage}
}}
& \makecell[c]{
\begin{minipage}[b]{0.62\linewidth}
    {
    Question: What are they doing, describe in detail. \vspace{1mm}\\
    \adjustbox{bgcolor=Gray,minipage=[t]{\linewidth}}{\textbf{RAP-Phi3-V}: \textlangle Anya\textrangle\ is feeding \textlangle Bond\textrangle, a white dog wearing a black bow tie. \textlangle Bond\textrangle\ is eating out of a bowl. \textlangle Anya\textrangle\ is kneeling down and watching \textlangle Bond\textrangle\ eat. \textlangle Bond\textrangle\ is standing on a tiled floor and looking at the bowl. \textlangle Anya\textrangle\ is wearing a pink sweater and has pink hair. \textlangle Bond\textrangle\ is wearing a black bow tie. The scene is indoors, and there is a TV in the background.}\vspace{2mm}\\
    Question: What is \textlangle Anya\textrangle\ wearing? \vspace{1mm}\\
    \adjustbox{bgcolor=Gray,minipage=[t]{\linewidth}}{\textbf{RAP-Phi3-V}:
    \textlangle Anya\textrangle\ is wearing a black sweater. }
    }
\end{minipage}
} \\
\midrule
Image& Concept & Information in the Database\\
\midrule
\makecell[c]{
\begin{minipage}[b]{0.2\linewidth}
    \centering
    {\includegraphics[width=1.5cm]{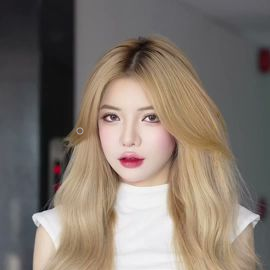}}
\end{minipage}
} &\makecell[c]{
\begin{minipage}[b]{0.12\linewidth}
    \centering
    {C}
\end{minipage}
}&\makecell[c]{
\begin{minipage}[b]{0.58\linewidth}
    {\textlangle C\textrangle\ is a young woman with long blonde hair, wearing a white sleeveless shirt. She has bright red lipstick and light makeup.}
\end{minipage}
} \\
\midrule
\multicolumn{2}{l}{\makecell[c]{
\begin{minipage}[b]{0.3\linewidth}
\centering
(Text-only conversation)
\end{minipage}
}}
& \makecell[c]{
\begin{minipage}[b]{0.62\linewidth}
    {
    \vspace{2mm}Question: What color is \textlangle C\textrangle's hair? \vspace{1mm}\\
    \adjustbox{bgcolor=Gray,minipage=[t]{\linewidth}}{\textbf{RAP-Phi3-V}:
    \textlangle C\textrangle's hair is blonde. }\vspace{2mm}\\
    Question: What color of clothes does she like? \vspace{1mm}\\
    \adjustbox{bgcolor=Gray,minipage=[t]{\linewidth}}{\textbf{RAP-Phi3-V}:
    She likes white clothes. }
    }
\end{minipage}
} \\
\bottomrule
\end{tabular}
\end{table*}

% Caption
% LLaVA
\begin{table*}[t]
\centering
\caption{Additional qualitative comparison on image captioning between RAP-LLaVA and other methods.}
\label{tab:Additional Qualitative Comparison1}
\vspace{-3mm}
\begin{tabular}{cc}
\toprule
Image & Caption\\
\midrule
\makecell[c]{
\begin{minipage}[b]{0.3\linewidth}
    \centering
    {\includegraphics[width=4.6cm]{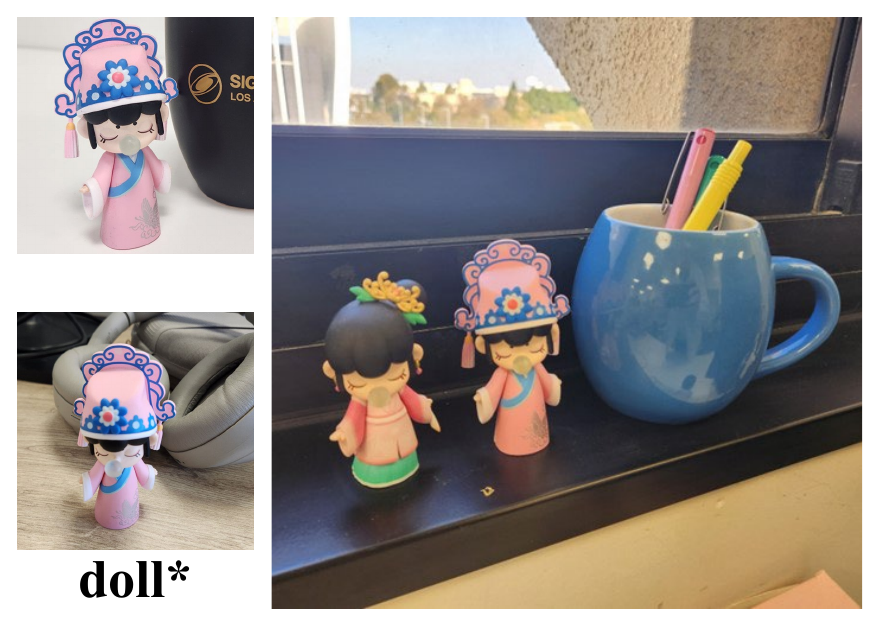}}
\end{minipage}
}& \makecell[c]{
\begin{minipage}[b]{0.62\linewidth}
    {LLaVA: A blue coffee cup sits next to two dolls on a window sill.\\
    LLaVA-LoRA:  \textcolor{cgreen}{\textlangle doll*\textrangle} positioned on a windowsill with a blue mug containing pencils\\
    MyVLM: \textcolor{cgreen}{\textlangle doll*\textrangle} is found sitting beside a window ledge with a cup and toothbrushes nearby. In the image, \textcolor{cgreen}{\textlangle doll*\textrangle} appears to be peacefully resting, possibly enjoying the view or taking a break.\\
    \adjustbox{bgcolor=Gray,minipage=[t]{\linewidth}}{\textbf{RAP-LLaVA}(Ours): \textcolor{cgreen}{\textlangle doll*\textrangle} and her friend are ready for a fun day at the office! They're ready to take on any challenge that comes their way.}}
\end{minipage}
} \\
\midrule
\makecell[c]{
\begin{minipage}[b]{0.3\linewidth}
    \centering
    {\includegraphics[width=4.6cm]{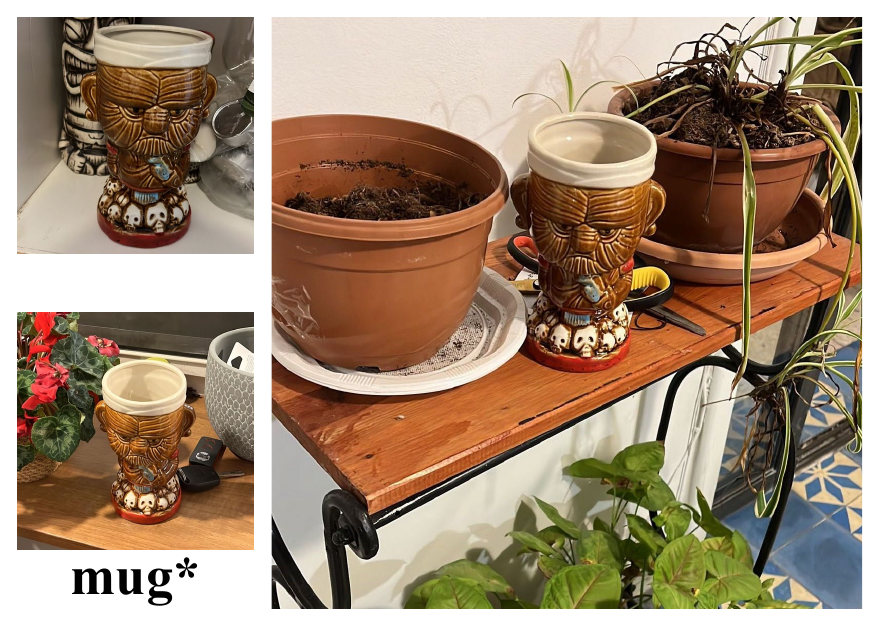}}
\end{minipage}
}& \makecell[c]{
\begin{minipage}[b]{0.62\linewidth}
    {LLaVA: A wooden shelf holds several potted plants, including a large clay pot and a small vase. The shelf is positioned near a wall, and the plants are arranged in a visually appealing manner.\\
    LLaVA-LoRA:  \textcolor{cgreen}{\textlangle mug*\textrangle} on a wooden shelf next to a plant and two potted plants. \\
    MyVLM: \textcolor{cgreen}{\textlangle mug*\textrangle} on a a blue and white tiled floor next to indoor plants, a shelf with pots and a vase.\\
 \adjustbox{bgcolor=Gray,minipage=[t]{\linewidth}}{\textbf{RAP-LLaVA}(Ours): A brown ceramic tiki mug with the face of a tiki head, \textcolor{cgreen}{\textlangle mug*\textrangle}, sits on a wooden shelf with two potted plants. }  }
\end{minipage}
} \\
\midrule
\makecell[c]{
\begin{minipage}[b]{0.3\linewidth}
    \centering
    {\includegraphics[width=4.6cm]{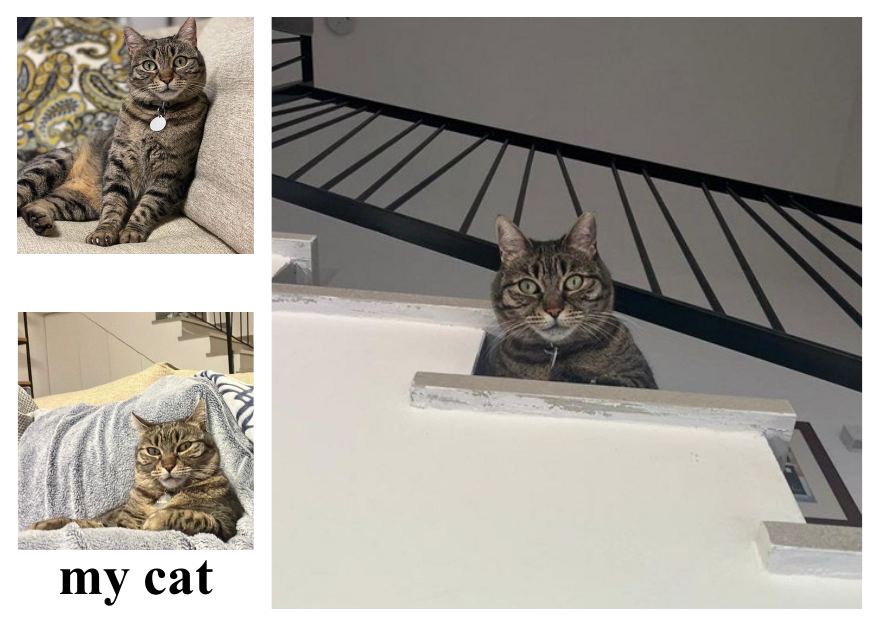}}
\end{minipage}
}& \makecell[c]{
\begin{minipage}[b]{0.62\linewidth}
    {LLaVA: A cat is sitting on a ledge near a staircase.\\
    LLaVA-LoRA: \textcolor{cgreen}{\textlangle my cat\textrangle} sitting on a staircase, looking down.\\
    MyVLM: \textcolor{cgreen}{\textlangle my cat\textrangle} in a typical pose, gripping the banister of a staircase. He is wearing a collar.\\
    \adjustbox{bgcolor=Gray,minipage=[t]{\linewidth}}{\textbf{RAP-LLaVA}(Ours): \textcolor{cgreen}{\textlangle my cat\textrangle} is relaxing on the stairs. Look at those eyes!  What a beautiful cat.}}
\end{minipage}
} \\
\midrule
\makecell[c]{
\begin{minipage}[b]{0.3\linewidth}
    \centering
    {\includegraphics[width=4.6cm]{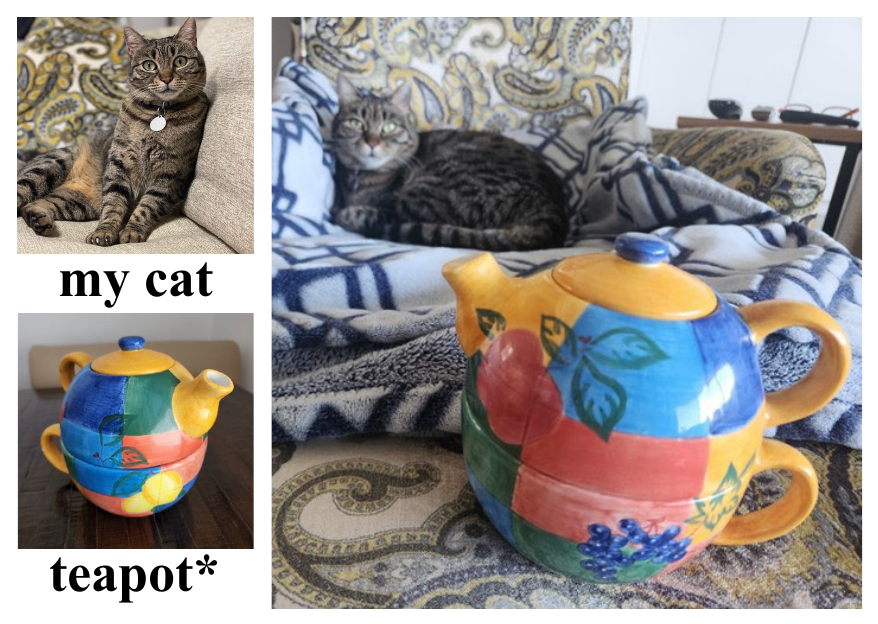}}
\end{minipage}
}& \makecell[c]{
\begin{minipage}[b]{0.62\linewidth}
    {LLaVA: A cat is laying on a blanket on a couch, next to a colorful teapot.\\
    LLaVA-LoRA: \textcolor{cgreen}{\textlangle my cat\textrangle} reclining on a chair with a \textcolor{cgreen}{\textlangle teapot*\textrangle} beside \\
    MyVLM: \textcolor{cgreen}{\textlangle teapot*\textrangle} on the couch near an orange, yellow, and blue teapot. \textcolor{cred}{The cat} is laying on a blanket, and the teapot is placed on a table.\\
    \adjustbox{bgcolor=Gray,minipage=[t]{\linewidth}}{\textbf{RAP-LLaVA}(Ours): \textcolor{cgreen}{\textlangle my cat\textrangle} is taking a well-deserved nap next to \textcolor{cgreen}{\textlangle teapot*\textrangle}.}}
\end{minipage}
} \\
\midrule
\makecell[c]{
\begin{minipage}[b]{0.3\linewidth}
    \centering
    {\includegraphics[width=4.6cm]{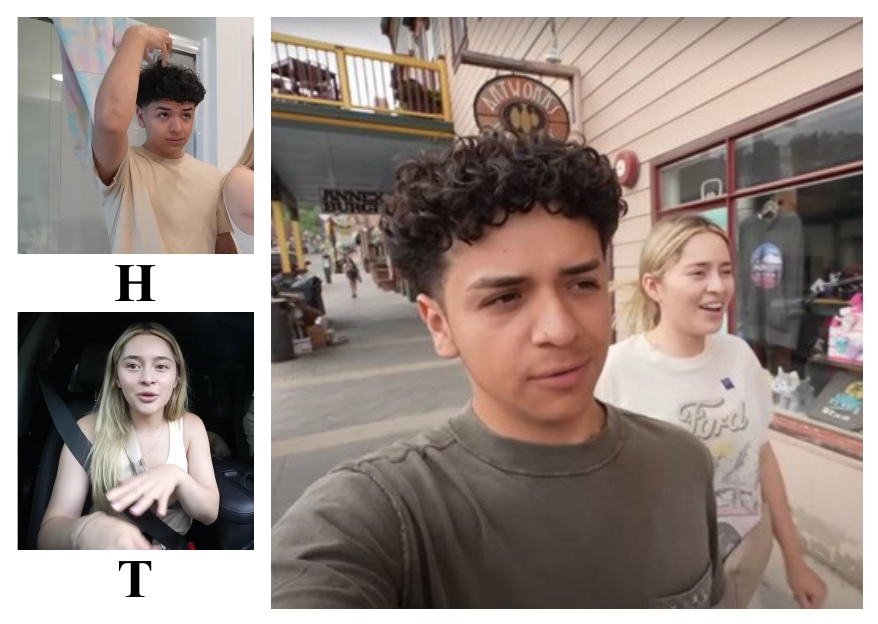}}
\end{minipage}
}& \makecell[c]{
\begin{minipage}[b]{0.62\linewidth}
    {LLaVA: A man and a woman are walking down a street, with the man taking a selfie using his cell phone. They are both smiling as they walk, and the man is holding his phone \dots\\
    LLaVA-LoRA: \textcolor{cgreen}{\textlangle H\textrangle} and \textcolor{cred}{\textlangle K\textrangle} are looking happy as they walk down the street together.\\
    MyVLM: \textcolor{cgreen}{\textlangle H\textrangle} and \textcolor{cgreen}{\textlangle T\textrangle} are looking very serious as they take in the scenery.\\
    \adjustbox{bgcolor=Gray,minipage=[t]{\linewidth}}{\textbf{RAP-LLaVA}(Ours): \textcolor{cgreen}{\textlangle T\textrangle} is walking down the street with her friend \textcolor{cgreen}{\textlangle H\textrangle}.}}
\end{minipage}
} \\
\midrule
\makecell[c]{
\begin{minipage}[b]{0.3\linewidth}
    \centering
    {\includegraphics[width=4.6cm]{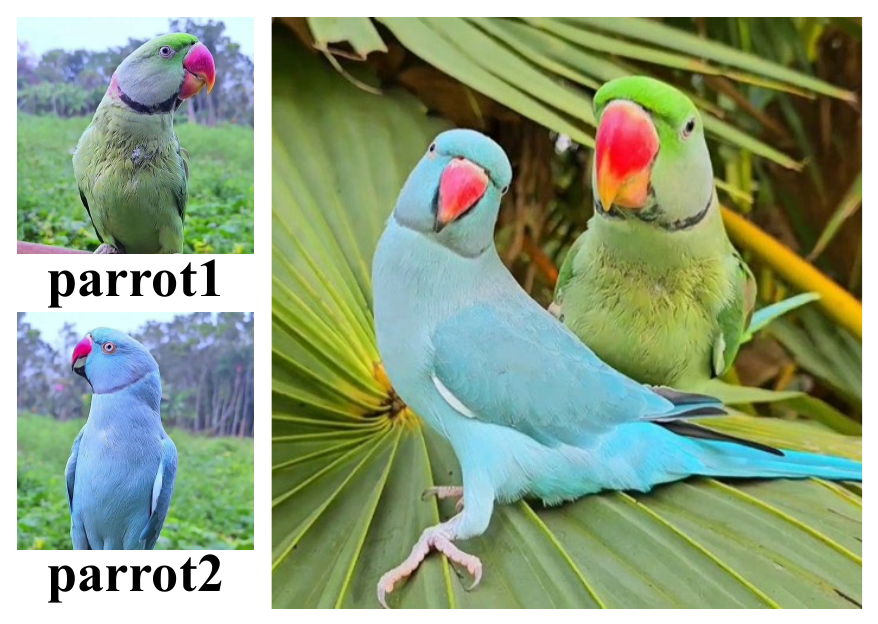}}
\end{minipage}
}& \makecell[c]{
\begin{minipage}[b]{0.62\linewidth}
    {LLaVA: Two colorful parrots are perched on a palm tree, sitting next to each other.\\
    LLaVA-LoRA: \textcolor{cgreen}{\textlangle parrot1\textrangle} and \textcolor{cgreen}{\textlangle parrot2\textrangle} are looking very serious as they perch on a palm frond.\\
    MyVLM: \textcolor{cgreen}{\textlangle parrot1\textrangle} and \textcolor{cgreen}{\textlangle parrot2\textrangle} look so cute together.\\
   \adjustbox{bgcolor=Gray,minipage=[t]{\linewidth}}{ \textbf{RAP-LLaVA}(Ours): \textcolor{cgreen}{\textlangle parrot1\textrangle} and \textcolor{cgreen}{\textlangle parrot2\textrangle} are both beautiful birds. \textcolor{cgreen}{\textlangle parrot1\textrangle} has a bright red beak and \textcolor{cgreen}{\textlangle parrot2\textrangle} has a light blue head and pink feet. They are both perched on a branch.}}
\end{minipage}
} \\
\bottomrule
\end{tabular}
\end{table*}
% Phi3-V
\begin{table*}[t]
\centering
\caption{Additional qualitative comparison on image captioning between RAP-Phi3-V and other methods.}
\label{tab:Additional Qualitative Comparison2}
\vspace{-3mm}
\begin{tabular}{cc}
\toprule
Image & Caption\\
\midrule
\makecell[c]{
\begin{minipage}[b]{0.3\linewidth}
    \centering
    {\includegraphics[width=4.6cm]{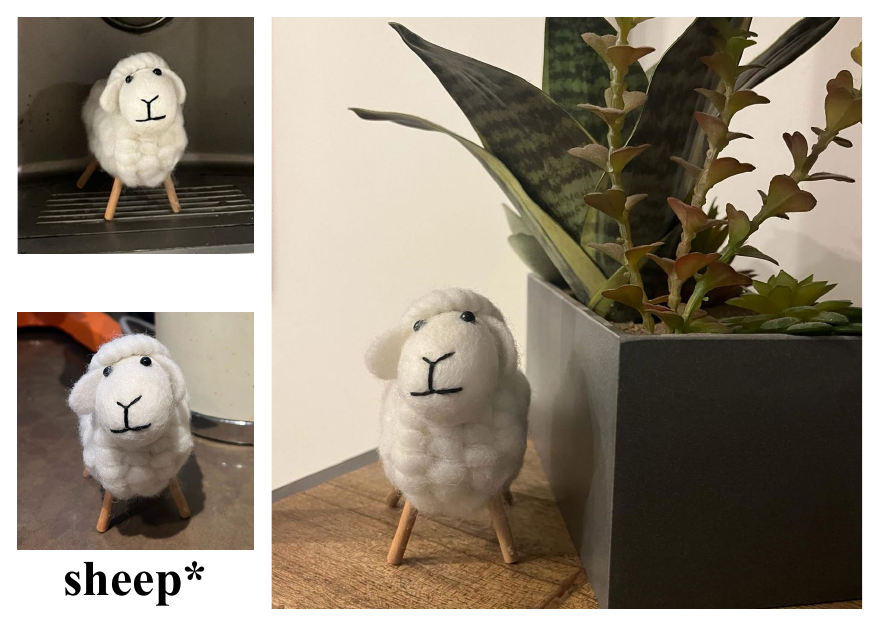}}
\end{minipage}
}& \makecell[c]{
\begin{minipage}[b]{0.62\linewidth}
    {Phi3-V: A small white sheep figurine is sitting on a wooden table next to a plant.\\
    LLaVA-LoRA:  \textcolor{cgreen}{\textlangle sheep*\textrangle} positioned next to a potted plant on a wooden table.\\
    MyVLM: \textcolor{cgreen}{\textlangle sheep*\textrangle} positioned within a square shape within a vase made of metal positioned on a table made of wood.\\
    \adjustbox{bgcolor=Gray,minipage=[t]{\linewidth}}{\textbf{RAP-Phi3-V}(Ours): \textcolor{cgreen}{\textlangle sheep*\textrangle} is ready to play on the shelf with the plant!}}
\end{minipage}
} \\
\midrule
\makecell[c]{
\begin{minipage}[b]{0.3\linewidth}
    \centering
    {\includegraphics[width=4.6cm]{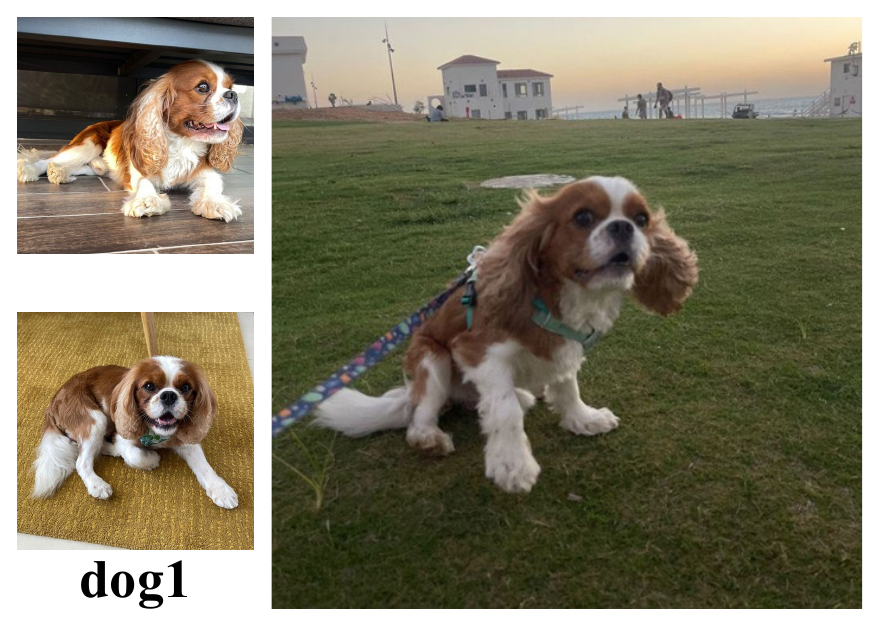}}
\end{minipage}
}& \makecell[c]{
\begin{minipage}[b]{0.62\linewidth}
    {Phi3-V: A brown and white dog is sitting on the grass with a leash attached to it.\\
    LLaVA-LoRA: \textcolor{cgreen}{\textlangle dog1\textrangle} sitting on the grass with a look of determination.\\
    MyVLM: on a cheerful note, \textcolor{cgreen}{\textlangle dog1\textrangle} restfully resides, surrounded by the soothing sounds of nature.\\
    \adjustbox{bgcolor=Gray,minipage=[t]{\linewidth}}{\textbf{RAP-Phi3-V}(Ours): This is a picture of \textcolor{cgreen}{\textlangle dog1\textrangle}, a cute dog with light brown and white fur. He looks happy and playful.}}
\end{minipage}
} \\
\midrule
\makecell[c]{
\begin{minipage}[b]{0.3\linewidth}
    \centering
    {\includegraphics[width=4.6cm]{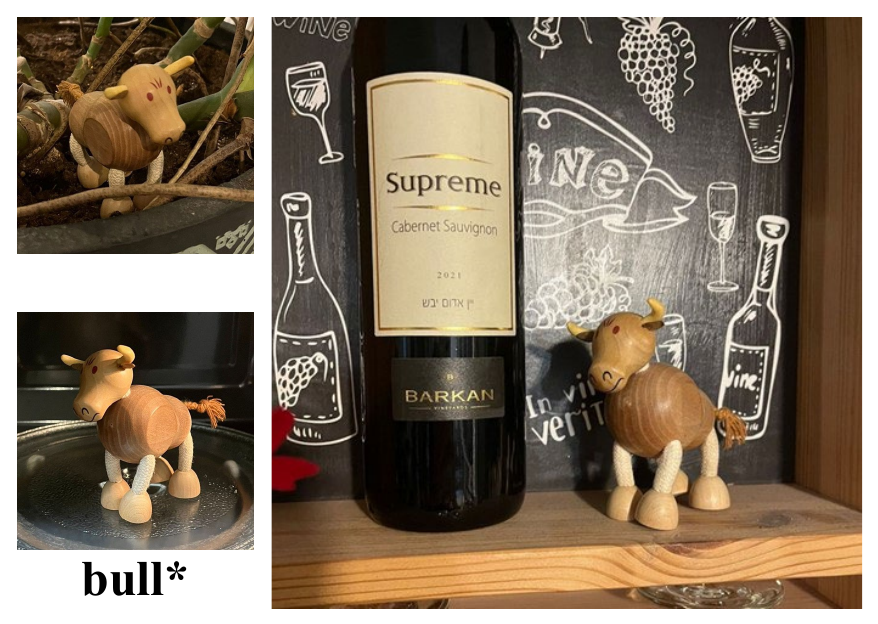}}
\end{minipage}
}& \makecell[c]{
\begin{minipage}[b]{0.62\linewidth}
    {Phi3-V: A bottle of wine is displayed next to a wooden cow statue.\\
    LLaVA-LoRA: \textcolor{cgreen}{\textlangle bull*\textrangle} positioned next to a bottle of supreme cabernet sauvignon.\\
    MyVLM: \textcolor{cred}{\textlangle gold pineapple\textrangle} pars the bottle of wine. A bottle of wine sits next to a bottle of wine.\\
    \adjustbox{bgcolor=Gray,minipage=[t]{\linewidth}}{\textbf{RAP-Phi3-V}(Ours): \textcolor{cgreen}{\textlangle bull*\textrangle} stands guard over a bottle of SUPREME wine.}}
\end{minipage}
} \\
\midrule
\makecell[c]{
\begin{minipage}[b]{0.3\linewidth}
    \centering
    {\includegraphics[width=4.6cm]{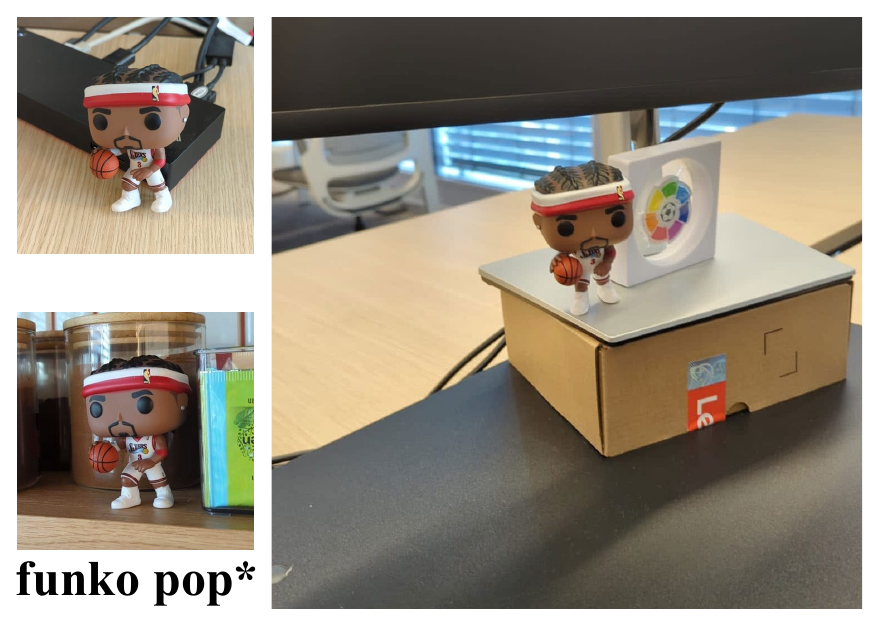}}
\end{minipage}
}& \makecell[c]{
\begin{minipage}[b]{0.62\linewidth}
    {Phi3-V: A small figurine of a basketball player is placed on top of a box.\\
    LLaVA-LoRA: \textcolor{cgreen}{\textlangle funko pop*\textrangle} sitting on a desk next to a box and a monitor. \\
    MyVLM: \textcolor{cgreen}{\textlangle funko pop*\textrangle} stands out in a photo set against a dark background.\\
    \adjustbox{bgcolor=Gray,minipage=[t]{\linewidth}}{\textbf{RAP-Phi3-V}(Ours): \textcolor{cgreen}{\textlangle funko pop*\textrangle} is ready to dominate the court!}}
\end{minipage}
} \\
\midrule
\makecell[c]{
\begin{minipage}[b]{0.3\linewidth}
    \centering
    {\includegraphics[width=4.6cm]{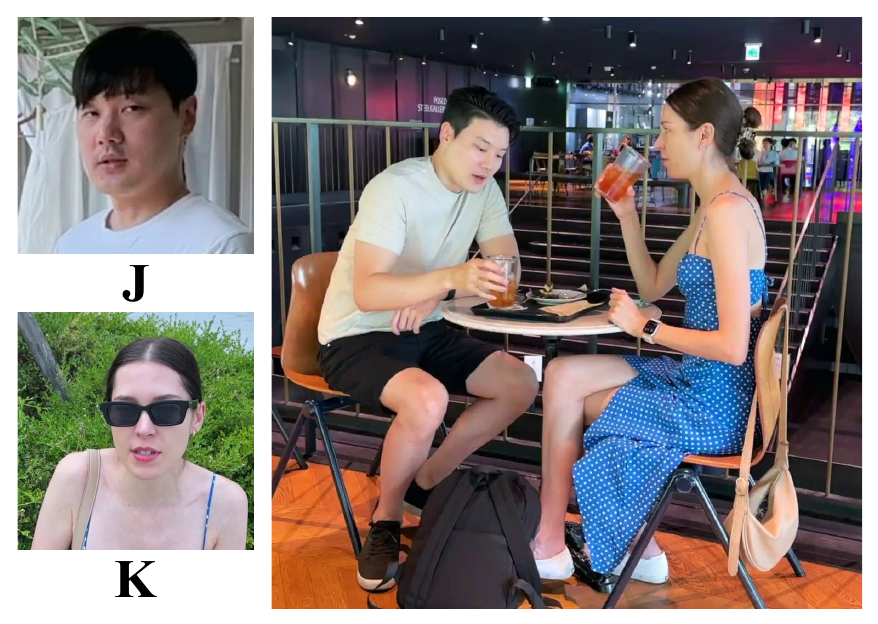}}
\end{minipage}
}& \makecell[c]{
\begin{minipage}[b]{0.62\linewidth}
    {Phi3-V: A man and woman are sitting at a table in a restaurant, having a conversation and enjoying their time together.\\
    LLaVA-LoRA: \textcolor{cgreen}{\textlangle J\textrangle} and \textcolor{cgreen}{\textlangle K\textrangle} enjoying a meal together.\\
    MyVLM: \textcolor{cgreen}{\textlangle J\textrangle} and \textcolor{cgreen}{\textlangle K\textrangle} enjoying their time together. \\
    \adjustbox{bgcolor=Gray,minipage=[t]{\linewidth}}{\textbf{RAP-Phi3-V}(Ours): \textcolor{cgreen}{\textlangle J\textrangle} and \textcolor{cgreen}{\textlangle K\textrangle} are enjoying a drink at a bar.}}
\end{minipage}
} \\
\midrule
\makecell[c]{
\begin{minipage}[b]{0.3\linewidth}
    \centering
    {\includegraphics[width=4.6cm]{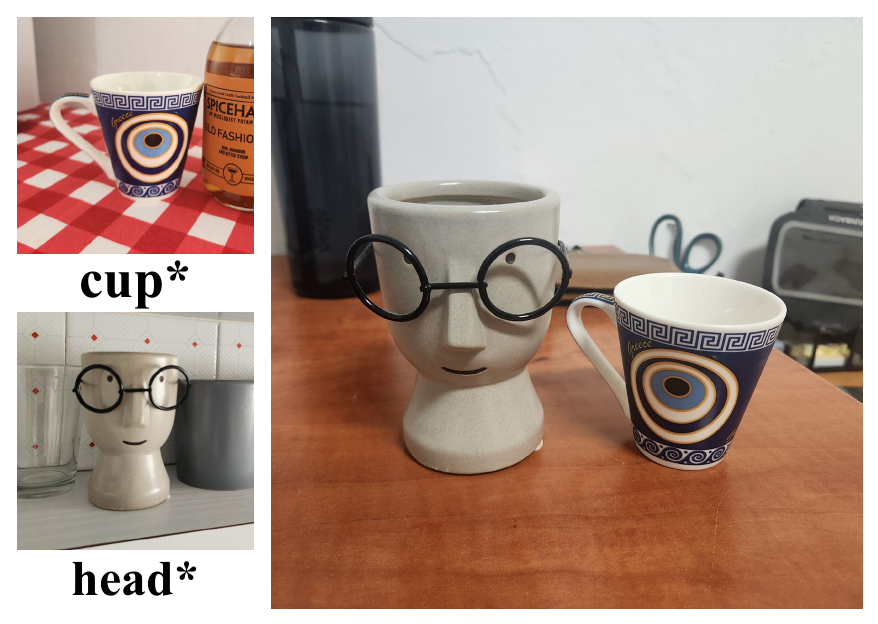}}
\end{minipage}
}& \makecell[c]{
\begin{minipage}[b]{0.62\linewidth}
    {Phi3-V: A small statue of a man wearing glasses is sitting next to a cup on a table.\\
    LLaVA-LoRA: A table hosts \textcolor{cgreen}{\textlangle cup*\textrangle} and a sculpture of a head wearing glasses.\\
    MyVLM: A table with \textcolor{cgreen}{\textlangle cup*\textrangle}, an espresso cup, and \textcolor{cred}{a ceramic mug} depicting a face wearing glasses sits on a wooden surface.\\
   \adjustbox{bgcolor=Gray,minipage=[t]{\linewidth}}{ \textbf{RAP-Phi3-V}(Ours): \textcolor{cgreen}{\textlangle head*\textrangle}  and \textcolor{cgreen}{\textlangle cup*\textrangle} are ready for their morning coffee.}}
\end{minipage}
} \\
\bottomrule
\end{tabular}
\end{table*}

% Description
% Image Description
% LLaVA
\begin{table*}[t]
\centering
\caption{Qualitative results of personalized image description obtained by RAP-LLaVA.}
\label{tab:Additional description1}
\vspace{-3mm}
\begin{tabular}{cc}
\toprule
Image & Description\\
\midrule
\makecell[c]{
\begin{minipage}[b]{0.3\linewidth}
    \centering
    {\includegraphics[width=4.6cm]{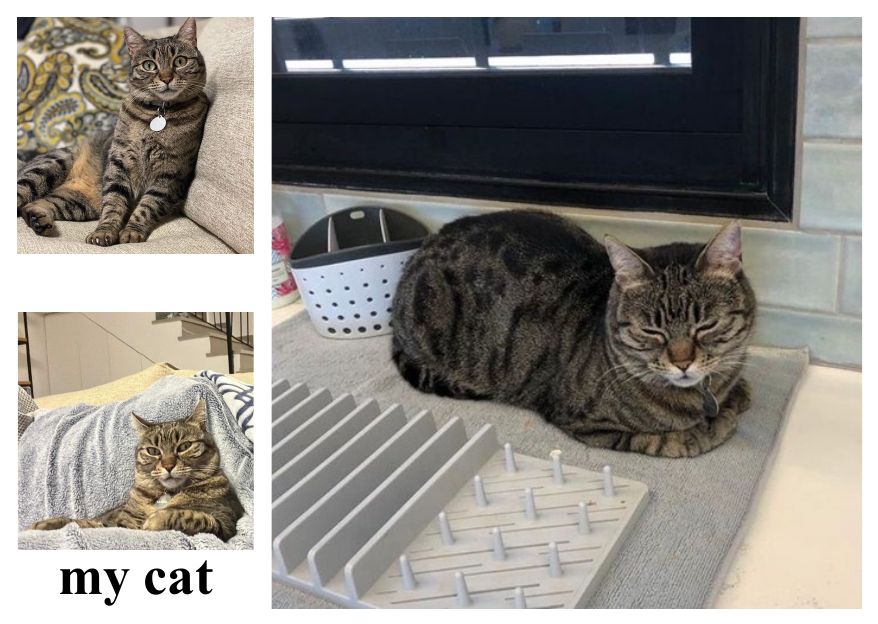}}
\end{minipage}
}& \makecell[c]{
\begin{minipage}[b]{0.62\linewidth}
    {LLaVA: The image features a large, fluffy cat sitting on a countertop, possibly in a kitchen. The cat is positioned near a sink and appears to be resting or sleeping. The countertop is covered with a towel, providing a comfortable surface for the cat to lay on \dots 
    \vspace{0.5mm}\\
    \adjustbox{bgcolor=Gray,minipage=[t]{\linewidth}}{\textbf{RAP-LLaVA}(Ours): The image shows \textcolor{cgreen}{\textlangle my cat\textrangle}, a tabby cat with its eyes closed, laying on a light grey surface. The cat's fur is a mix of light and dark brown stripes.  The cat is relaxed and comfortable.} }
\end{minipage}
} \\
\midrule
\makecell[c]{
\begin{minipage}[b]{0.3\linewidth}
    \centering
    {\includegraphics[width=4.6cm]{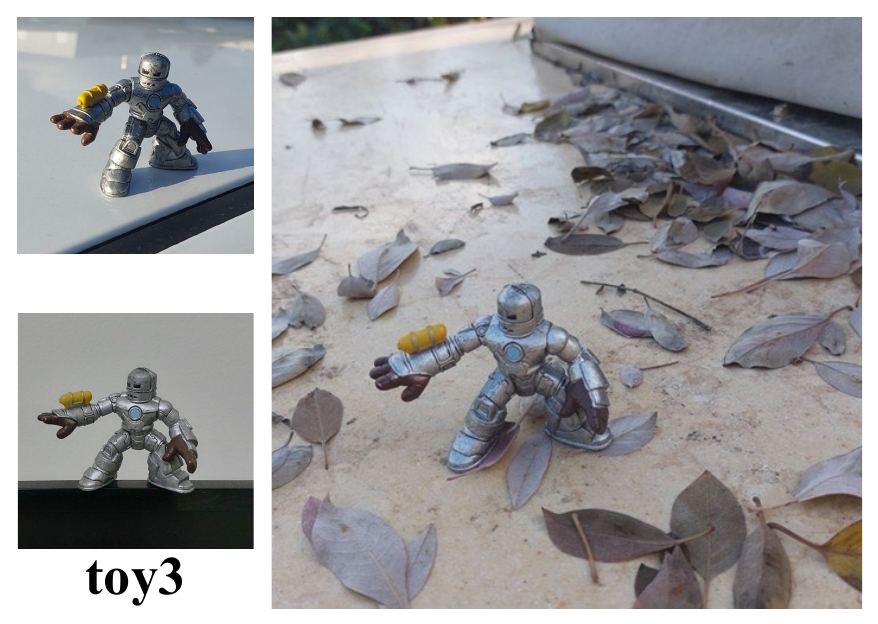}}
\end{minipage}
}& \makecell[c]{
\begin{minipage}[b]{0.62\linewidth}
    {LLaVA: The image features a small metal robot figure, possibly a toy, standing on a table or counter. The robot is positioned in the center of the scene, surrounded by a variety of leaves scattered around it. The leaves are of different sizes and \dots
    \vspace{0.5mm}\\
    \adjustbox{bgcolor=Gray,minipage=[t]{\linewidth}}{\textbf{RAP-LLaVA}(Ours): The image shows a toy figurine of Iron Man, a superhero from Marvel comics. He is wearing a silver and red suit of armor and is standing on a surface covered in leaves. The toy is called \textcolor{cgreen}{\textlangle toy3\textrangle}.} }
\end{minipage}
} \\
\midrule
\makecell[c]{
\begin{minipage}[b]{0.3\linewidth}
    \centering
    {\includegraphics[width=4.6cm]{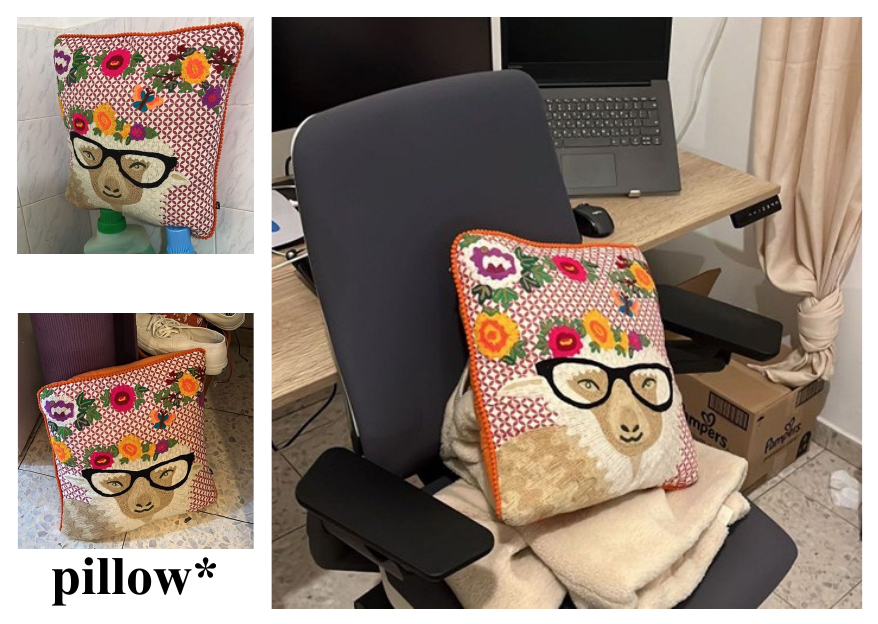}}
\end{minipage}
}& \makecell[c]{
\begin{minipage}[b]{0.62\linewidth}
    {LLaVA: The image features a chair with a pillow on it, which has a picture of a sheep on it. The chair is positioned in front of a desk with a laptop and a monitor on it. The laptop is placed on the left side of the desk, while the monitor is on the right side \dots 
    \vspace{0.5mm}\\
    \adjustbox{bgcolor=Gray,minipage=[t]{\linewidth}}{\textbf{RAP-LLaVA}(Ours): The image shows a chair with a \textcolor{cgreen}{\textlangle pillow*\textrangle} on it. The \textcolor{cgreen}{\textlangle pillow*\textrangle} is a pattern of flowers, leaves, and a sheep wearing glasses with a flower crown. There is also a laptop and a computer monitor in the background.}}
\end{minipage}
} \\
\midrule
\makecell[c]{
\begin{minipage}[b]{0.3\linewidth}
    \centering
    {\includegraphics[width=4.6cm]{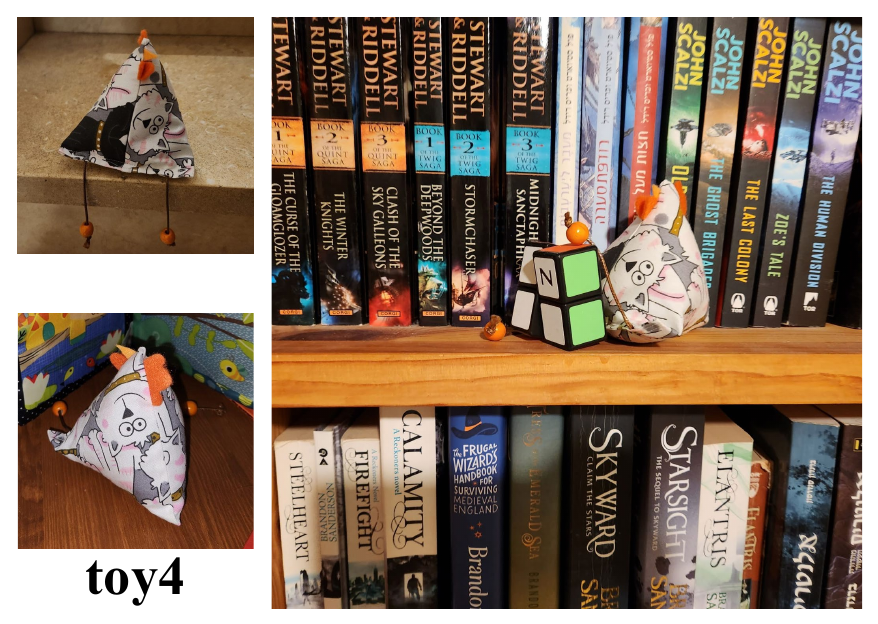}}
\end{minipage}
}& \makecell[c]{
\begin{minipage}[b]{0.62\linewidth}
    {LLaVA: The image features a bookshelf filled with various books. On the shelf, there are two Rubik's cubes, one green and one white, sitting next to each other. The bookshelf is well-stocked with books, occupying most of \dots
    \vspace{0.5mm}\\
   \adjustbox{bgcolor=Gray,minipage=[t]{\linewidth}}{ \textbf{RAP-LLaVA}(Ours): The image shows a bookshelf with books on it. There are two toys on the shelf. One is a \textcolor{cgreen}{\textlangle toy4\textrangle}, which looks like a cat with orange fuzzy ears and a brown cord attached. The other is a Rubik's cube.} }
\end{minipage}
} \\
\midrule
\makecell[c]{
\begin{minipage}[b]{0.3\linewidth}
    \centering
    {\includegraphics[width=4.6cm]{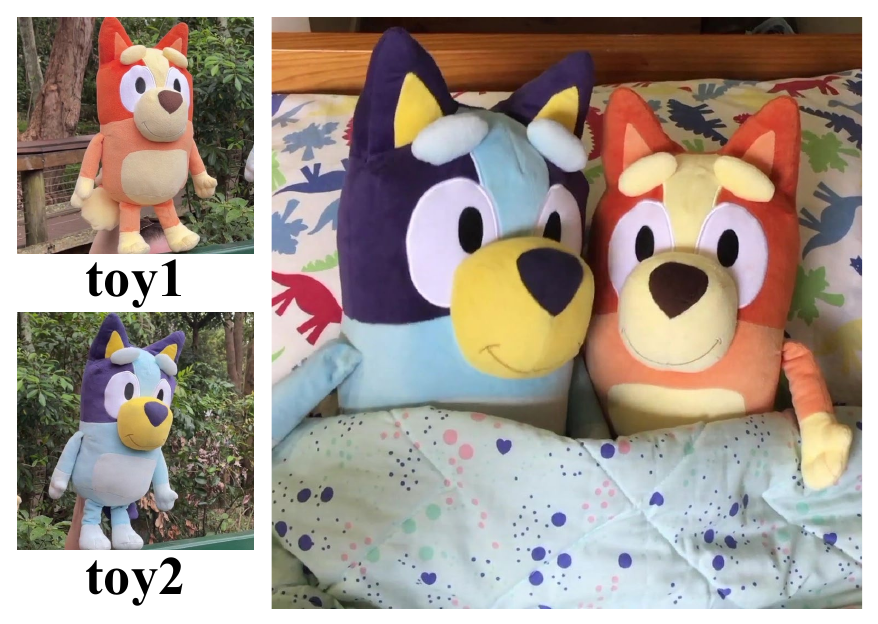}}
\end{minipage}
}& \makecell[c]{
\begin{minipage}[b]{0.62\linewidth}
    {LLaVA: The image features a bed with two stuffed animals, a dog and a cat, lying on it. The dog is positioned on the left side of the bed, while the cat is on the right side. The stuffed animals are placed close to each other, giving the impression of a cozy and \dots
    \vspace{0.5mm}\\
    \adjustbox{bgcolor=Gray,minipage=[t]{\linewidth}}{\textbf{RAP-LLaVA}(Ours): The image shows two plush toys, \textcolor{cgreen}{\textlangle toy1\textrangle} and \textcolor{cgreen}{\textlangle toy2\textrangle}, lying in bed together. \textcolor{cgreen}{\textlangle toy2\textrangle} is a light blue plush with a purple patch on its head and yellow ears. \textcolor{cgreen}{\textlangle toy1\textrangle} is an orange plush with a yellow belly and a brown nose. They are both looking at the camera.}}
\end{minipage}
} \\
\midrule
\makecell[c]{
\begin{minipage}[b]{0.3\linewidth}
    \centering
    {\includegraphics[width=4.6cm]{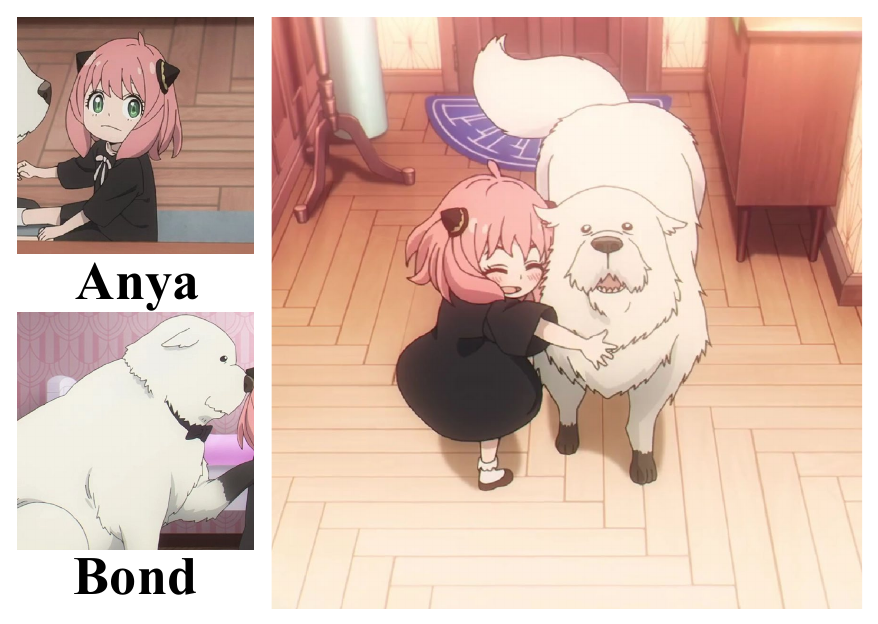}}
\end{minipage}
}& \makecell[c]{
\begin{minipage}[b]{0.62\linewidth}
    {LLaVA: The image features a young girl hugging a large white dog in a room. The girl is wearing a black dress and appears to be enjoying her time with the dog. The dog is positioned close to the girl, and they seem to be sharing a tender moment together \dots
    \vspace{0.5mm}\\
    \adjustbox{bgcolor=Gray,minipage=[t]{\linewidth}}{\textbf{RAP-LLaVA}(Ours): \textcolor{cgreen}{\textlangle Anya\textrangle} is a young girl with pink hair and big green eyes. She is wearing a black dress and black shoes. She is standing in a room with wooden floors and is hugging a white dog named \textcolor{cgreen}{\textlangle Bond\textrangle}. \textcolor{cgreen}{\textlangle Bond\textrangle} is wearing a black bow tie. The dog is standing on its hind legs and is looking at the camera. The room is decorated with a blue rug and a wooden cabinet \dots}}
\end{minipage}
} \\
\bottomrule
\end{tabular}
\end{table*}
% Image Description
% Phi3-V
\begin{table*}[t]
\centering
\caption{Qualitative results of personalized image description obtained by RAP-Phi3-V.}
\label{tab:Additional description2}
\vspace{-3mm}
\begin{tabular}{cc}
\toprule
Image & Description\\
\midrule
\makecell[c]{
\begin{minipage}[b]{0.3\linewidth}
    \centering
    {\includegraphics[width=4.6cm]{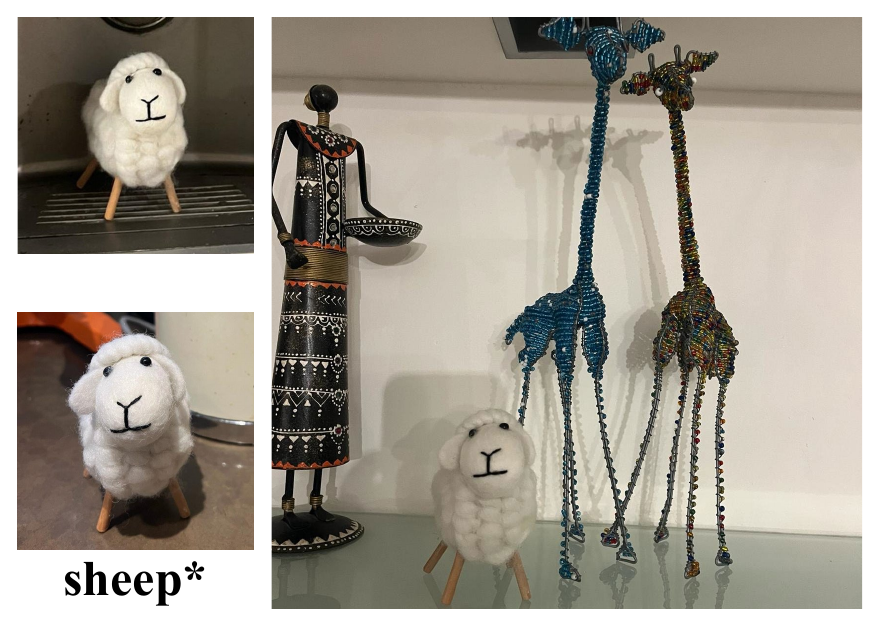}}
\end{minipage}
}& \makecell[c]{
\begin{minipage}[b]{0.62\linewidth}
    {Phi3-V: The image features a display of various animal figurines on a shelf. There are three giraffes, two of which are standing and one sitting, placed next to each other. A sheep figurine is also present on the shelf, located between the giraffes \dots 
    \vspace{0.5mm}\\
    \adjustbox{bgcolor=Gray,minipage=[t]{\linewidth}}{\textbf{RAP-Phi3-V}(Ours): The image shows a white \textcolor{cgreen}{\textlangle sheep*\textrangle} sitting on a shelf next to a giraffe figurine. The \textcolor{cgreen}{\textlangle sheep*\textrangle} is made of felt and has black eyes and a black nose. The giraffe figurine is made of metal and has a colorful necklace around its neck. The shelf is white \dots}}
\end{minipage}
} \\
\midrule
\makecell[c]{
\begin{minipage}[b]{0.3\linewidth}
    \centering
    {\includegraphics[width=4.6cm]{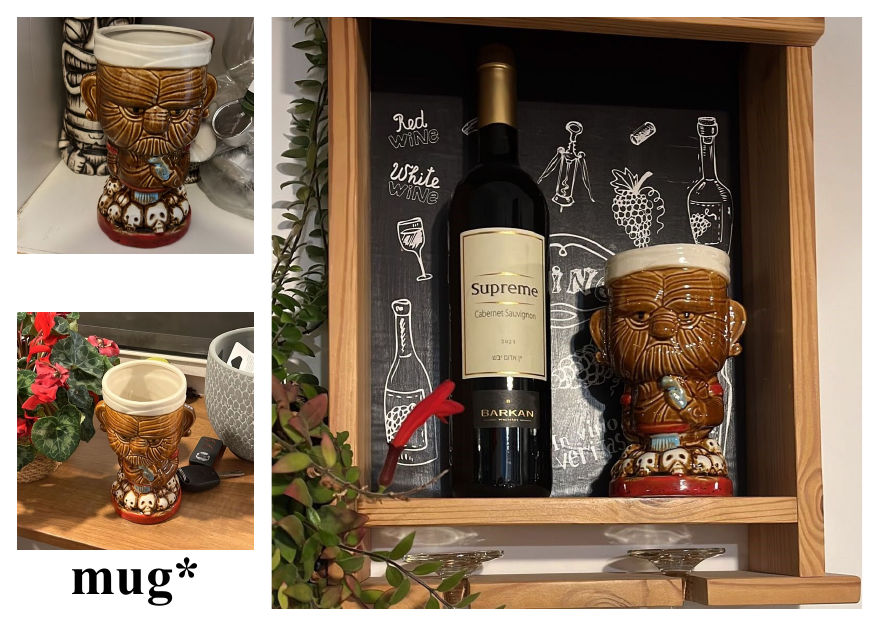}}
\end{minipage}
}& \makecell[c]{
\begin{minipage}[b]{0.62\linewidth}
    {Phi3-V: The image features a wooden shelf with two wine bottles and two wine glasses. One of the wine bottles is positioned towards the left side of the shelf, while the other is located more towards the center. The wine glasses are placed on the right side of the \dots
    \vspace{0.5mm}\\
    \adjustbox{bgcolor=Gray,minipage=[t]{\linewidth}}{\textbf{RAP-Phi3-V}(Ours): The image shows a wooden shelf with three glassware items on it. The top glassware item is a \textcolor{cgreen}{\textlangle mug*\textrangle} with a tiki head design. The other glassware items are wine glasses. There is a sign on the wine bottle that says "SUPREME". There is also a plant in a pot in the background.} }
\end{minipage}
} \\
\midrule
\makecell[c]{
\begin{minipage}[b]{0.3\linewidth}
    \centering
    {\includegraphics[width=4.6cm]{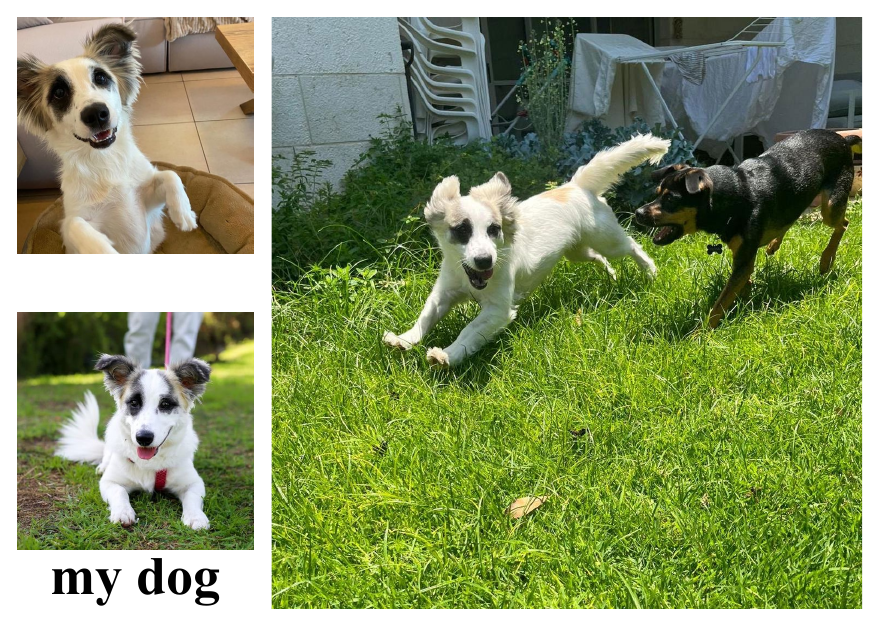}}
\end{minipage}
}& \makecell[c]{
\begin{minipage}[b]{0.62\linewidth}
    {Phi3-V: The image features two dogs playing together in a grassy yard. One dog is located on the left side of the yard, while the other dog is on the right side. They are both running and enjoying their time outside. In the background, there are several chairs \dots 
    \vspace{0.5mm}\\
    \adjustbox{bgcolor=Gray,minipage=[t]{\linewidth}}{\textbf{RAP-Phi3-V}(Ours): The image shows two dogs playing in a grassy backyard. The dog on the left is a white and gray dog with long fur. The dog on the right is a black and brown dog with long fur. The dogs are both running and playing on the grass. The dog on the left is named \textcolor{cgreen}{\textlangle my dog\textrangle}. }}
\end{minipage}
} \\
\midrule
\makecell[c]{
\begin{minipage}[b]{0.3\linewidth}
    \centering
    {\includegraphics[width=4.6cm]{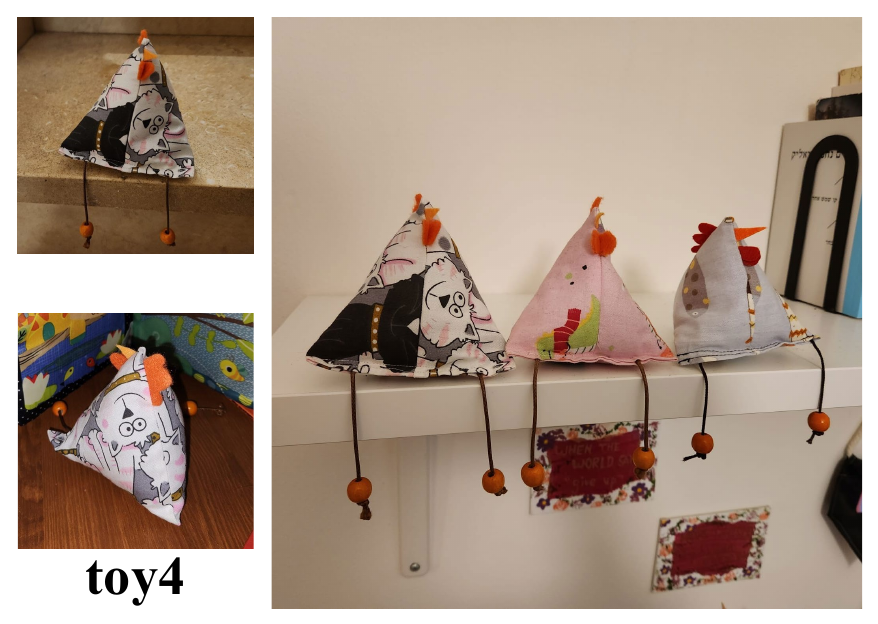}}
\end{minipage}
}& \makecell[c]{
\begin{minipage}[b]{0.62\linewidth}
    {Phi3-V: The image features a white shelf with several decorative items on it. There are three chicken-shaped decorations hanging from the shelf, each with a different design. Additionally, there are three more chicken decorations placed on the shelf, making \dots
    \vspace{0.5mm}\\
   \adjustbox{bgcolor=Gray,minipage=[t]{\linewidth}}{ \textbf{RAP-Phi3-V}(Ours): The image shows three handmade stuffed toys. The toys are all shaped like chickens and are hanging on a white shelf. The toys are all different colors and have different designs on them. The toy on the left is called \textcolor{cgreen}{\textlangle toy4\textrangle}.} }
\end{minipage}
} \\
\midrule
\makecell[c]{
\begin{minipage}[b]{0.3\linewidth}
    \centering
    {\includegraphics[width=4.6cm]{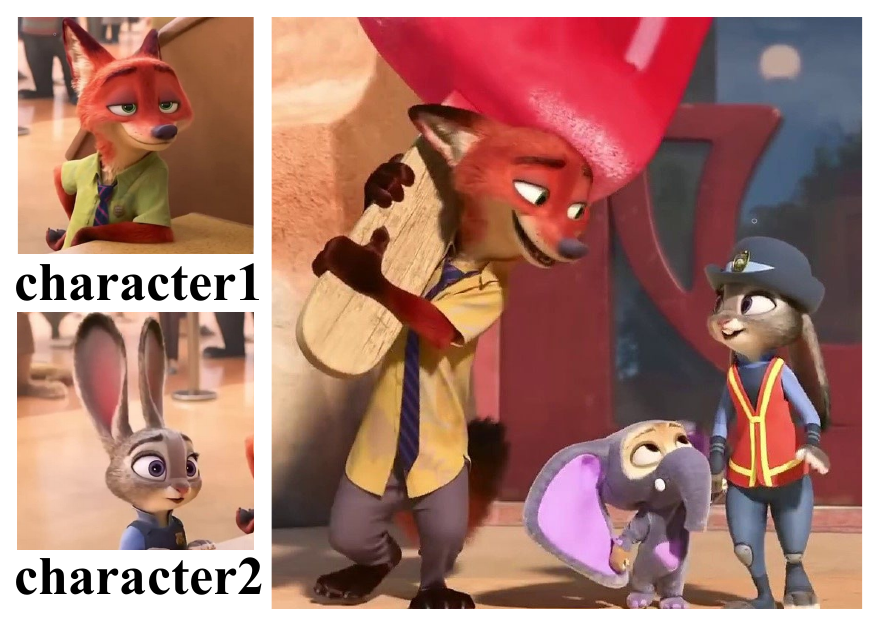}}
\end{minipage}
}& \makecell[c]{
\begin{minipage}[b]{0.62\linewidth}
    {Phi3-V: The image features a group of three cartoon characters, two of which are wearing hats. They are standing together on a sidewalk, with one character positioned towards the left side of the frame, another in the center, and \dots
    \vspace{0.1mm}\\
    \adjustbox{bgcolor=Gray,minipage=[t]{\linewidth}}{\textbf{RAP-Phi3-V}(Ours): The image shows \textcolor{cgreen}{\textlangle character1\textrangle} and \textcolor{cgreen}{\textlangle character2\textrangle} standing in front of a brick wall. \textcolor{cgreen}{\textlangle character1\textrangle} is a cartoon fox wearing a yellow button-down shirt and a purple tie. He has big eyes, fluffy fur, and a sly grin on his face. \textcolor{cgreen}{\textlangle character2\textrangle} is a cartoon rabbit with large ears, a police officer's hat, and a red and yellow safety vest. She has a sweet, innocent look on \dots}}
\end{minipage}
} \\
\midrule
\makecell[c]{
\begin{minipage}[b]{0.3\linewidth}
    \centering
    {\includegraphics[width=4.6cm]{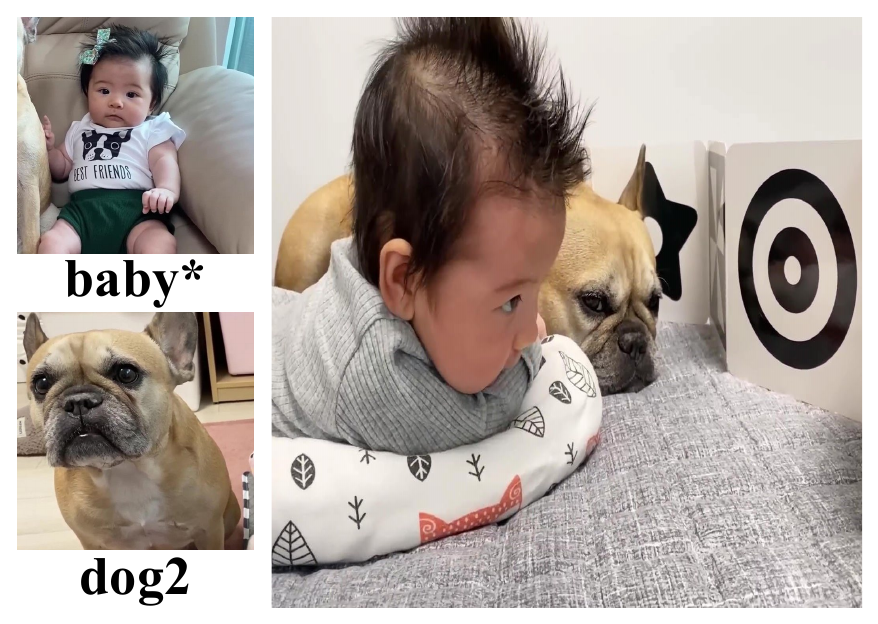}}
\end{minipage}
}& \makecell[c]{
\begin{minipage}[b]{0.62\linewidth}
    {Phi3-V: The image features a baby lying on a bed, with a dog resting on the floor nearby. The baby appears to be playing with a toy, possibly a puzzle, while the dog watches attentively \dots
    \vspace{0.5mm}\\
    \adjustbox{bgcolor=Gray,minipage=[t]{\linewidth}}{\textbf{RAP-Phi3-V}(Ours): The image shows a baby, \textcolor{cgreen}{\textlangle baby*\textrangle}, sleeping in a carrier. The baby is wrapped in a colorful blanket and is secured by a black strap. A french bulldog, \textcolor{cgreen}{\textlangle dog2\textrangle}, is sleeping next to the baby. The baby and dog are both sleeping peacefully.}}
\end{minipage}
} \\
\bottomrule
\end{tabular}
\end{table*}

% Recognition
\begin{table*}[t]
\centering
\caption{Qualitative results of personalized concept recognition obtained by RAP-LLaVA. We use \textcolor{cgreen}{green} rectangle to show the bounding box in the image.}
\label{tab:additional recognition}
\vspace{-3mm}
\begin{tabular}{cc}
\toprule
Image & Conversation\\
\midrule
\makecell[c]{
\begin{minipage}[b]{0.3\linewidth}
    \centering
    {\includegraphics[width=4.6cm]{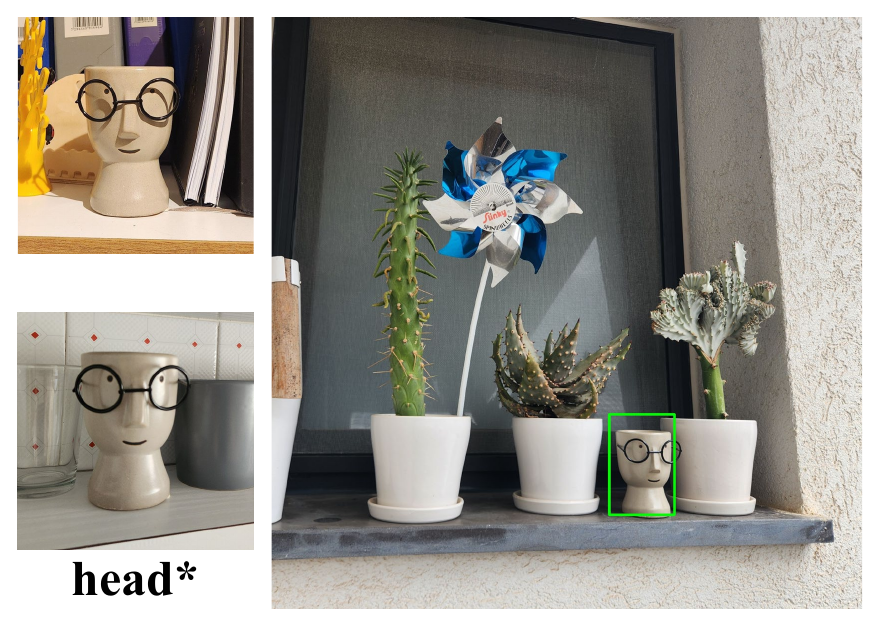}}
\end{minipage}
}& \makecell[c]{
\begin{minipage}[b]{0.62\linewidth}
    {Question: Is \textlangle head*\textrangle\ in this image?
    \vspace{1mm}\\
    \adjustbox{bgcolor=Gray,minipage=[t]{\linewidth}}{\textbf{RAP-LLaVA}: Yes, \textlangle head*\textrangle\ is in the photo.} \vspace{2mm}\\
    Question: Give \textlangle head*\textrangle's bounding box in the image.
    \vspace{1mm}\\
    \adjustbox{bgcolor=Gray,minipage=[t]{\linewidth}}{\textbf{RAP-LLaVA}: [0.57, 0.67, 0.68, 0.84]} 
}
\end{minipage}
} \\
\midrule
\makecell[c]{
\begin{minipage}[b]{0.3\linewidth}
    \centering
    {\includegraphics[width=4.6cm]{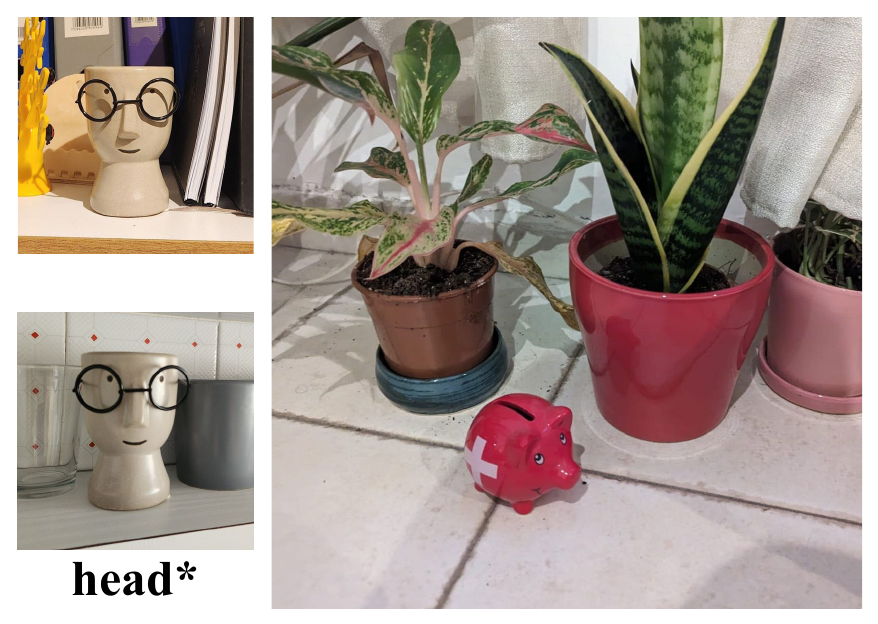}}
\end{minipage}
}& \makecell[c]{
\begin{minipage}[b]{0.62\linewidth}
    {Question: Is \textlangle head*\textrangle\ in this image?
    \vspace{1mm}\\
    \adjustbox{bgcolor=Gray,minipage=[t]{\linewidth}}{\textbf{RAP-LLaVA}: No, \textlangle head*\textrangle\  is not part of this photograph.} }
\end{minipage}
} \\
\midrule
\makecell[c]{
\begin{minipage}[b]{0.3\linewidth}
    \centering
    {\includegraphics[width=4.6cm]{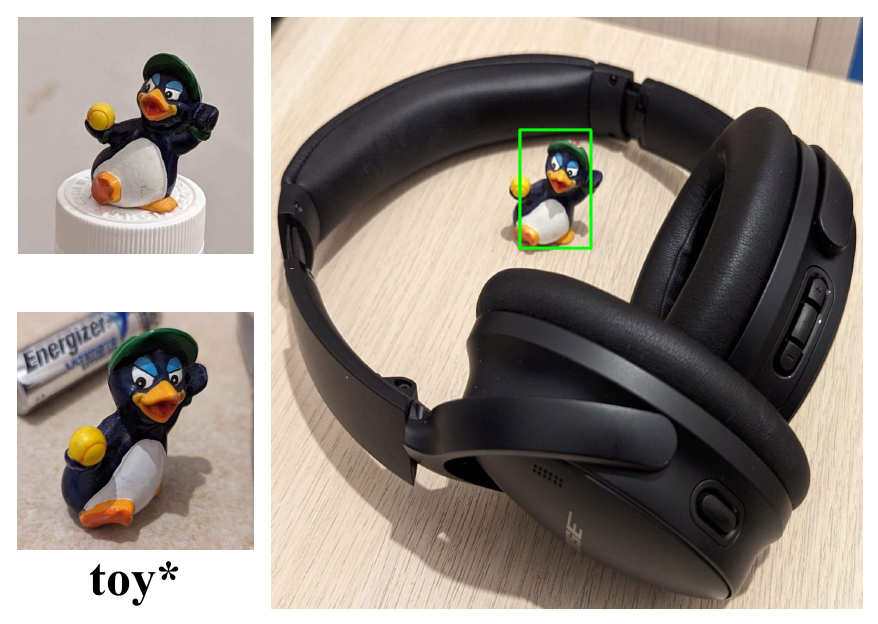}}
\end{minipage}
}& \makecell[c]{
\begin{minipage}[b]{0.62\linewidth}
    {Question: Can you identify if \textlangle toy*\textrangle\ is captured in this picture?
    \vspace{1mm}\\
    \adjustbox{bgcolor=Gray,minipage=[t]{\linewidth}}{\textbf{RAP-LLaVA}: Yes, \textlangle toy*\textrangle\ is visible in this picture.}
    \vspace{2mm}\\
    Question: Please provide \textlangle toy*\textrangle's bounding coordinates in the image.
    \vspace{1mm}\\
    \adjustbox{bgcolor=Gray,minipage=[t]{\linewidth}}{\textbf{RAP-LLaVA}: [0.42, 0.19, 0.54, 0.39]}
    }
\end{minipage}
} \\
\midrule
\makecell[c]{
\begin{minipage}[b]{0.3\linewidth}
    \centering
    {\includegraphics[width=4.6cm]{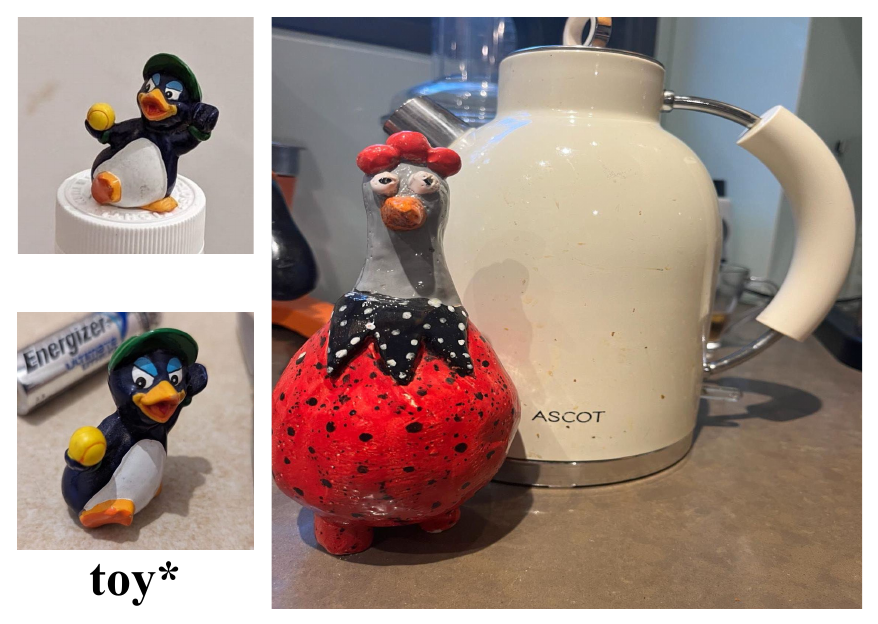}}
\end{minipage}
}& \makecell[c]{
\begin{minipage}[b]{0.62\linewidth}
    {Question: Can you identify if \textlangle toy*\textrangle\ is captured in this picture?
    \vspace{1mm}\\
   \adjustbox{bgcolor=Gray,minipage=[t]{\linewidth}}{ \textbf{RAP-LLaVA}: \textlangle toy*\textrangle\ is not captured in this image.} }
\end{minipage}
} \\
\midrule
\makecell[c]{
\begin{minipage}[b]{0.3\linewidth}
    \centering
    {\includegraphics[width=4.6cm]{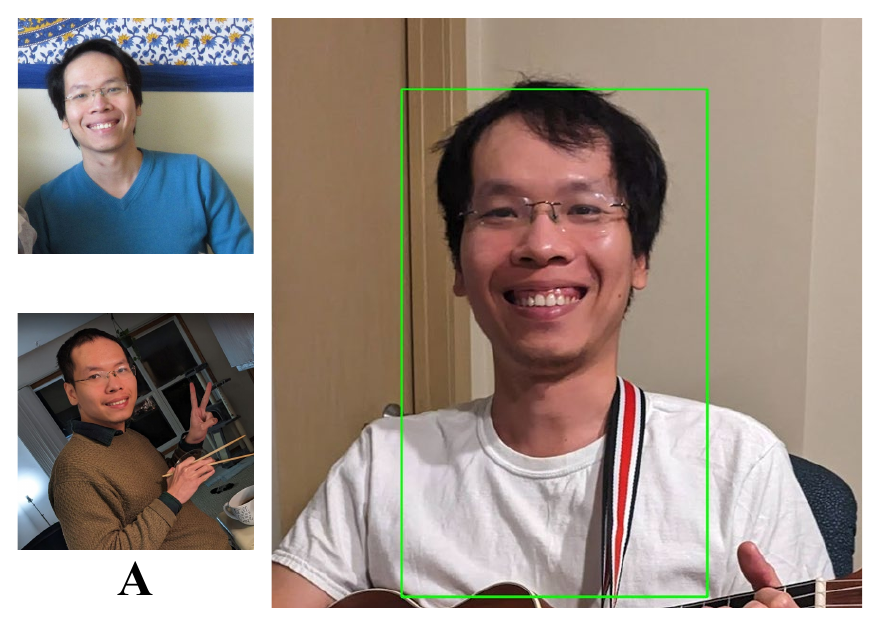}}
\end{minipage}
}& \makecell[c]{
\begin{minipage}[b]{0.62\linewidth}
    {Question: Can you confirm if \textlangle A\textrangle\ appears in this photo?
    \vspace{1mm}\\
    \adjustbox{bgcolor=Gray,minipage=[t]{\linewidth}}{\textbf{RAP-LLaVA}: Yes, \textlangle A\textrangle\ is in this photo.}
    \vspace{2mm}\\
    Question: Describe \textlangle A\textrangle’s position in the image.
    \vspace{1mm}\\
    \adjustbox{bgcolor=Gray,minipage=[t]{\linewidth}}{\textbf{RAP-LLaVA}: [0.24, 0.12, 0.72, 0.98]}
    }
\end{minipage}
} \\
\midrule
\makecell[c]{
\begin{minipage}[b]{0.3\linewidth}
    \centering
    {\includegraphics[width=4.6cm]{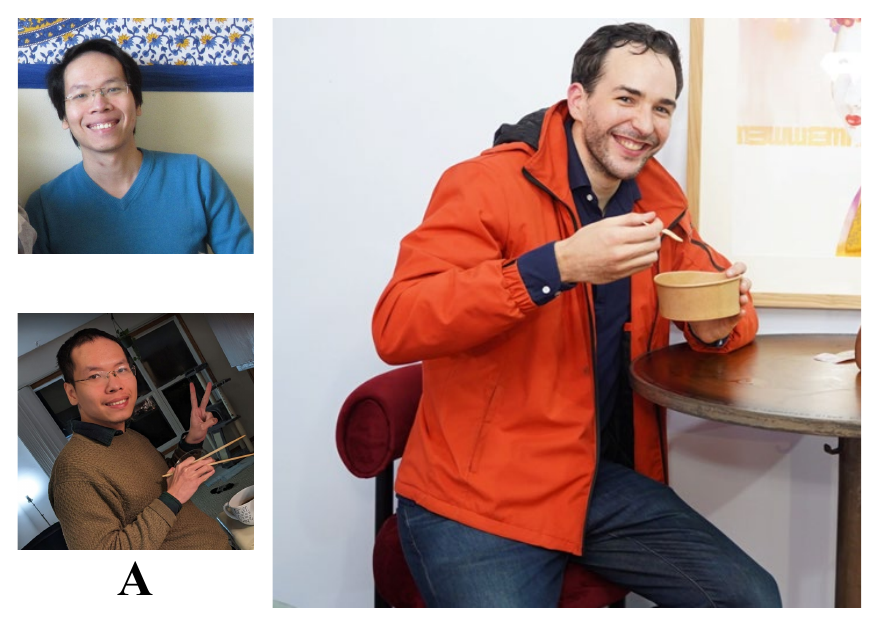}}
\end{minipage}
}& \makecell[c]{
\begin{minipage}[b]{0.62\linewidth}
    {Question: Can you confirm if \textlangle A\textrangle\ appears in this photo?
    \vspace{1mm}\\
    \adjustbox{bgcolor=Gray,minipage=[t]{\linewidth}}{\textbf{RAP-LLaVA}: \textlangle A\textrangle\ is not present in this photograph.}}
\end{minipage}
} \\
\bottomrule
\end{tabular}
\end{table*}
\begin{table*}[t]
\centering
\caption{Qualitative results of personalized concept recognition obtained by RAP-LLaVA. We use \textcolor{cgreen}{green} rectangle to show the bounding box in the image.}
\label{tab:additional recognition2}
\vspace{-3mm}
\begin{tabular}{cc}
\toprule
Image & Conversation\\
\midrule
\makecell[c]{
\begin{minipage}[b]{0.3\linewidth}
    \centering
    {\includegraphics[width=4.6cm]{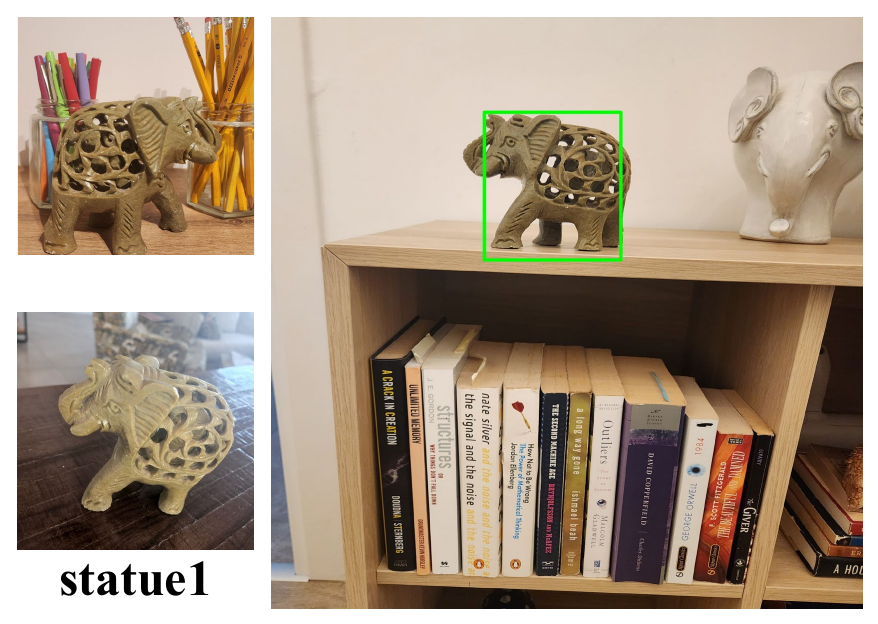}}
\end{minipage}
}& \makecell[c]{
\begin{minipage}[b]{0.62\linewidth}
    {Question: Is \textlangle statue1\textrangle\ included in this shot?
    \vspace{1mm}\\
    \adjustbox{bgcolor=Gray,minipage=[t]{\linewidth}}{\textbf{RAP-LLaVA}: Yes, \textlangle statue1\textrangle\ is included in this shot.} \vspace{2mm}\\
    Question: Give \textlangle statue1\textrangle's bounding box in the image.
    \vspace{1mm}\\
    \adjustbox{bgcolor=Gray,minipage=[t]{\linewidth}}{\textbf{RAP-LLaVA}: [0.36, 0.16, 0.59, 0.41]} 
}
\end{minipage}
} \\
\midrule
\makecell[c]{
\begin{minipage}[b]{0.3\linewidth}
    \centering
    {\includegraphics[width=4.6cm]{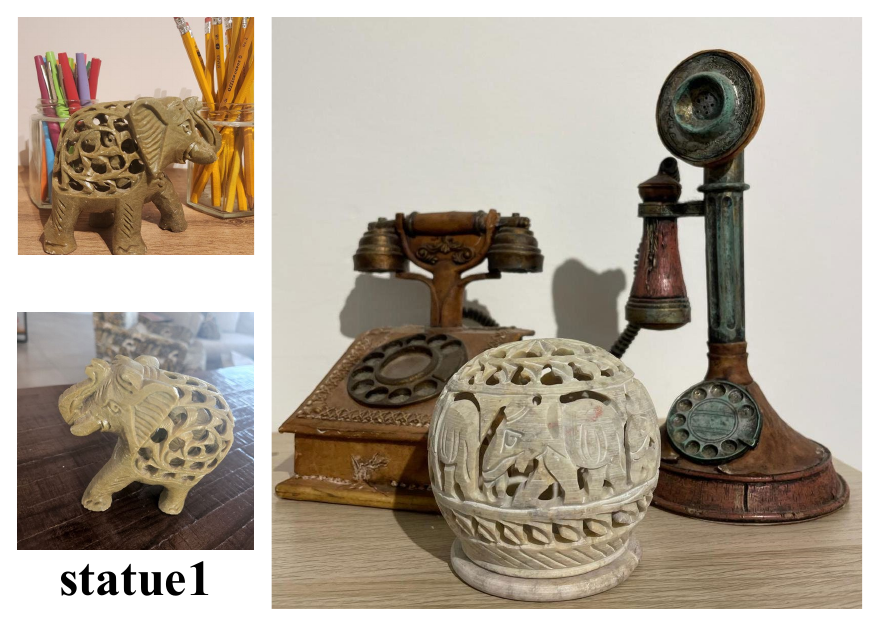}}
\end{minipage}
}& \makecell[c]{
\begin{minipage}[b]{0.62\linewidth}
    {Question: Is \textlangle statue1\textrangle\ included in this shot?
    \vspace{1mm}\\
    \adjustbox{bgcolor=Gray,minipage=[t]{\linewidth}}{\textbf{RAP-LLaVA}: \textlangle statue1\textrangle\ is not captured in this picture.} }
\end{minipage}
} \\
\midrule
\makecell[c]{
\begin{minipage}[b]{0.3\linewidth}
    \centering
    {\includegraphics[width=4.6cm]{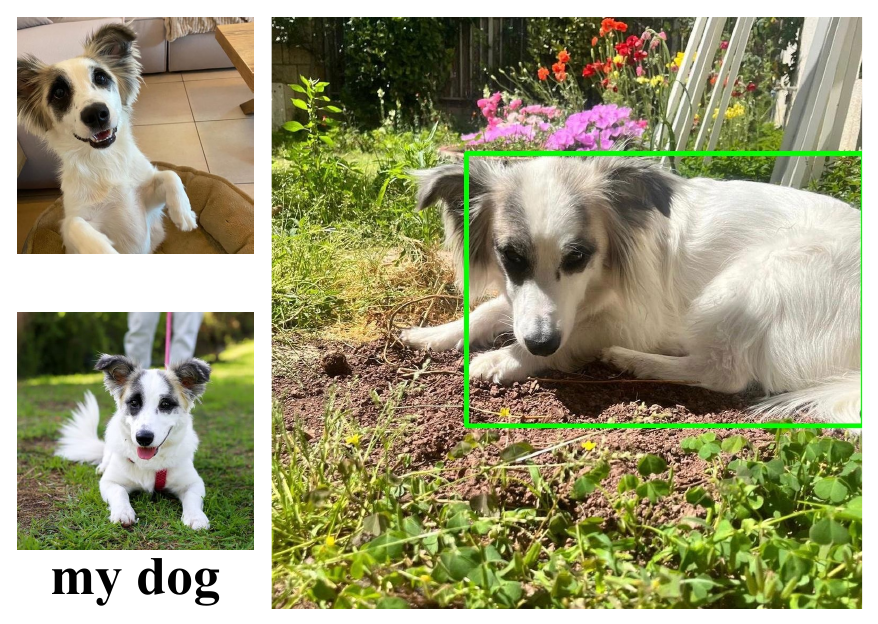}}
\end{minipage}
}& \makecell[c]{
\begin{minipage}[b]{0.62\linewidth}
    {Question: Is \textlangle my dog\textrangle\ captured in this image?
    \vspace{1mm}\\
    \adjustbox{bgcolor=Gray,minipage=[t]{\linewidth}}{\textbf{RAP-LLaVA}: Yes, I recognize \textlangle my dog\textrangle\ in this picture.}
    \vspace{2mm}\\
    Question: Specify \textlangle my dog\textrangle's bounding box in the photograph.
    \vspace{1mm}\\
    \adjustbox{bgcolor=Gray,minipage=[t]{\linewidth}}{\textbf{RAP-LLaVA}: [0.33, 0.23, 1.0, 0.69]}
    }
\end{minipage}
} \\
\midrule
\makecell[c]{
\begin{minipage}[b]{0.3\linewidth}
    \centering
    {\includegraphics[width=4.6cm]{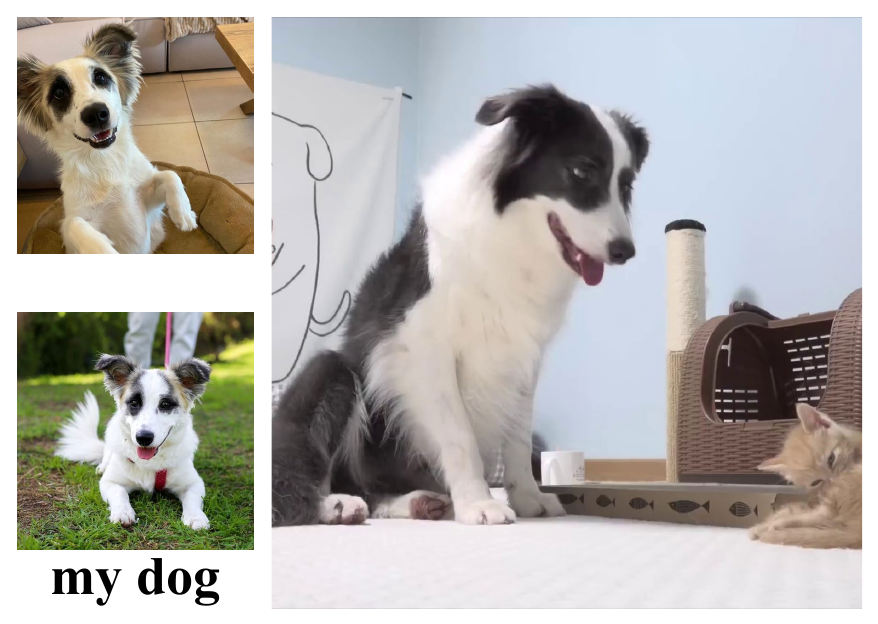}}
\end{minipage}
}& \makecell[c]{
\begin{minipage}[b]{0.62\linewidth}
    {Question: Is \textlangle my dog\textrangle\ captured in this image?
    \vspace{1mm}\\
   \adjustbox{bgcolor=Gray,minipage=[t]{\linewidth}}{ \textbf{RAP-LLaVA}: No, \textlangle my dog\textrangle\ is not included in this shot.} }
\end{minipage}
} \\
\midrule
\makecell[c]{
\begin{minipage}[b]{0.3\linewidth}
    \centering
    {\includegraphics[width=4.6cm]{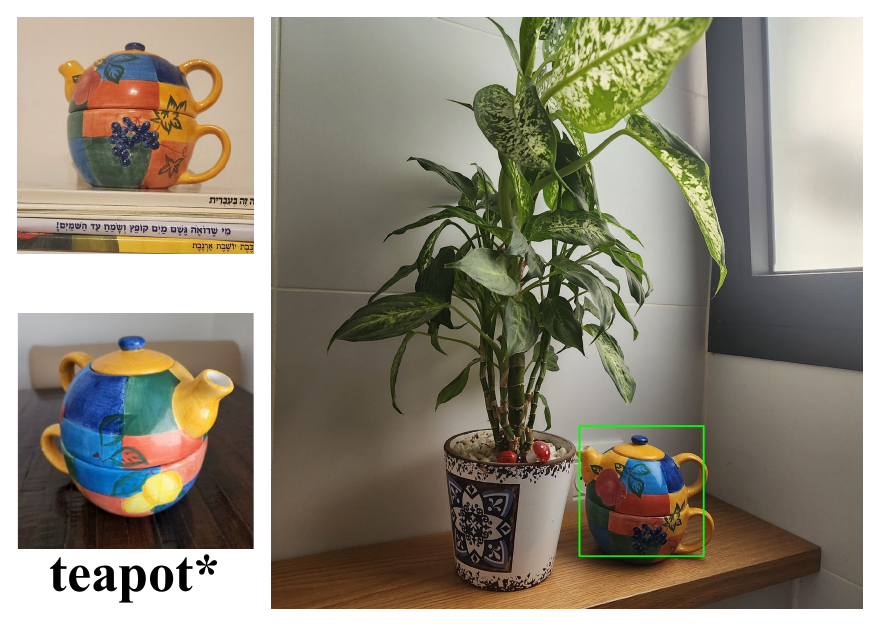}}
\end{minipage}
}& \makecell[c]{
\begin{minipage}[b]{0.62\linewidth}
    {Question: Is \textlangle teapot*\textrangle\ visible in this image?
    \vspace{1mm}\\
    \adjustbox{bgcolor=Gray,minipage=[t]{\linewidth}}{\textbf{RAP-LLaVA}: Yes, \textlangle teapot*\textrangle\ is visible in this picture.}
    \vspace{2mm}\\
    Question: Describe \textlangle teapot*\textrangle’s position in the image.
    \vspace{1mm}\\
    \adjustbox{bgcolor=Gray,minipage=[t]{\linewidth}}{\textbf{RAP-LLaVA}: [0.52, 0.69, 0.73, 0.91]}
    }
\end{minipage}
} \\
\midrule
\makecell[c]{
\begin{minipage}[b]{0.3\linewidth}
    \centering
    {\includegraphics[width=4.6cm]{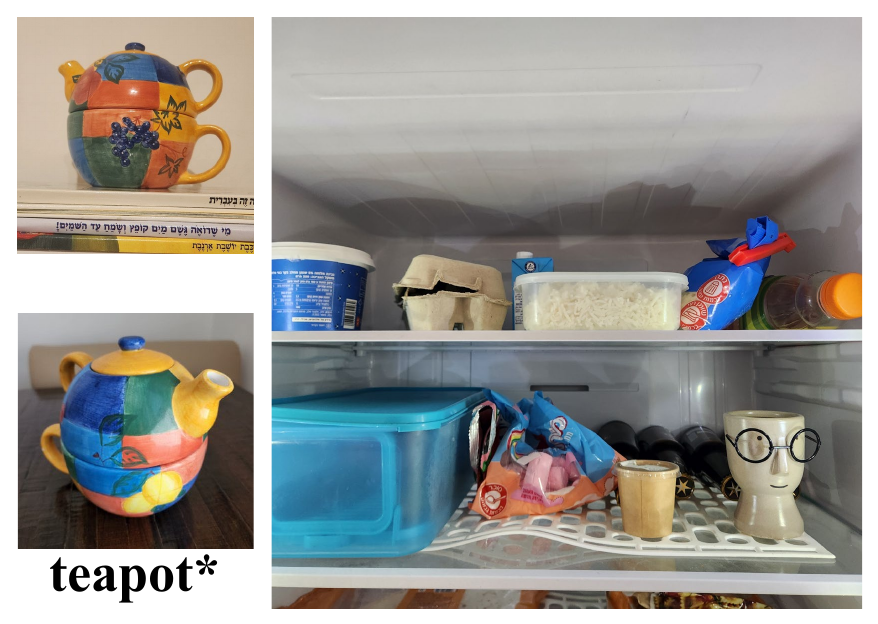}}
\end{minipage}
}& \makecell[c]{
\begin{minipage}[b]{0.62\linewidth}
    {Question: Is \textlangle teapot*\textrangle\ visible in this image?
    \vspace{1mm}\\
    \adjustbox{bgcolor=Gray,minipage=[t]{\linewidth}}{\textbf{RAP-LLaVA}: No, \textlangle teapot*\textrangle\ is not visible in this picture.}}
\end{minipage}
} \\
\bottomrule
\end{tabular}
\end{table*}

\begin{table*}[ht]
\centering
\caption{Instructions for visual grounding and recognition.}
\label{tab:inst_for_grounding}  
\vspace{-3mm}
\cornersize*{0.75cm}
\ovalbox{\parbox{130mm}{
\vspace{1mm}
\quad Visual recognition:
\vspace{1mm}
\begin{itemize}[leftmargin=18 pt, itemsep= 3 pt,topsep = 1pt]
\item Is $\left \langle \text{V} \right \rangle$ visible in this picture?
\item Is $\left \langle \text{V} \right \rangle$ in this image?
\item Do you see $\left \langle \text{V} \right \rangle$ in the photo?
\item Is $\left \langle \text{V} \right \rangle$ present in this photograph?
\item Can you identify if $\left \langle \text{V} \right \rangle$ is captured in this picture?
\item Is $\left \langle \text{V} \right \rangle$ depicted in this image?
\item Does the picture feature $\left \langle \text{V} \right \rangle$?
\item Can you confirm if $\left \langle \text{V} \right \rangle$ appears in this photo?
\item Is $\left \langle \text{V} \right \rangle$ included in this shot?
\item Is $\left \langle \text{V} \right \rangle$ shown in this image?
\item Can you tell if $\left \langle \text{V} \right \rangle$ is part of this photograph?
\item Is there any sign of $\left \langle \text{V} \right \rangle$ in this picture?
\item Can you detect $\left \langle \text{V} \right \rangle$ in the photo?
\item Is $\left \langle \text{V} \right \rangle$ captured in this image?
\item Do you recognize $\left \langle \text{V} \right \rangle$ in this picture?
\end{itemize}
\vspace{1mm}
\quad Visual grounding:
\vspace{1mm}
\begin{itemize}[leftmargin=18 pt, itemsep= 1 pt,topsep = 1pt]
\item Give \textlangle V\textrangle's bounding box in the image.
\item Describe \textlangle V\textrangle's position in the image.
\item Please provide the coordinates of the bounding box for \textlangle V\textrangle\ in the given image.
\item Specify the rectangular boundaries of \textlangle V\textrangle\ in the image.
\item Give \textlangle V\textrangle's position in the following image.
\item Please provide \textlangle V\textrangle's bounding coordinates in the image.
\item Indicate the bounding box for \textlangle V\textrangle\ in the image.
\item Show the bounding box for \textlangle V\textrangle\ in the picture.
\item Specify \textlangle V\textrangle's bounding box in the photograph.
\item Mark \textlangle V\textrangle's bounding box within the image.
\end{itemize}
}}
\end{table*}

\begin{table*}[ht]
  \centering
  \caption{Instructions for image captioning.}
  \label{tab:inst_caption}
  \vspace{-3mm}
\cornersize*{0.75cm}
\ovalbox{\parbox{130mm}{
\vspace{1mm}
\quad Image caption:
\vspace{1mm}
\begin{itemize}[leftmargin=18 pt, itemsep = 1 pt,topsep = 1pt]
\item Give a caption of the image.
\item Give a personalized caption of this image.
\item Provide a brief caption of the image.
\item Summarize the visual content of the image.
\item Create a short caption of the image.
\item Offer a short and clear interpretation of the image.
\item Describe the image concisely.
\item Render a concise summary of the photo.
\item Provide a caption of the given image.
\item Can you provide a personalized caption of this photo?
\item Could you describe this image concisely?
\end{itemize}
}}
\end{table*}

\begin{table*}[ht]
  \centering
  \caption{Instructions for image description.}
  \label{tab:inst_description}
  \vspace{-3mm}
\cornersize*{0.75cm}
\ovalbox{\parbox{130mm}{
\vspace{1mm}
\quad Image description:
\vspace{1mm}
\begin{itemize}[leftmargin=18 pt, itemsep= 1 pt,topsep = 1pt]
\item Describe the image.
\item Give a description of the image.
\item Give a description of the image in detail.
\item Give a short description of the image.
\item Describe the image in detail.
\item Please provide a description of the image.
\item Can you give me details about the image?
\item Could you explain what's shown in the image?
\end{itemize}
}}
\end{table*}

\begin{table*}[th]
\centering
\caption{Seed questions used for question answering synthesis.}
\label{tab:seed_q}
\vspace{-3mm}
\cornersize*{0.75cm}
\ovalbox{\parbox{130mm}{
\vspace{1mm}
\quad Person:
\vspace{1mm}
\begin{itemize}[leftmargin=18 pt, itemsep= 1 pt,topsep = 1pt]
\item What is \textlangle H\textrangle's hair color?
\item What is \textlangle H\textrangle's height (estimated)?                     
\item What is \textlangle H\textrangle's skin tone?
\item What is \textlangle H\textrangle's eye color?
\item What style of clothing is \textlangle H\textrangle\ wearing?
\item Does \textlangle H\textrangle\ have any visible tattoos?
\item Does \textlangle H\textrangle\ wear glasses or contact lenses?
\item Does \textlangle H\textrangle\ have any facial hair?
\item What is \textlangle H\textrangle's approximate age?
\item What is \textlangle H\textrangle's build or body type?
\item What is \textlangle H\textrangle\ doing?
\end{itemize}
\vspace{1mm}
\quad Object:
\vspace{1mm}
\begin{itemize}[leftmargin=18 pt, itemsep= 3 pt,topsep = 1pt]
\item What color is \textlangle O\textrangle?
\item What pattern is on \textlangle O\textrangle?
\item What shape does \textlangle O\textrangle\ have?
\item What size is \textlangle O\textrangle?
\item What is the texture of \textlangle O\textrangle?
\item Is \textlangle O\textrangle\ shiny or matte?
\item What material is \textlangle O\textrangle\ made of?
\item Does \textlangle O\textrangle\ have any patterns or designs on it?
\item Is \textlangle O\textrangle\ new or worn?
\item Does \textlangle O\textrangle\ have any visible brand or logo?
\item Is \textlangle O\textrangle\ functional or decorative?
\end{itemize}
\vspace{1mm}
\quad Multi-concept question:
\vspace{1mm}
\begin{itemize}[leftmargin=18 pt, itemsep= 3 pt,topsep = 1pt]
\item What do \textlangle C$_1$\textrangle\ and \textlangle C$_2$\textrangle\ have in common?
\item What activity are \textlangle C$_1$\textrangle\ and \textlangle C$_2$\textrangle\ engaged in?
\item Where could \textlangle C$_1$\textrangle\ and \textlangle C$_2$\textrangle\ be located?
\item What is the most noticeable difference between \textlangle C$_1$\textrangle\ and \textlangle C$_2$\textrangle?
\item What are they doing?
\end{itemize}}}
\end{table*}

\section{Limitation}
\label{appendix:limitation}
Our proposed RAP framework is a retrieval-based method. The limitations of RAP mainly concern the additional computational cost of generation and the precision of the retriever. While incorporating external information effectively generates more specific answers, it inevitably increases the context length for MLLMs, leading to additional computational overhead during the generation process. We will further explore ways to mitigate this computational burden. Another limitation is that the personalization performance of our RAP-MLLMs depends on the retriever's capability. This proposes a need for a robust multimodal retriever that can discern intricate features to enhance retrieval precision. Despite these limitations, RAP offers a timely solution for MLLM personalization. By retrieving from a user's specific database, RAP facilitates reliable and flexible personalized generation, which is valuable in practical applications.

\section{Potential Ethics Issues}
\label{appendix:ethics}
In our RAP framework, users can deploy and manage their own databases and personalized MLLMs locally. This design gives them greater flexibility to customize the model to fit their specific needs while ensuring privacy and security. However, improper use of personalized MLLMs can result in bias or other ethical concerns. For example, the model may unintentionally generate biased information or spread misleading or harmful content. To mitigate these risks, it is crucial to implement effective mechanisms for bias detection, fairness evaluation, and responsible model usage.

\section{Examples of the Personalized Database}
\label{appendix:database}
We give some visualized examples of our database in Table \ref{eg_database}. For each concept in the database, users need to provide an image with its name and optional text information. %Note that the concepts shown here such as doll* are just for readability, users can assign any names according to their preferences. 
During inference, the images, names and other information of retrieved concepts are integrated into the input for the MLLM. Users have the flexibility to define the name and personalized description based on their preferences, and our RAP-MLLMs will generate answers according to the provided information.

\begin{table*}[t]
\centering
\caption{Examples of our database. A concept should be provided with an image and its personalized description.}
\label{eg_database}
\vspace{-3mm}
\begin{tabular}{ccc}
\toprule
Image& Concept & Information\\
\midrule
\makecell[c]{
\begin{minipage}[b]{0.15\linewidth}
    \centering
    {\includegraphics[width=2cm, height=2cm]{figure/database/Anya.jpg}}
\end{minipage}
}&\makecell[c]{
\begin{minipage}[b]{0.12\linewidth}
    \centering
    {Anya}
\end{minipage}
} &\makecell[c]{
\begin{minipage}[b]{0.58\linewidth}
    {A young girl with pink hair and big green eyes.}
\end{minipage}
} \\
\midrule
\makecell[c]{
\begin{minipage}[b]{0.15\linewidth}
    \centering
    {\includegraphics[width=2cm, height=2cm]{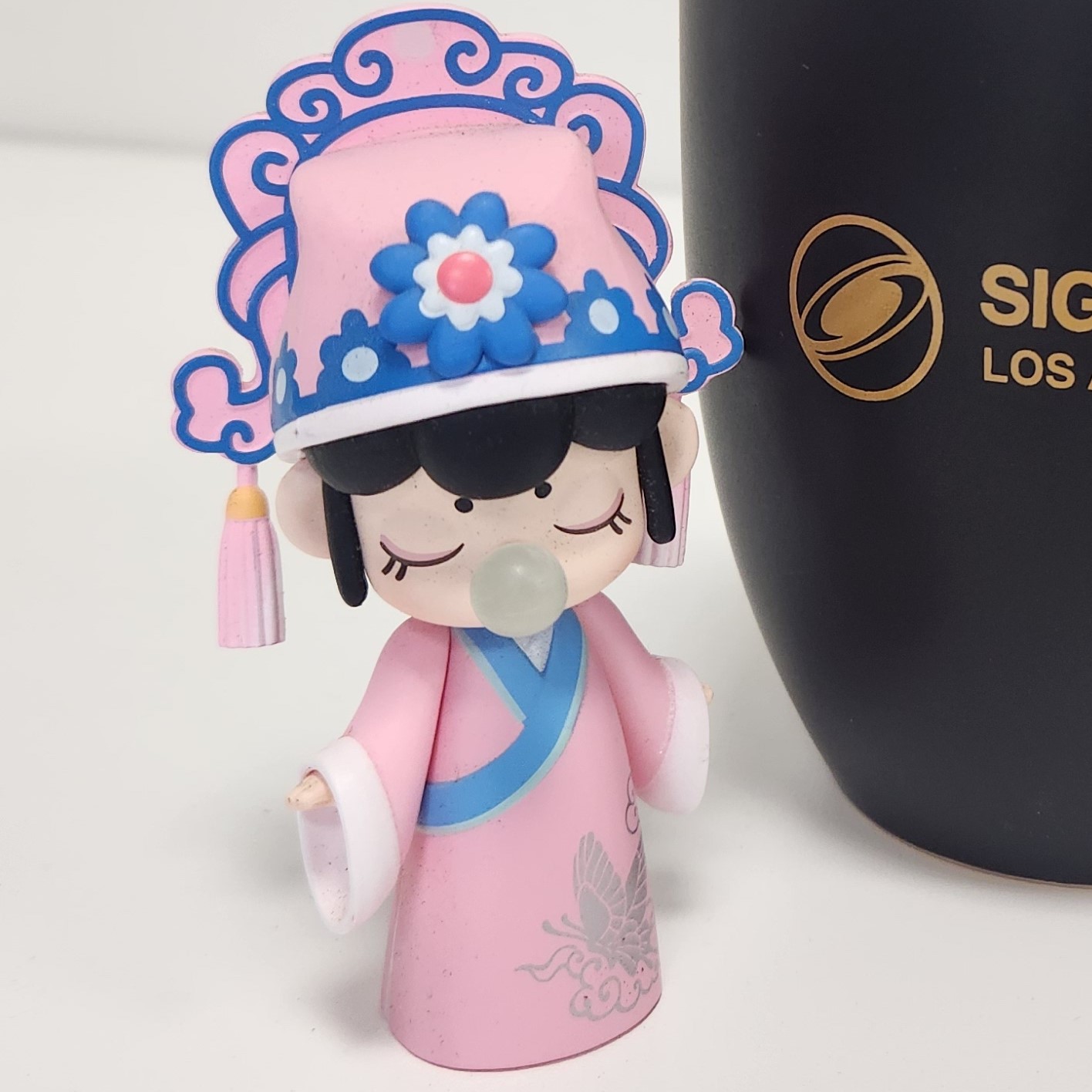}}
\end{minipage}
}&\makecell[c]{
\begin{minipage}[b]{0.12\linewidth}
    \centering
    {doll*}
\end{minipage}
} &\makecell[c]{
\begin{minipage}[b]{0.58\linewidth}
    {This is a cute figurine of a girl wearing a pink and blue dress, holding a white bubble.}
\end{minipage}
} \\
\midrule
\makecell[c]{
\begin{minipage}[b]{0.15\linewidth}
    \centering
    {\includegraphics[width=2cm, height=2cm]{figure/database/Bingo.jpg}}
\end{minipage}
}&\makecell[c]{
\begin{minipage}[b]{0.12\linewidth}
    \centering
    {toy1}
\end{minipage}
} &\makecell[c]{
\begin{minipage}[b]{0.58\linewidth}
    {A plush toy. It is orange with a yellow belly and a brown nose.}
\end{minipage}
} \\
\midrule
\makecell[c]{
\begin{minipage}[b]{0.15\linewidth}
    \centering
    {\includegraphics[width=2cm, height=2cm]{figure/database/Bluey.jpg}}
\end{minipage}
}&\makecell[c]{
\begin{minipage}[b]{0.12\linewidth}
    \centering
    {toy2}
\end{minipage}
} &\makecell[c]{
\begin{minipage}[b]{0.58\linewidth}
    {This is a plush toy of the bluey character. It is a light blue color with a purple patch on its head, and its ears are yellow.}
\end{minipage}
} \\
\midrule
\makecell[c]{
\begin{minipage}[b]{0.15\linewidth}
    \centering
    {\includegraphics[width=2cm, height=2cm]{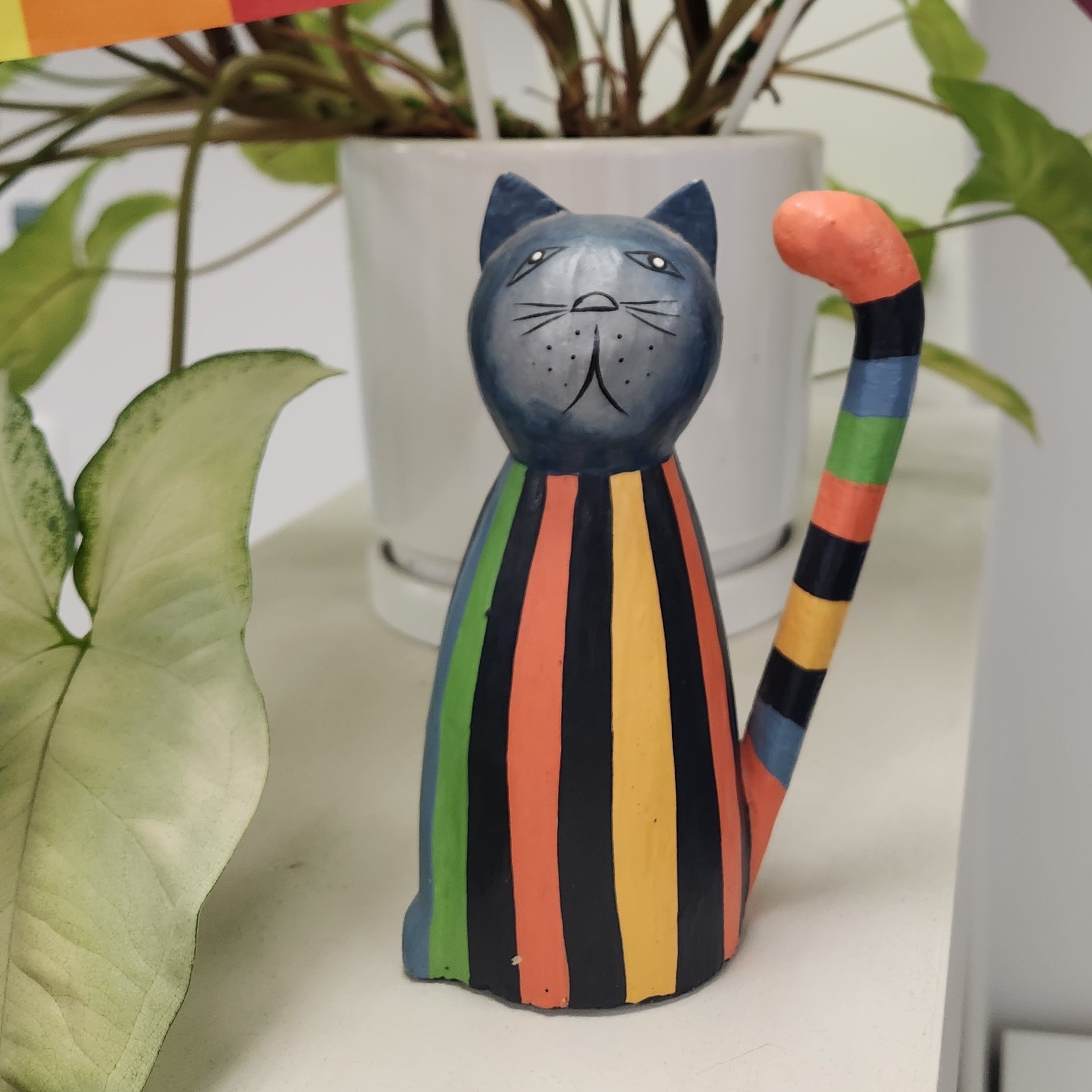}}
\end{minipage}
}&\makecell[c]{
\begin{minipage}[b]{0.12\linewidth}
    \centering
    {statue*}
\end{minipage}
} &\makecell[c]{
\begin{minipage}[b]{0.58\linewidth}
    {This is a figurine of a cat. The cat has a blue body with yellow, red, and green stripes and a long tail that is also striped.}
\end{minipage}
} \\
\midrule
\makecell[c]{
\begin{minipage}[b]{0.15\linewidth}
    \centering
    {\includegraphics[width=2cm, height=2cm]{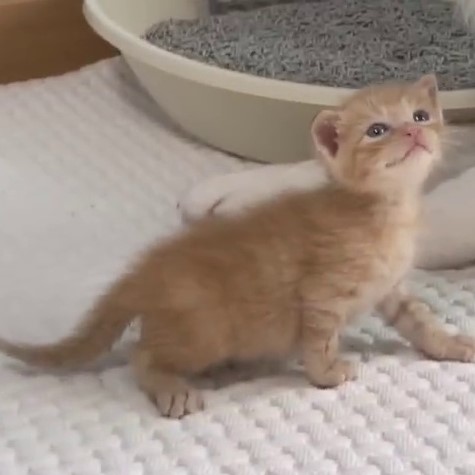}}
\end{minipage}
}&\makecell[c]{
\begin{minipage}[b]{0.12\linewidth}
    \centering
    {cat*}
\end{minipage}
} &\makecell[c]{
\begin{minipage}[b]{0.58\linewidth}
    {A small ginger kitten with bright blue eyes looks up at the camera.}
\end{minipage}
} \\
\midrule
\makecell[c]{
\begin{minipage}[b]{0.15\linewidth}
    \centering
    {\includegraphics[width=2cm, height=2cm]{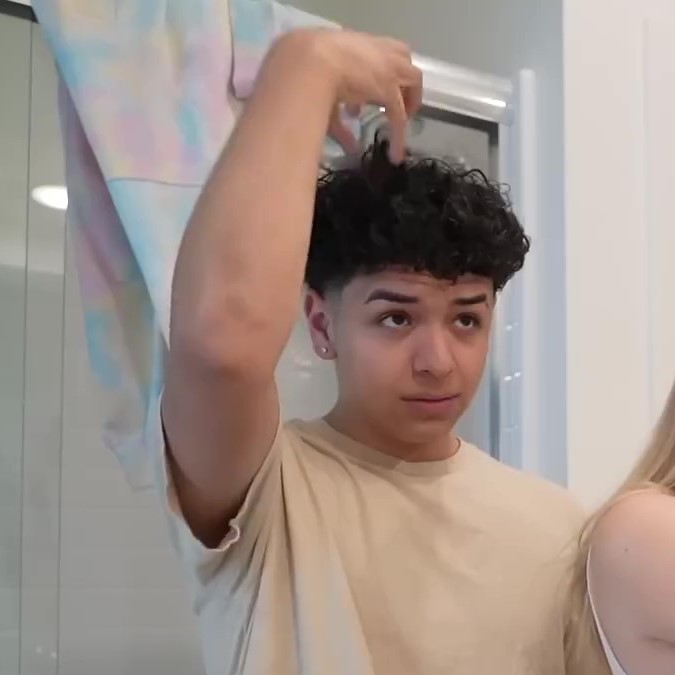}}
\end{minipage}
}&\makecell[c]{
\begin{minipage}[b]{0.12\linewidth}
    \centering
    {H}
\end{minipage}
} &\makecell[c]{
\begin{minipage}[b]{0.58\linewidth}
    {A young man is wearing a plain tan t-shirt. His hair is short and curly.}
\end{minipage}
} \\
\midrule
\makecell[c]{
\begin{minipage}[b]{0.15\linewidth}
    \centering
    {\includegraphics[width=2cm, height=2cm]{figure/database/maeve_dog.jpeg}}
\end{minipage}
}&\makecell[c]{
\begin{minipage}[b]{0.12\linewidth}
    \centering
    {my dog}
\end{minipage}
} &\makecell[c]{
\begin{minipage}[b]{0.58\linewidth}
    {A white and gray dog with long fur. He has black eyes.}
\end{minipage}
} \\
\midrule
\makecell[c]{
\begin{minipage}[b]{0.15\linewidth}
    \centering
    {\includegraphics[width=2cm, height=2cm]{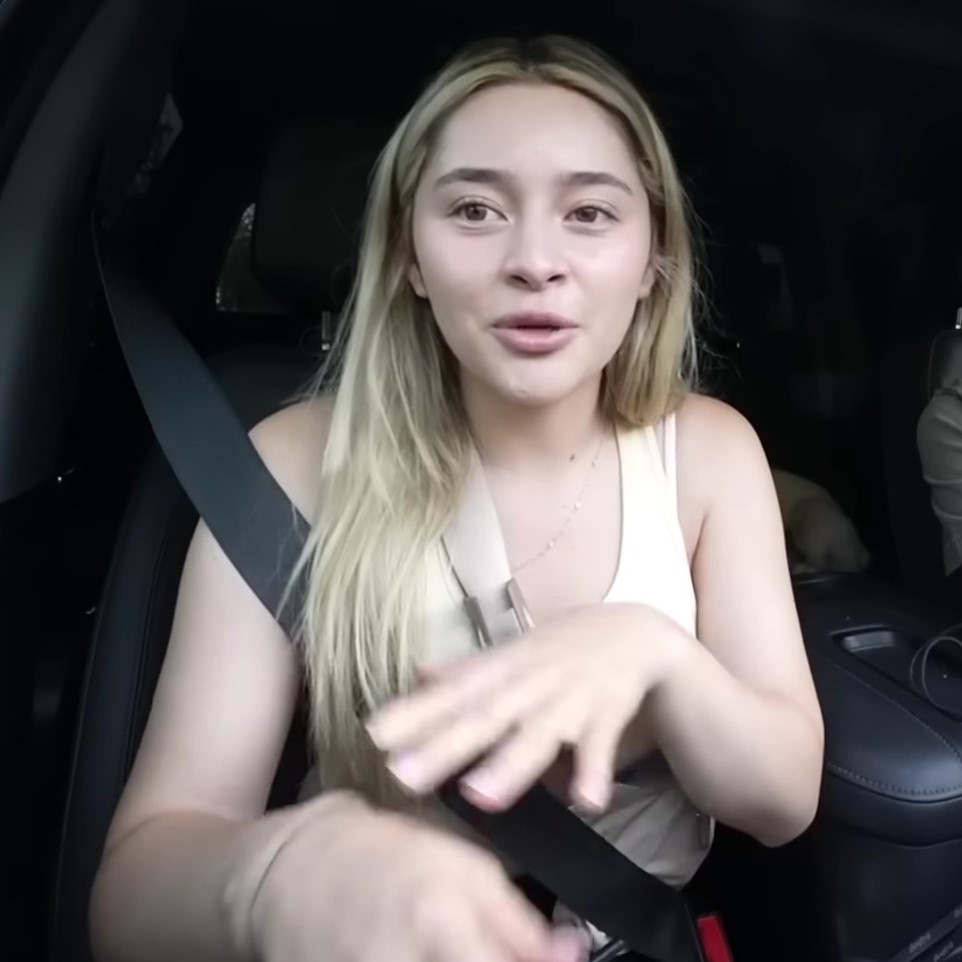}}
\end{minipage}
}&\makecell[c]{
\begin{minipage}[b]{0.12\linewidth}
    \centering
    {T}
\end{minipage}
} &\makecell[c]{
\begin{minipage}[b]{0.58\linewidth}
    {A young woman with blonde hair is wearing a white tank top and blue jeans.}
\end{minipage}
} \\
\bottomrule
\end{tabular}
\end{table*}

% WARNING: do not forget to delete the supplementary pages from your submission 

\end{document}